# Operational Framework for Recent Advances in Backtracking Search Optimisation Algorithm: A Systematic Review and Performance Evaluation

Bryar A. Hassan[1,3], Tarik A. Rashid[2]

[1]Kurdistan Institution for Strategic Studies and Scientific Research, Sulaimani, Iraq

[2]Computer Science and Engineering Department, University of Kurdistan Hewler, Erbil, Iraq

[3]Department of Information Technology, Sulaimani Polytechnic University, Sulaimani, Iraq

Email (corresponding): bryar.hassan@kissr.edu.krd

Email: tarik.ahmed@ukh.edu.krd

## Abstract

Backtracking search optimisation algorithm (BSA) is a commonly used meta-heuristic optimisation algorithm and was proposed by Civicioglu in 2013. When it was first used, it exhibited its strong potential for solving numerical optimisation problems. Additionally, the experiments conducted in previous studies demonstrated the successful performance of BSA and its non-sensitivity toward the several types of optimisation problems. This success of BSA motivated researchers to work on expanding it, e.g., developing its improved versions or employing it for different applications and problem domains. However, there is a lack of literature review on BSA; therefore, reviewing the aforementioned modifications and applications systematically will aid further development of the algorithm. This paper provides a systematic review and meta-analysis that emphasise on reviewing the related studies and recent developments on BSA. Hence, the objectives of this work are two-fold: (i) First, two frameworks for depicting the main extensions and the uses of BSA are proposed. The first framework is a general framework to depict the main extensions of BSA, whereas the second is an operational framework to present the expansion procedures of BSA to guide the researchers who are working on improving it. (ii) Second, the experiments conducted in this study fairly compare the analytical performance of BSA with four other competitive algorithms: differential evolution (DE), particle swarm optimisation (PSO), artificial bee colony (ABC), and firefly (FF) on 16 different hardness scores of the benchmark functions with different initial control parameters such as problem dimensions and search space. The experimental results indicate that BSA is statistically superior than the aforementioned algorithms in solving different cohorts of numerical optimisation problems such as problems with different levels of hardness score, problem dimensions, and search spaces. This study can act as a systematic and meta-analysis guide for the scholars who are working on improving BSA.

## 1. Background and Introduction

In artificial intelligence (AI), besides the existing traditional symbolic and statistical methods [1], computational intelligence (CI) has emerged as a recent research area [2]. According to [3], CI deals with collections of nature-inspired algorithms and approaches that in turn have been used to deal with complex practical problems that cannot be feasibly or effectively solved by traditional methods. CI comprises three subfields: fuzzy logic, evolutionary computation, and neural network.

The concept of swarm intelligence (SI) is a part of evolutionary computation. It deals with artificial and natural systems that comprise several individuals and possess the ability of self-organisation and decentralised control. This concept, which was primarily initiated by Gerardo Beni and Jing Wang in 1989 in the context of cellular robotic systems, is used in AI [4]. Typically, the SI system comprises a collection of agents. The agents interact with their own environment or with each other. SI systems often draw inspiration from the biological systems and nature [5,6]. Consequently, the ideas of termite and ant colonies, fish schooling, bird flocking, evolution, and human genetics have been used in the SI algorithms. The idea of genetic algorithm (GA) based on the bio-inspired operators was proposed by Emanuel Falkenauer [7], whereas differential evolution (DE) inspired by evolutionary development was invented by Storn and Price [8]. Furthermore, an algorithm inspired by the ant foraging behaviour in the real-world is called ant optimisation algorithm [9], whereas an algorithm based on bird flocking that is called particle swarm optimisation (PSO) was introduced by Kennedy and Eberhart [10]. Moreover, several creatures have motivated the SI scholars to develop original optimisation algorithms.

Additionally, some human artefacts such as multi-robot systems and some computer programs to solve data analysis and optimisation problems can also be considered to fall in the SI domain. In this context, Pinar Civicioglu introduced BSA as an EA [11]. BSA is one of the recent evolutionary algorithms (EAs) used to solve and optimise numerical problems. It was developed in 2013, and since then, several scholars have cited and elaborated it in the SI research community. Consequently, several research articles on the developments and applications of BSA are being published each year. Fig. 1 depicts the number of research articles on BSA that have been published since 2013 on the platforms of Google Scholar, Web of Science, and Scopus databases. The Figure illustrates the cumulative growth of BSA citations in the literature from the last few years. The number of published articles citing the standard BSA reached the peak in 2016, followed by 2018 and 2017. Overall, the number of articles based on BSA has increased yearly. The increasing use of BSA has attracted more researchers to study its development, and different types of applications and problems have been solved effectively and efficiently by using this metaheuristic algorithm. The current work mainly uses Google Scholar, Web of Science, Scopus, ScienceDirect, EBSCO Information Services, ResearchGate, and IEEE Xplore Digital Library.

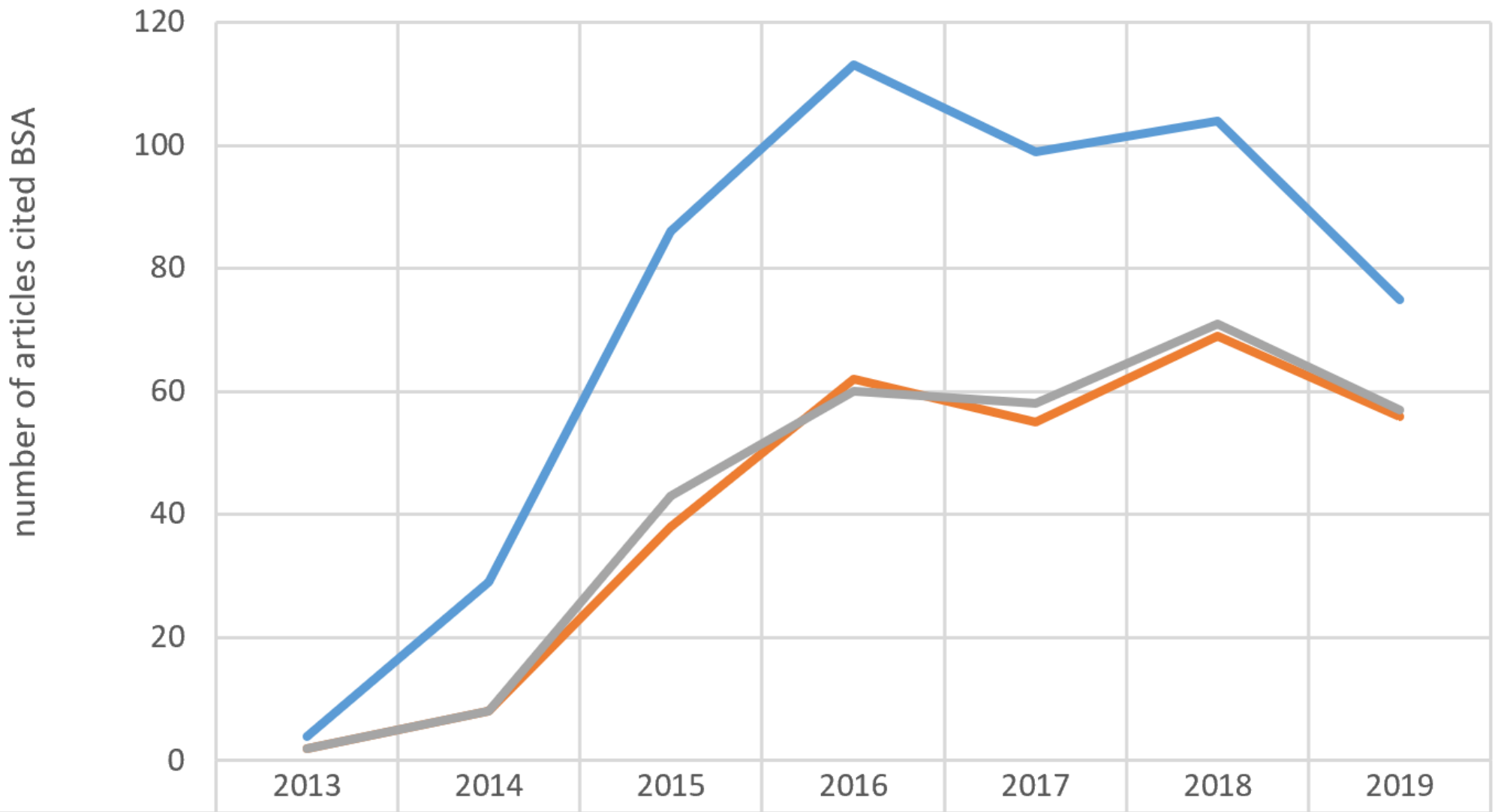

| | 2013 | 2014 | 2015 | 2016 | 2017 | 2018 | 2019 |
|---|---|---|---|---|---|---|---|
| GoogleScholar | 4 | 29 | 86 | 113 | 99 | 104 | 75 |
| Web of Science | 2 | 8 | 38 | 62 | 55 | 69 | 56 |
| Scopus | 2 | 8 | 43 | 60 | 58 | 71 | 57 |

**Fig. 1: Number of yearly published articles cited the standard BSA algorithm since 2013**

Owing to the adaptability of BSA to different applications and optimisation problems, several scholars have proposed new algorithms based on the original BSA, whereas few others have attempted to employ the original BSA in different applications to solve a variety of problems. Numerous research articles are focusing on the advancements of BSA such as modifications, hybridisation, and extensions. The number of published articles on the modifications of BSA is considerably higher than those on hybridising the BSA with other techniques. BSA was primarily used to deal with numerical optimisation problems; however, currently, it has been extended to deal with multi-objective, constrained, and binary optimisation problems. Currently several published articles on extending the BSA for different optimisation problems are available. Although BSA was proposed to solve numerical optimisation problems, it has also tackled several other optimisation problems. The problems tackled by BSA can be categorized into three types: the fundamental problems handled by this algorithm (n-queens problem, knight's tour puzzle, random word generator for the Ambrosia game, and maze generation) [12–16], travelling of salesperson (TSP) [17,18], and scheduling program [19,20].

Although BSA has been recently developed, it has already been used in several real-world applications and academic fields. The latest applications and implementation categories of BSA are control and robotics engineering, electrical and mechanical engineering, AI, information and communication engineering, and material and civil engineering. Additionally, it has been intensively used in the area of electrical and mechanical engineering. However, few articles have been published on BSA regarding the applications of AI, material and civil engineering, and control and robotics engineering, in decreasing order. It has also been affirmed by [21] that BSA is one of the most used EAs in the field of engineering. On this basis, BSA is considered as one of the most popular and widely used SI-based optimisation algorithms.

Furthermore, performance evaluation of BSA is an interesting research area. However, owing to several reasons, a fair and equitable performance analysis of BSA is a challenging task. These reasons can be selecting initialisation parameters of the competitive algorithms, system performance during the evaluation, programming style used in the algorithms, the balance of randomisation, and the cohorts and hardness score s of the selected benchmark problems. In this study, an experiment was conducted to fairly compare BSA with the other algorithms: PSO, artificial bee colony (ABC), firefly algorithm (FF), and DE by considering all the aforementioned aspects, owing to the lack of equitable experimental evaluation studies on BSA in the literature. Consequently, we focus on the uses and performance evaluation of BSA to provide a baseline of the studies. We also focus on the recent development of BSA to find footprints to guide scholars for conducting more studies in the future.

This paper is divided into eight sections. Section 1 introduces the fundamentals of EAs. BSA and its convergence analysis are described in Section 2. Recent studies on BSA and its advances such as its modifications, hybridisation, and extensions (single and multi-objective optimisations, binary optimisation, constrained optimisation, and numerical optimisation) are presented in Section 3. The leading implementations of BSA that include the common problems solved by BSA and its and real-world applications are discussed in Section 4. The analytical evaluation of BSA in the literature is reviewed in Section 5. Based on this, a general and operational framework regarding the main implementations and uses of BSA is proposed in Section 6. Section 7 discusses the experiments conducted in this study to fairly evaluate the analytical performance of BSA compared to other 4 comparable algorithms with respect to the initialisation parameters, problem dimensions, search space, and cohorts of the problems. Finally, the main contributions and shortcomings of BSA are discussed in Section 8. They are followed by the concluding remarks and a discussion of the future directions BSA to aid the scholars studying this algorithm.

## 2. Backtracking search optimisation algorithm

In recent years, BSA has gained significant attention owing to its various applications, especially in the fields of science and engineering. Backtracking was introduced in 1848 and was then known as ‘8-queens puzzle’. Further, Nauck created an 8*8 chessboard in 1950 for finding a feasible solution to this problem [15]. Later, Pinar Civicioglu introduced BSA as EA for solving numerical optimisation problems [11]. Currently, BSA is one of the popular EAs used widely for real-valued numerical, non-linear, and non-differentiable optimisation problems. The naive idea of backtracking originates in an improvement of the brute force approach. Backtracking automatically searches for an optimal solution to a problem among all the available solutions. The available solutions are presented as list of vectors, and the optimal solution is reached by moving back and forth or sideways in the vector’s domain. Pseudocode 1 outlines the structure of BSA.

```
1. Initialization
WHILE stopping criteria are not met DO
    2. Selection-I
    FOR generation of trial-population DO
        3. Mutation
        4. Corossover
5. Selection-II
6. Fitness evaluation
```

**Pseudocode 1: The structure of BSA**

Some websites related to BSA share the BSA sources codes in different programming languages and the ideas, publications, and latest advances on BSA. Such websites are listed in Table I.

**Table 1: Public websites on BSA**

| Website name | Reference, author(s) | URL |
|---|---|---|
| backtracking search optimization algorithm (BSA) | [11], P. Civicioglu | www.pinarcivicioglu.com |
| Mathworks for backtracking search optimization algorithm | [22], P. Civicioglu | www.mathworks.com |
| Backtracking tutorial using C program Code | [23], K. DUSKO | www.thegeekstuff.com |
| Applications of backtracking | [24], Huybers | www.huybers.net |

### 2.1. Standard BSA

BSA is a population-based EA that uses the DE mutation as a recombination-based strategy [25]. According to [11,26], BSA comprises six steps: initialisation, selection-I, mutation, crossover, selection-II, and fitness evaluation. Fig. 2 illustrates the flowchart of BSA. The steps in the flowchart are described in the following sub-sections.

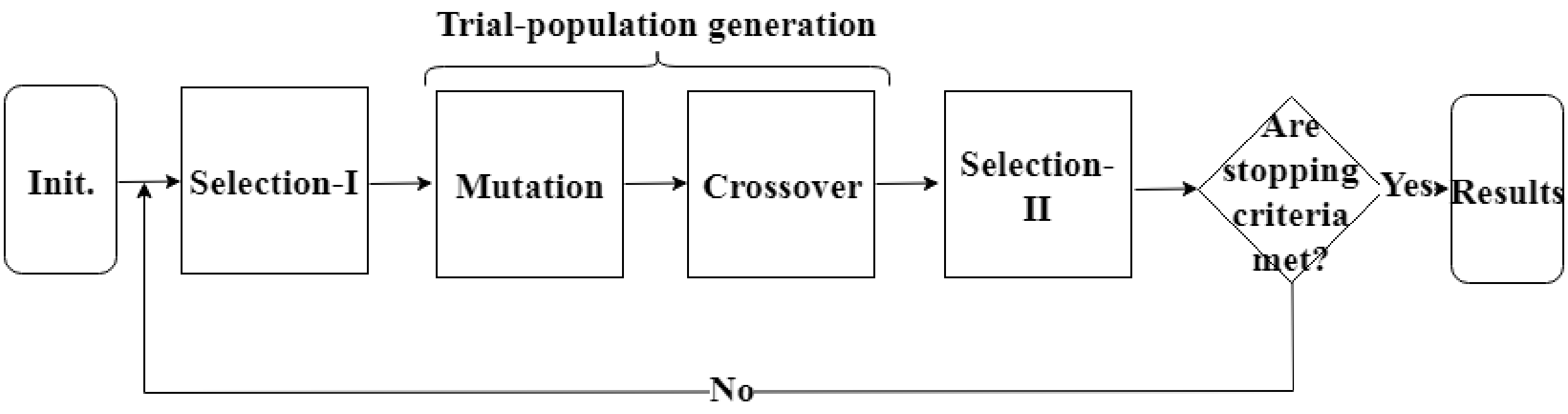

**Fig. 2: The flowchart of BSA**

**A. Initialisation:** The initial population ($P$) in BSA is generated randomly, and it comprises D variables and N individuals. It is expressed by Eq. (1) as follows:

$$P_{ij} \sim U(low_j,\ up_j) \qquad (1)$$

For $i=1, 2,...., N$ and $n=1, 2,...., D$, where $N$ is the population size, $D$ is the problem dimension, and $U$ is the uniform distribution.

**B. Selection-I.** There are two selection operators in BSA: the pre-selection operator, selection-I, and the post-selection operator, selection-II. The pre-selection operator is used to obtain the historical population ($P^{old}$) that is further used to calculate the direction of the search. The value of $P^{old}$ is calculated by using the following three steps:

1. The initial historical population is generated randomly by using Eq. (2) as follows:

$$P_{ij} \sim U(low_j,\ up_j) \tag{2}$$

2. At the beginning of each iteration, $P^{old}$ is redefined as by using Eq. (3) as follows:

$$P_{ij}^{old} \sim U(low_j, up_j) \tag{3}$$

   if $a < b$ then $P^{old} :=\ P|a, b \sim U(0,1)$,

   where: = is the update operation and $a$ and $b$ are random numbers.

3. After determining the historical population, the order of its individuals is changed by using Eq. (4) as follows:

$$P^{old} \coloneqq Permuting(P^{old}) \tag{4}$$

The permutation function is a random shuffling function.

**C. Mutation.** BSA generates the initial trial population (mutant) by applying the mutation operator as shown in Eq. (5).

$$Mutant = P + F.\ (P^{old} - P) \tag{5}$$

where F controls the amplitude of the search direction matrix ($P^{old} - P$).

**D. Crossover.** BSA's crossover generates the final form of the trial population ($T$). It comprises two steps:

1. A binary-valued matrix (map) is generated, where the map size is N X D, which indicates the individuals of trail population T.
2. The initial value of the binary integer matrix is set to 1, where $n \epsilon \{1,2,..., N\}$ and $m \epsilon \{1,2,..., D\}$.

The value of $T$ is updated by using Eq. (6) as follows:

$$T_{n,m}\ :=\ P_{n,m} \tag{6}$$

**E. Selection-II.** Selection-II is called BSA greedy selection. The individuals of trail population T are replaced with the individuals in population $P$ when their fitness values are better than those of the individuals in population $P$. The global best solution is selected based on the overall best individual having the best fitness value.

**F. Fitness evaluation.** A set of individuals is evaluated by the fitness evaluation. The list of individuals is used as input, and fitness is considered as the output.

These aforementioned processes are repeated except for the initialisation step until the stopping criteria are fulfilled. The stages of BSA are described in Fig. 3 as a set of processes. A population of the individuals *(P)* is created randomly in the initialisation step. Further, individual historical populations (*oldP*) are created in the Selection-I step. The value of *oldP* gets updated in this step, and further, the individuals of *oldP* are randomly re-arranged. The value of *oldP* remains unchanged until further iterations. Next, a population of the initial trial *(mutant)* is created from *oldP* and *P* by a mutation process. Further, by using *P* and *Mutant*, a population of the final cross-trial (*T*) is created. Finally, *P* gets updated with *T* individuals in the Selection-II step by considering the best fitness and selecting the corresponding individual. The steps from Selection-I to Selection-II are iterated until the stopping criteria are fulfilled. Fig. 3 illustrates detailed flowchart of BSA.

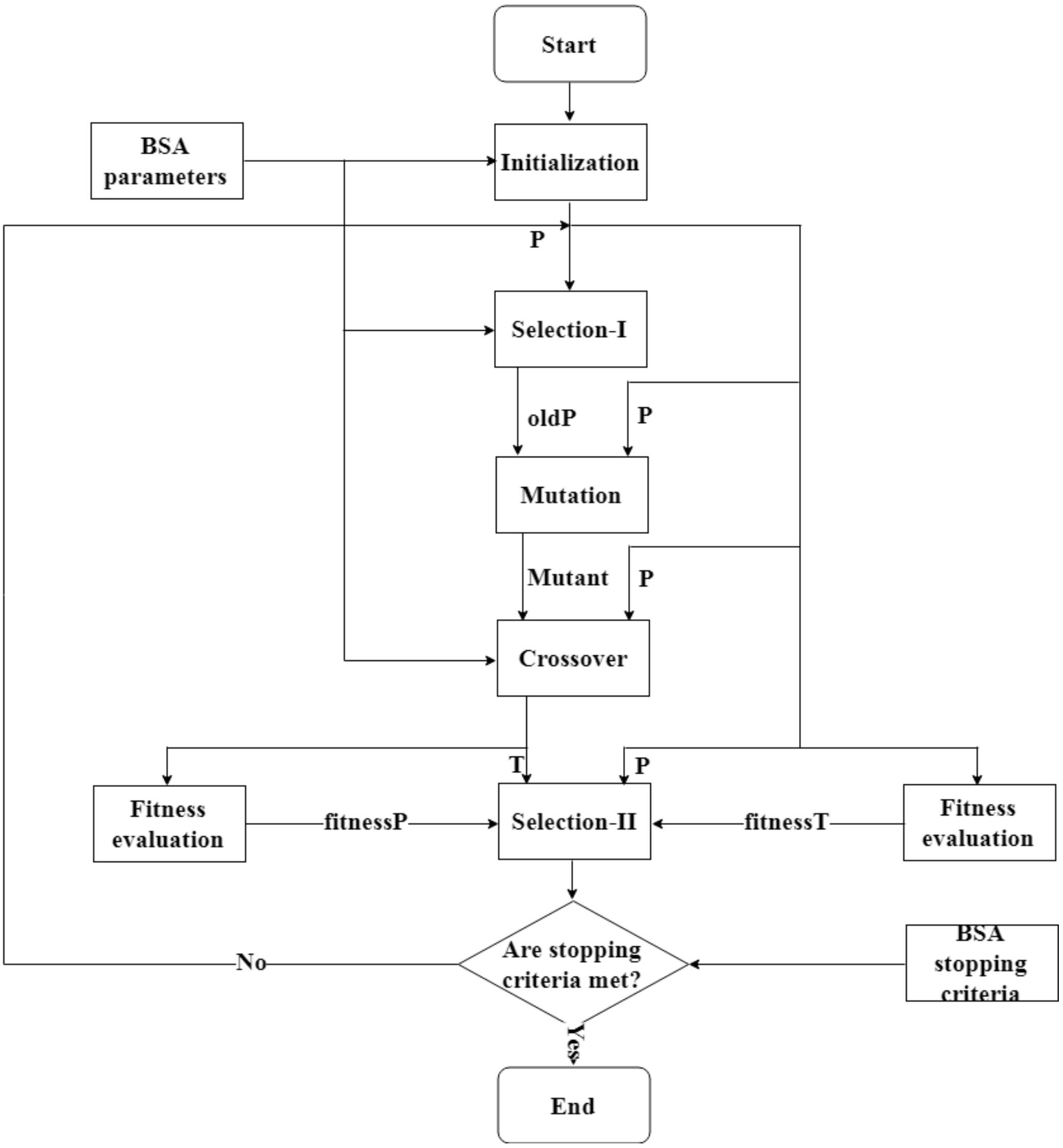

**Fig. 3: Detailed flowchart of BSA (adapted from** [26]**)**

## 2.2. BSA convergence analysis

Currently, most researchers are focusing on the performance improvement of BSA. Convergence analysis is an essential tool to analyse the speed of an algorithm during its iterative process [27,28]. This analysis process is usually represented as a convergence rate to find the desired solution. Therefore, the crossover and mutation parameters of BSA may differ owing to the dependence on the suitable solution of convergence. Additionally, it is recommended to adjust the scale factors within the mutation procedure [29]. This ensures that the convergence level and the variety of population are approached to their balance. In this manner, the BSA performance is improved by optimisation through an iterative process.

BSA can be mathematically analysed in the context of initialisation, selection, mutation, and crossover. The initial step of BSA selects current population, *P*, and historic population, *oldP*. This step is denoted by Eq. (7) as follows:

$$P_{ij} = P_j^{min} + \left(P_j^{max} - P_j^{min}\right) * rand\ (0, 1) \tag{7}$$

where, i = 1, 2, …, *N* and J = 1, 2, ……, *D*

To contribute more, *N* approaches the population size and *D* approaches the problem dimension. At this stage, the selection process chooses the historical population. It can be expressed by Eq. (8) as follows:

$$oldP \ := \begin{cases} P & a,b \\ oldP & otherwise \end{cases} \tag{8}$$

$$oldP \ := \ permutting\ (oldP) \tag{9}$$

Further, the BSA mutation procedure creates a real sample vector. The search direction matrix, (*oldP* – *P*), is estimated and controlled by F that is randomly generated within a standard normal distribution range. The BSA mutation is expressed by Eq. (10) as follows:

$$Mutant = P + F * (oldP - P) \tag{10}$$

Then, the BSA generates a heterogeneous and complicated crossover strategy. The crossover process creates the final version of the trial population. At the forefront, the matrix map of N * D binary integers is obtained. Moreover, depending on the BSA map value, the suitable size of the mutated individual is updated by using the corresponding individual in P. The crossover process can be expressed by Eq. (11) as follows:

$$V_{ij} = \begin{cases} P_{ij} & map_{ij}=1 \\ Mutant_{ij} & otherwise \end{cases} \tag{11}$$

The rate of crossover for a modified scaling factor is expressed in Eq. (12) [30]

$$F = F_{min} + (F_{max} - F_{min}) \frac{F_{i\,max} - F_{i\,min}}{F_{0\,max} - F_{0\,min}} \tag{12}$$

Therefore, the rate of convergence of BSA is expressed by minimising the sequence of the error term among the current and the historical solution. It is known that if the sequence approaches 0, then it provides an insight into how speedily the convergence happens. The convergence analysis can be explained as the Marciniak and Kuczynski (M-K) method [31–34].

$$X_{n+1} = Z_n - \frac{f\ (Zn)}{f\ [Z_n, y_n\ ] + f\ [z_n, x_n, y_n](z_n - y_n)} \quad (13)$$

Assume that the error in $x_n$ is $e^n = x_{n-}$ α with the beneficial approach of Taylor expansion, $x = \alpha$. The convergence is six, and the order is seven

$$f_{(x_n)=f'}\ (\alpha)\ [e_n + c_2\, e_n^2 + c_3\, e_n^3 + c_4\, e_n^4 + c_5\, e_n^5 + c_6\, e_n^6 + c_7\, e_n^7 + O(e_n^8)] \quad (14)$$

In addition

$$f_{(x_n)=f'}\ (\alpha)\ [1 + 2\, c_2\, e_n^2 + 3\, c_3\, e_n^3 + 4\, c_4\, e_n^4 + 5\, c_5\, e_n^5 + 6\, c_6\, e_n^6 + 7\, c_7\, e_n^7 + O(e_n^7)] \quad (15)$$

whereas,

$$C_k = \frac{1}{k}\ !\frac{\left(f^k\ (\alpha)\right)}{f'\ (\alpha)}, k = 2, 3, \dots.. \quad (16)$$

Therefore, we can estimate the iterative method by using the M-K method.

## 3. Variants of BSA

Several optimisation problems have been successfully tackled by using BSA. Researchers have been motivated to optimise the performance of the standard BSA owing to its claimed limitations in terms of deficiency and the speed of convergence. A considerable number of scholars have proposed new algorithms based on the original BSA, whereas some others have attempted to employ it in different applications to solve various problems. In this section, we focus on the advancements of the BSA variants and formally published researches conducted on them. These variants are categorised into three: modifications, hybridisation, and extensions of BSA. In Fig. 4, a general framework to depict the main variants of BSA is proposed.

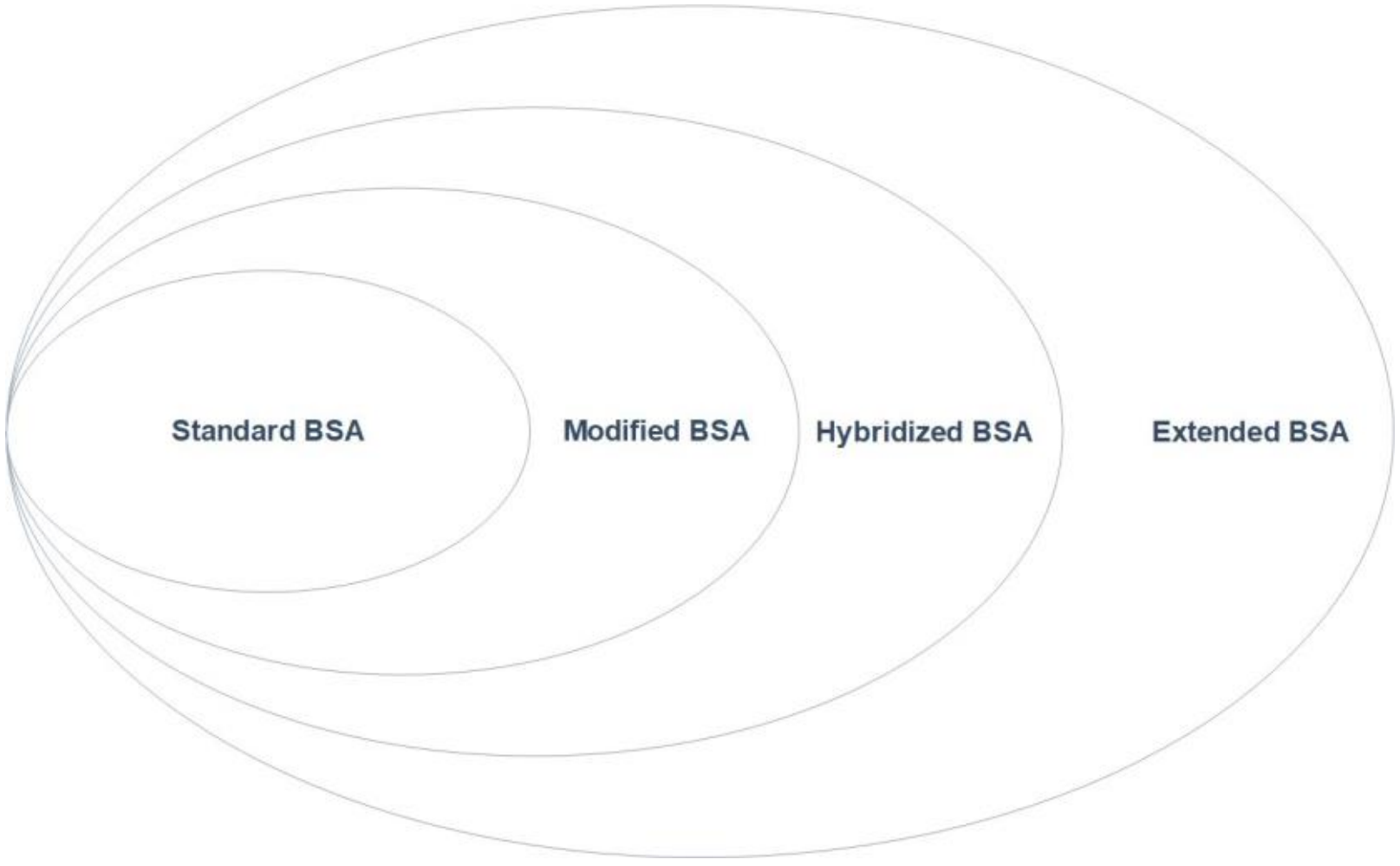

**Fig. 4: The proposed general framework for main extensions of BSA**

### 3.1. Modifications of BSA

To improve the effectiveness and performance of BSA, and suit with its use in a specific application, scholars have modified BSA many times. This section describes seven modified algorithms based on BSA. The modified algorithms are: BSAISA, COBSA, BBSA, HMOBSA, BSA based CVRP, SSBSA, LBSA, IBSA(s), the improved BSA versions, MBSA(s), chaos-based BSA, BSA-based neuro-fuzzy network, MLBSA, SBSA, modified BSA, BSA-IFOC, opposition-based BSAs, OBSAs, IG-BSA, Q-BSA, mBSAs, and other modifications. These algorithms are briefed in Table 2.

**Table 2: The modified algorithms based on BSA**

| Modified algorithm | Source, author(s) | Aim of the improvement | Modified by using | Result |
|---|---|---|---|---|
| BSAISA | [35], H. Wang et al. | Efficiency | Simulated annealing (SA) | Efficiency of BSA was improved |
| COBSA | [29], L. Zhao et al. | Convergence precision and speed | Control factor of the population to the equation of variation | A solution for improvement in the algorithm in terms of convergence speed and precision was obtained. |
| BBSA | [36], M.S. Ahmed et al. | Managing home energy consumption | Optimal real-time schedule controller | Performance improvement was obtained by reducing not only the consumption of energy but also the electricity bill and saving energy during peak hours compared to other binary EAs. |
| HMOBSA | [20], C. Lu et al. | Effectively scheduling energy-efficient with controllable transportation times to propose an energy-saving strategy | Selection II strategy | An effective algorithm with satisfactory performance was obtained. |
| BSA based CVRP | [37], M. Akhtar et al. | Efficiency reducing costs and emissions, and enhanced collection | Vehicle routing problem (CVRP) | Efficiency of the BSA-based CVRP was improved. |
| SSBSA | [38], H. Wang et al. | Convergence performance and exploitation of BSA | SA and species evolution principle | The performance of SSBSA improved compared to BSA and other competitive algorithms. |
| LBSA | [39], D. Chen et al. | Convergence speed, application domain maximisation | Populations | The performance of LBSA was improved compared to BSA and other classical algorithms. It also was more effective |
| IBSA | [30], S. Nama et al. | Performance, applicability to solve engineering problems | Control parameter (F) | IBSA's performance was enhanced, and it was appropriate for solving engineering problems |
| IBSA | [40], J. Kartite, and M. Cherkaoui | Mitigating some limitations of BSA, minimising renewable energy costs | A new mutation operator | The effectiveness of IBSA was demonstrated. |
| The improved BSA versions | [41], H.-C. Tsai | Effectively minimising the diversity of optimisation problems | Mutation strategies and one-dimensional crossover design | The improved BSA versions outperformed its counterpart algorithms (PSO, UPS, ABC, SaDE, and joint approximate diagonalization of eigen (JADE)) concerning convergence precision. |
| MBSA | [42], J. Lu, and J. Ding | Optimising output weights of deep stochastic configuration network (DSCN) for prediction intervals (PIS) constructions | Population diversity and searchability expansion | MBSA was able to deal with PIs with good quality |
| MBSA | [43], D. Chen et al | Enhancing performance of the standard BSA | Three primary strategies in combination with the standard BSA. | MBSA had considerably better performance compared with its counterpart algorithms |
| Chaos-based BSA | [44], Y.Ç. Kuyu, and F. Vatansever | Performance | Chaos maps were generated instead of generating random numbers | Chaos-based BSA outperformed the standard BSA. |
| BSA-based neuro-fuzzy network | [45], A. Chatzipavlis et al. | Performance | Neuro-fuzzy network utilised in the standard BSA | The BSA-based neuro-fuzzy network had better performance compared to other competitive methods for modelling beach realignment. |
| MLBSA | [46], K. Yu et al. | Estimating parameters of photovoltaic (PV) models | Multiple learning strategies to balance between exploitation and exploration capabilities | MLBSA had better performance compared with other algorithms in the literature concerning reliability, computational efficiency, and accuracy. |
| SBSA | [47], M.A. Ahandani et al. | Performance | Defining a new operator to the initial trial population, and providing effective group searchability | SBSA was entirely robust and effective |
| Modified BSA | [48], H. Yang et al. | Stability of PV microgrids | Stability of PV | The Modified BSA was effective for enhancing the stability of PV microgrids once it suffered from small disturbance. |
| BSA-IFOC | [49], J.A. Ali et al. | Performance | Indirect field-oriented control (IFOC) method for operating a traditional induction motor (TIM) | BSA-IFOC was more robust of BSA-IFOC than PSO. |
| Opposition-based BSAs | [50], Q. Xu et al. | Performance | Potential solution and its related opposite solution | The performance of the opposition-based BSAs was considerably enhanced compared to the standard BSA. |
| OBSAs | [51], J. Lin | Identifying parameters of hyperchaotic system | Combining the idea of opposition-based learning (OBL) into the standard BSA | OBSA was robust and effective. |
| IG-BSA | [52], M. Brévilliers et al. | Enhancing quality of the standard BSA | Populations | IG-BSA outperformed the standard BSA, SPSO2011, and CMAES |
| Q-BSA | [53], L.A. Passos et al. | Solving optimisation problems | Mapping the search space to a complicated space | Q-BSA was powerful and had fast convergence. |
| mBSAs | [54], S. Vitayasak et al. | minimising dynamic facilities layout problem (DFLP) with a heterogeneous resource. | minimising the combination of redesign costs and material flow | (i) mBSAs had better solution than GA for large-scale problems, and it the costs used to generate the layout of the best mBSA were considerably lower than for the standard BSA.<br>(ii) mBSAs enlarged the diversity of solutions, and increased exploration capability accordingly.<br>(iii) The execution time needed for the mBSAs was less than GA by 70%. |
| Other modifications | [55], H. Ferradi et al. | Performance | Multiplication algorithm | The performance was improved |

**3.1.1. BSAISA.** Some researchers proposed a modified BSA that is inspired by a probabilistic technique called SA to approximate its global optimum. This improves efficiency of BSA. For example, [35] proposed a modified version of BSA inspired by SA (BSAISA). The aim of BSAISA is to overcome the limitation of the current standard version of BSA by modifying its control factor amplitude (F) based on the criterion of the metropolis algorithm in SA and using a method called self-adaptive $\varepsilon$-constrained to manage the constraints. The performance of BSAISA was evaluated and compared with the standard BSA and other well-known EAs in solving several strictly constrained benchmarks and few problems in the field of engineering. The results showed that BSAISA was the most effective algorithm compared to the standard BSA and other well-known algorithms in terms of the convergence speed and the computation cost. However, BSAISA was limited in its robustness. The study suggests conducting further studies to improve the robustness of BSAISA. A future study could combine a niche technology with BSAISA [56,57]. Moreover, [58] presented a searching technique for the n-queens problem by using SA. The proposed technique led to a faster search than the standard BSA.

**3.1.2. COBSA.** Zhao et al. proposed COBSA based on the standard BSA for overcoming the inefficiency of convergence precision and speed in the standard BSA [29]. COBSA's improvement is achieved through the use of the control factor of the population to the equation of variation. The experiments conducted in [29] indicated that COBSA improved the standard BSA in terms of convergence speed and precision. However, it showed low convergence precision in a few benchmark functions in the experiment compared to other EAs such as ABC and modified ABC (MABC). COBSA's low convergence precision can be studied in the future to improve its performance on the related benchmark functions. Further, in [59], COBSA as an improved EA of BSA was proposed and implemented based upon the parameter estimation in a non-linear model of Muskingum, and the routing process of the flood was employed as a calibration of the optimal parameters. This improved COBSA showed high efficiency in calibrating the parameters of the non-linear Muskingum model compared to other methods. This study concluded that COBSA can offer improved performance for a precise solution and is more efficient compared to other algorithms. Statistically, it was confirmed that COBSA showed improved performance than PSO, GA, and DE for estimating the parameters of the non-linear Muskingum model. Therefore, the improved COBSA was able to achieve better convergence precision and speed, and offered an efficient alternative for estimating the parameters of the non-linear Muskingum model.

**3.1.3. BBSA.** In [36], a new algorithm, binary BSA (BBSA), which is based on BSA was proposed for home energy consumption management. The proposed algorithm used an optimal real-time schedule controller for managing the energy consumption at home. Experimental results showed that BBSA provided improved performance, thereby reducing the consumption of energy and electricity bill and saving energy during peak hours compared to other binary EAs such as BPSO.

**3.1.4. HMOBSA.** According to [20], BSA should be hybridised to be used for multi-objective optimisation. The proposed hybridised algorithm, called HMOBSA, aims to solve energy-efficient permutation flow shop scheduling problem. It proposes effective energy-saving strategy. The evaluation results showed that HMOBSA performed satisfactorily, and it was effective compared to non-dominated sorting genetic algorithm II (NSGA-II) and multi-objective EA based on the decomposition (MOEA/D). Additionally, the proposed energy-saving scenario showed improved performance compared to its competitors

**3.1.5. BSA based CVRP.** In waste management, collection and transportation of waste incur considerable expenses. Based on this, [37] suggested an algorithm—BSA with the CVRP approach for waste collection and routing optimisation. In this algorithm, the threshold waste level (TWL) with scheduling concepts was employed. The results showed that the algorithm reduced costs and emissions and enhanced the collection efficiency. Therefore, it can be implemented as a tool for optimising waste collection routes to reduce expenses and environmental damage. Additional studies considering more uncertainties and constraints can be undertaken in the future. Furthermore, to check the feasibility of the proposed algorithm, a pilot project could be implemented as a case study.

**3.1.6. SSBSA.** According to the recent study [38], BSA was modified to enhance its convergence performance and exploitation. The modified algorithm called SSBSA was inspired by SA and the species evolution principle. It proposes two strategies by providing an acceptance probability and a specified retain mechanism. In SSBSA, the historical population (oldP) and its amplitude control parameter (F) are preserved in accordance with the fitness feedback of the following iteration. Additionally, the adaptive control parameter (F) decreases with the increasing number of iterations, and it is redesigned through learning the acceptance probability. Consequently, the advantages of retaining the specified previous information and the new redesigned F improve the SSBSA's exploitation capability and speed-up the convergence rate. The proposed algorithm was executed on 14 benchmark functions and engineering problems to evaluate its performance. The results revealed the performance improvement of SSBSA compared to BSA and other competitive algorithms.

**3.1.7. LBSA.** A new algorithm, LBSA, with a modified structure of BSA was proposed [39]. It aims to improve the convergence speed and expand the application domain of BSA. In it, the historical information was combined with the global best information of the current generation to renew the individuals based on random probability. Additionally, the positions of the remaining individuals were renewed by learning the knowledge from the worst and best individuals, and other random individuals from the present generation. Consequently, LBSA has an advantage that some of its individuals learn from the best individuals to update their positions. Learning individuals from each other improved convergence speed and enlarged the population diversity. Finally, the performance of LBSA was tested on the CEC2014 and CEC2005 benchmark functions. The results showed that LBSA has superior performance compared to other classical algorithms, and it enhances the performance of the standard BSA.

**3.1.8. IBSA(s).** There are two BSA modifications named IBSA in the literature [30,40]. In [30], in an improvement over the standard BSA, a new adaptive amplitude parameter (F) was suggested. The newly improved algorithm is called IBSA. To perform the validation of IBSA, it was applied on the CEC2005 benchmark functions, and the results were compared with some existing algorithms. Further, it was used to minimise certain engineering problems on retaining the walls supporting the c-backfill. The results showed that IBSA was appropriate for solving problems such as the aforementioned ones. Additionally, simulation result indicated that IBSA showed superior performance with a high confidence level than the other algorithms. In another work [40], the standard BSA was improved to mitigate certain limitations of BSA. The improved algorithm, IBSA, was suggested to minimise the renewable energy costs. In this IBSA, a new mutation operator was designed based on the scaling factor value. Tests conducted to evaluate the performance of IBSA revealed the effectiveness of IBSA.

**3.1.9. Improved BSA versions.** Recently, improved BSA versions were suggested to minimise several optimisation problems effectively [41]. In the improved BSA versions, ten mutation strategies and one-dimensional crossover design were employed. Additionally, two parameters were used on the basis of best positions and historical mean, and three schemes of search factors were suggested. Experiments were performed to evaluate the performance of the newly introduced algorithm. The results indicated that the improved BSA version had superior performance than the standard BSA with respect to the convergence speed. Moreover, the improved BSA version outperformed its counterpart algorithms (PSO, UPS, ABC, SaDE, and JADE) in terms the convergence precision.

**3.1.10. MBSA(s).** There are two BSA modifications named MBSA in the literature [42,43] Firstly, as stated in [42], a new method based on the standard BSA was suggested for optimising the output weights of deep stochastic configuration networks (DSCNs) for the construction of prediction intervals (PIs). MBSA aims to improve the population diversity and searchability while maintaining the convergence speed. The performance evaluation of MBSA on standard datasets revealed that MBSA satisfactorily dealt with the PIs. Another modified BSA based on niching and learning strategies to enhance the performance of the standard BSA was proposed in [43]. Three primary strategies: niching, learning, and mutation were combined with the standard BSA to produce this MBSA. This MBSA was evaluated to examine its effectiveness on some well-known benchmark problems and three chaotic problems in the time series prediction on the basis of neural networks. The results revealed that MBSA showed improved performance compared with its counterpart algorithms for most of the benchmark functions and the chaotic problems.

**3.1.11. Chaos-based BSA.** In [44], BSA was modified based on the chaotic-based approach instead of some randomisation operations used in the algorithm. In the chaos-based BSA, values were generated via chaos maps instead of using the normal method of generating random values. To measure the statistical results and convergence rates of the chaos-based BSA, its performance was evaluated on 12 CEC2013 benchmark problems. The results indicated that the chaos-based BSA outperformed the original algorithm. The proposed model can be used further to minimise other optimisation problems.

**3.1.12. BSA-based neuro-fuzzy network.** In [45], the neuro-fuzzy network is utilised in the standard BSA on a few input variables correlated to beach morphology and wave forcing to model the beach realignment. The proposed approach is called BSA-based neuro-fuzzy network. The suggested model was evaluated on the data of a tourist beach. The experimental results indicated that the BSA-based neuro-fuzzy network showed superior performance compared to other competitive methods for modelling the beach realignment.

**3.1.13. MLBSA.** In [46], BSA was modified based on multiple learning strategies to balance between the exploitation and exploration capabilities. The modified algorithm, MLBSA, was used to determine the parameters of PV models such as the PV module and the diode. In the introduced algorithm, individuals learn from each other in two ways. Firstly, the individuals learn from the historical and the current population information concurrently to improve the exploration capability and preserve the population diversity. Secondly, the individuals learn the best individuals of the present population to enhance the exploitation capability and convergence speed. Experimental results indicated that MLBSA showed superior performance in terms of reliability, computational efficiency, and accuracy compared to the other algorithms in the literature.

**3.1.14. SBSA.** In [47], two concepts were proposed to enhance the performance of the standard BSA to identify the parameters of chaotic systems. The first concept was to define a new operator to the initial trial population, whereas the second was to provide effective group searchability. The suggested algorithm is called shuffled BSA (SBSA). In SBSA, the populations are classified into a number of groups, wherein which each group evolves itself based on the BSA's evolution process. This grouping process provides better search space exploration, and an autonomous search for each group expands the exploitation capability of BSA. The performance of SBSA was investigated by identifying the parameters for ten standard chaotic systems. The results indicated that both the suggested concepts were robust and effective. Additionally, SBSA provided promising and consistent results over repeated executions.

**3.1.15. Modified BSA.** In [48], a coordinated optimisation approach by considering the time delay effect of islanded PV microgrids, was proposed based on the modified BSA to enhance the stability of (PV microgrids. The experimental results indicated that the suggested approach was useful to enhance the stability of PV microgrids when they suffer from small disturbances.

**3.1.16. BSA-IFOC.** In [49], an adaptive proportional-integral (PI) controller with BSA was used to enhance the indirect field-oriented control (IFOC) method for operating a traditional induction motor (TIM). Additionally, BSA-IFOC was also exploited to minimise and tune the mean absolute error (MAE) to enhance the performance of TIM with respect to mechanical load and speed changes. The results of BSA-IFOC when compared with those of PSO showed improved robustness of BSA-IFOC for all the tests regarding the transient response and damping capability under different speeds and loads.

**3.1.17. Opposition-based BSAs.** Several opposition-based BSA were proposed in [50] to enhance the performance of the standard BSA. The key feature of these algorithms is that the potential solution and its corresponding opposite solution were considered concurrently to gain optimum approximation. The opposition-based BSAs were tested on 78 benchmark functions, and the results demonstrated that the performance of the opposition-based BSAs was considerably higher compared to the standard BSA.

**3.1.18. OBSAs.** An opposition BSA (OBSA) approach was proposed in another study [51]. The suggested algorithm was used to identify the parameters of the hyperchaotic system by combining the idea of opposition-based learning (OBL) with the standard BSA. OBSA was tested and compared with SPSO2011 and the standard BSA. The simulation results showed that OBSA was robust and effective.

**3.1.19. IG-BSA.** The authors of this study [52] proposed a new algorithm, Idol-BSA, to enhance the quality of the standard BSA. The modified algorithm was used the historical population to change some idols from the past generations. Additionally, the present population was developed under the effect of these idols. Further, the diversification strategy was added to the proposed algorithm. IG-BSA was tested on 19 benchmark functions. Experimental results showed that IG-BSA outperformed the standard BSA, SPSO2011, and CMAES.

**3.1.20. Q-BSA.** Quite recently, quaternion-based BSA was proposed to solve optimisation problems [53]. The proposed approach that is abbreviated as Q-BSA maps the search space to an extremely complex space such that each individual was represented by a quaternion. The features of quaternion algebra were used to minimise the optimisation problems. Q-BSA was tested on eight benchmark problems, and the results showed that it was powerful in terms of effectiveness and efficiency. Moreover, it had improved convergence speed compared to other counterpart techniques. Additionally, the authors intend to design octonion-based BSA that will be a tensor-based extension of BSA to map eight-dimensional entirely complex search space.

**3.1.21. mBSAs.** In [54], a novel modified BSA was outlined to solve dynamic facilities layout problem (DFLP) with a heterogeneous resource. The suggested algorithm, called mBSAs, aimed to minimise the combination of redesign costs and material flow. To validate its performance, three mBSAs were evaluated against GA and BSA using 7 benchmark functions. The result showed: (i) mBSAs had better solution than GA for large-scale problems, and it the costs used to generate the layout of the best mBSA were considerably lower than for the standard BSA. (ii) mBSAs enlarged the diversity of solutions, and increased exploration capability accordingly. (iii) The execution time needed for the mBSAs was less than GA by 70%. Additionally, the results indicated that BSA is particularly suitable for minimising large and complex engineering problems.

**3.1.22. Other modifications.** Some other modifications were performed on BSA. In [55], a new multiplication algorithm was used in the microprocessor by using backtracking search optimisation to determine the multiplication-friendly encoding of the operands. The results indicated that the new algorithm had superior performance in terms of speed over the classical ones.

### 3.2. Hybridisation of BSA

The combination of BSA and some other evolutionary or traditional algorithms is called hybridised BSA. The use of hybridised BSA is to take advantages of BSA as well as the other algorithms and overcome the weaknesses of each other. This section describes seventeen hybridized BSA algorithms, which are HHBSA, MBSOA, BPNN-BSA, HBSA, HBD, Fast Hybrid BSA-DE-SA, IBSA, BSABCM, BSANN, BSAF, BSA-WM, BGBSA, MFE-BSA, w-BSAFCM, BSA-SCA, BS-HH, BSOA-SVM, hDEBSA, E-BSADE, BSA-SQP, BSA-based MPPT algorithm, SRBSAMVS, BSA-based LSSVM, BSA-based ANN model, APSS&AMFS–BSA-ESN, BSA-SVMEEMD/VMD-HSBA-DAWNN, BSAxGWO, BSANCS, IHBSA, LSSVM-based IHBSA, ELM-BSA, MOBBSA, IBSA, VMD-BSA-RELM, BSO-FCM, SEBAL-BSA, BSA-based clustering algorithm, RBM-BSASA-BP, Hybrid BSA-SQP, HBSDE, BSA-NNRWs-N, BSA-DE, and BA-ACO. These hybridised algorithms are summarised in Table 3.

**Table 3: The hybridised algorithms based on BSA**

| Hybridised algorithm | Source, author(s) | Aim of the improvement | Hybridised by using | Result |
|---|---|---|---|---|
| HHBSA | [60], F. Zou et al. | Efficiency and performance of the standard BSA | Control parameter (F) | It had effective performance in global optimisation searches. |
| MBSOA | [61], A.F. Ali | Speeding up convergence and improving optimal solution | Random walk and direction exploitation algorithm (RWDE) | It was a successful and effective algorithm for solving non-linear problems, especially ELD problem. |
| BPNN-BSA | [62], M.A. Hannan et al.; [63], M.S.H. Lipu et al. | Robustness, accuracy, and performance | Back-propagation neural network (BPNN) | Performance of BPNN-BSA was improved. |
| HBSA | [64], Q. Lin et al. | Balancing between local and global searches | Combination of two local searches | Performance of HBSA was improved, and it was effective. |
| HBSA | [65], P. Chen et al. | Solving both energy consumption for PFSP and makespan | SA | HBSA was effective in comparison with GA and branch and bound algorithm. |
| HBSA | [66], S. Nama, and A. Saha | Minimising complex optimisation problems | Quadratic approximation (QA) | HBSA was statistically successful, and it had excellent performance compared to the variants of PSO. |
| HBSA | [67], S. Wang et al. | Solving global and numerical optimisations | Hooke-Jeeves pattern search | Efficiency and effectiveness of HBSA were |
| | | | | demonstrated compared to BSA. |
| HBD | [68], L. Wang et al.; [69], S. Das et al. | Convergence speed of BSA with the optimal solution | DE | Efficiency and effectiveness of HBD was enhanced |
| Fast Hybrid BSA-DE-SA | [70], M. Brévilliers et al. | Convergence speed and quality of BSA | DE with SA | Performance of Fast Hybrid BSA-DE-SA was improved. |
| IBSA | [71], Z. Su et al. | Overcoming drawbacks of discrete points when it was generated in the controls parameterisation | hp-adaptive Gauss pseudo-spectral methods (hpGPM) | It had better performance in terms of accuracy, convergent speed, and robustness. |
| BSABCM | [72], A. Askarzadeh, and L. dos Santos Coelho | Estimating fuel cell electrochemical model's unknown parameters to achieve accurate solutions | Burger's chaotic map (BSABCM) | It had a promising result in terms of performance. |
| BSANN | [73], S.K. Agarwal et al. | Classifying three mental tasks (hand movement generation of left or right, and imagination of word) | Artificial neural network (ANN) | It had better accuracy. |
| BSAF | [74], J.A. Ali et al. | Controlling three-phase induction motor (TIM) | Fuzzy logic controller | It had a better performance. |
| BSA-WM | [75], P.S. Pal et al. | Identifying types of nonlinear Hammerstein models | Wavelet-based mutation | It had better accuracy and precision. |
| BGBSA | [76], W. Zhao et al. | Enhancing convergence performance of BSA | Guided BSA | Better efficiency and effectiveness of BGBSA were guaranteed. |
| MFE-BSA | [77], S. Pare et al. | Addressing the issue of colour image multilevel thresholding | Modified fuzzy entropy (MFE) | It had a practical result about robustness, preciseness, and stability. |
| w-BSAFCM, | [78], G. Toz et al. | Achieving image-clustering method | Classic fuzzy C-means (FCM) | It was useful for solving the problem of image clustering |
| BSA-SCA | [79], O.E. Turgut | Designing rube and shell evaporator optimally | Since-cosine algorithm (SCA) | A better accuracy of BSA-SCA was provided. |
| BS-HH | [80], J. Kacprzyk, and W. Pedrycz | Dealing with the distributed assembly permutation flow-shop scheduling problem (DAPFSP) in supply chain management | Hyper-heuristic algorithm | It was more effective. |
| BS-HH | [81], J. Lin | Solving flexible job-shop scheduling problem with fuzzy processing time (FJSPF). | Hyper-heuristic algorithm | BS-HH outperformed it counterparts. |
| BSOA-SVM | [82], H. Ao et al. | Improving effectiveness and accuracy of BSA | Support vector machine (SVM) | low computation time with high accuracy of BSOA-SVM were provided |
| hDEBSA | [83], S. Nama, and A.K. Saha | Enhancing convergence speed and performance of the standard BSA | Control parameter (F) | Performance of hDEBSA was better than BSA and its competitive algorithms. |
| E-BSADE | [84], S. Nama et al. | Performance | Time-consuming parameters of both BSA and DE. | Performance of E-BSADE was better than other competitive algorithms. |
| BSA-SQP | [85], W.U. Khan et al. | Utilising it for active noise control systems (ANC) | Exploration searchability | BSA-SQP was accurate and able to minimise complex problems with fast convergence. |
| BSA-based MPPT algorithm | [86], M. Sriram, and K. Ravindra | Analysing characteristics of PV array | Maximum power point tracking (MPPT) algorithm | BSA-based MPPT algorithm was performed better compared to competitive algorithms (INC and ABC based MPPT) except P&O. |
| SRBSAMVS | [87], A. Gosain, and K. Sachdeva | Minimising MVS problem | Exploitation and exploration ability | SRBSAMVS was robust in terms of increasing user queries and processing cost of queries. |
| BSA-based LSSVM | [88], Z. Tian et al. | Predicting short-term wind speed | LSSVM | The introduced model was valuable for engineering applications. |
| BSA-based ANN model | [89], M. Konar | Maximising engine performance of unmanned aerial vehicles (UAV) and flight time | ANN | BSA-based ANN model was highly assisted the designers for designing UAV. |
| APSS&AMFS–BSA-ESN | [90], Z. Wang et al. | Finding out the most suitable output weights of ESN | Echo state networks | (i) The accuracy of the optimised ESNs was better than the basic ESN. (ii) Performance of APSS&AMFS BSA-ESN was relatively better than the basic ESN, the three optimised ESNs, and the other competitive approaches. |
| BSA-SVM | [91], V. Thai et al. | Diagnosing gear faults | SVM | BSA-SVM was more accurate than GA or PSO for diagnosing gear faults. |
| EEMD/VMD-HSBA-DAWNN | [92], S. Sun et al. | Forecasting wind speed modelling strategy | Feature selection, signal decomposition, and parameter optimisation, a new mixture of DAWNN method | It was effectual for multi-step WSF strategy. |
| BSAxGWO | [93], D. Jia et al. | Enhancing search capabilities | Grey wolf optimiser algorithm (GWO) | BSAxGWO was efficient in minimising global optimisation problems. |
| BSANCS | [94], Z. Xu et al. | Enhancing searchability | Negatively correlated search (NCS) | BSANCS was effective and feasible in |
| | | | | enhancing solution efficiency and exploration ability. |
| IHBSA | [95], S. Yan et al. | Using it for fuzzy clustering, and minimising cluster number and cluster centroids concurrently | T-S fuzzy model | IHBSA was accuracy and effectiveness. |
| LSSVM-based IHBSA | [96], C. Zhang et al. | Model-identification of pumped turbine governing system | Least-squares support vector machine (LSSVM) | LSSVM-based IHBSA was achieved higher accuracy, generalisation performance, and better robust in comparison to other counterpart models. |
| ELM-BSA. | [97], L. Chen et al. | Forecasting flood occurrence | Extreme learning machine (ELM) | ELM-BSA had better results compared to its competitive models |
| MOBBSA | [98], A. Pourdaryaei et al. | Forecasting electricity price | Multi-objective binary BSA | MOBBSA was more effective in forecasting electricity price compared to the simulation results of artificial neural network (ANN) and ANFIS. |
| IBSA | [99], J. Zhou et al. | Minimising parameter identification problem for pump-turbine governing system (PTGS) | Chaotic local, adaptive mutation scale factor, search operator, the elastic boundary processing strategy, and orthogonal initialisation technique | IBSA was performed better in comparison with GSA, PSO, and the standard BSA in terms of parameter identification accuracy and quality. |
| VMD-BSA-RELM | [100], J. Zhou et al. | Improving wind forecasting accuracy | Variational mode decomposition (VMD), and regularised extreme learning machine (RELM) | VMD-BSA-RELM was performed better than its both multi- and single-step forecasting by 50% |
| BSO-FCM | [101], W. Zhang et al. | Tackling high-order fuzzy-trend forecasting problem | Fuzzy c-means algorithm (FCM) | BSO-FCM showed remarkable forecasting accuracy than other benchmark approaches on the TAILEX. |
| SEBAL-BSA | [102], U.H. Atasever, and C. Ozkan | Adapting to Landsat 8 | Surface energy balance algorithm for land (SEBAL) | SEBAL-BSA was remarkably successful and demonstrated that it was a user-friendly method for water management related organisations. |
| BSA-based clustering algorithm | [103], G. Jothi et al. | Developing graphical user interface (GUI) for acute lymphoblastic leukaemia (ALL) image | Clustering algorithm | It had higher efficiency and more robust compared to PSO, ABC, and DE. |
| | | classification and segmentation | | |
| RBM-BSASA-BP | [104], H. Li et al. | Forecasting time-series problem | Restricted boltzmann machine (RBM), and simulated annealing | RBM-BSASA-BP was robust and reliable. So, it can be used for more wide-range of time series forecasting issues. |
| Hybrid BSA-SQP | [105], G. Mohy-ud-din | Minimising economic emission dispatch (EED) problem | Sequential quadratic programming (SQP) | Efficiency of Hybrid BSA-SQP concerning reducing emissions and operative cost was provided. |
| HBSDE | [106], K. Lenin et al | Minimising reactive power issue | DE | HDSDE had better performance in decreasing power loss. |
| BSA-NNRWs-N | [107], B. Wang et al. | Minimising hidden parameters of single-layer feed-forward | ANN with a combination of random weights (NNRWs) | BSA-NNRWs-N had a promising performance. |
| BSA-DE | [108], S. Mallick et al. | Optimally designing analogue CMOS amplifier circuits | DE | BSA-DE was promising in comparison to the results of DE, ABC, harmony search (HS), and PSO |
| BA-ACO | [109], B. Benhala et al. | Optimising analogue circuit performance | Ant colony optimisation (ACO) | Validity of BA-ACO were indicated. |

**3.2.1. HHBSA.** Owing to the limitations of BSA in quickly falling into local optimal, learning from an optimum individual, and adjustment of control parameter F, a new hybridised hierarchical BSA (HHBSA) was proposed to improve the efficiency and performance of the standard BSA [60]. HHBSA introduces two layers of the strategy of randomised regrouping and population such that the population diversity can be improved. Moreover, a strategy for mutation is employed to increase the ability of BSA. HHBSA was verified and tested by using benchmark functions with some optimisation problems in the real-world. The results indicated that HHBSA showed comparable performance to some existing EAs, and performed effectively in global optimisation searches. However, HHBSA when employed on large-scale and complex optimisation problems can lead to reduced search capability. To improve the performance of HHBSA, its searchability can be further studied by focusing on large-scale optimisation problems.

**3.2.2. MBSOA.** For tackling non-linear optimisation problems in economic load dispatch (ELD) in electrical generators, an improved BSA algorithm named memetic BSA (MBSOA) was proposed. MBSOA is a combination of the RWDE with the original BSA [61]. RWDE was used as a local search algorithm in the original BSA to speed up the convergence and improve the optimal solution in each iteration. In this case, MBSOA was as used for power sharing among the electrical generators units to satisfy the constraints of the generation limit for each unit and reduce the cost of electric power generation. The results of testing the proposed algorithm on six electrical generators revealed that MBSOA was a successful and effective in solving non-linear problems, especially the ELD problem.

**3.2.3. BPNN-BSA.** A recent study [62] presented BSA with the back-propagation neural network (BPNN) to improve the robustness and accuracy of the BPNN model by determining the best value of the learning rate and neurons of the hidden layers. This hybrid algorithm was used to estimate a lithium-ion battery's state of charge (SOC) for ensuring its safe operation and preventing it from over-discharging and -charging. The evaluation results showed the following: Firstly, the BPNN-BSA model was more successful in terms of performance compared with other neural networks in making high accuracy estimation of the SOC under various temperatures and EV profiles. Secondly, the performance of BPNN improved substantially by utilising BSA. Finally, and most importantly, the proposed hybrid algorithm could estimate the value of SOC for the lithium-ion battery in terms of improved accuracy and reduced time compared to the original BPNN. In another study [63], two heuristic techniques: PSO and BSA were hybridised with the BPNN for the estimation of an optimal state of energy (SOE) of a lithium-ion battery and to increase the accuracy of BPNN. The hybridised algorithms are called BPNN-PSO and BPNN-BSA. The results indicated that both the proposed hybrid algorithms showed satisfactory accuracy with an error rate of ±5%. The performance evaluation conducted at 25 °C showed that BPNN-BSA had superior performance than BPNN-PSO in decreasing both the root mean square error (RMSE) and MAE; the decrease was 2.8 and 4.4% respectively.

**3.2.4. HBSA(s).** A total of four different algorithms, all named HBSA have been proposed in the literature [64–67]. In the first proposition, a hybrid BSA (HBSA) on the basis of a new EA—backtracking search optimisation with combination of two local searches was proposed to (i) balance the local search (exploitation) and the global search (exploration), (ii) to solve discrete problems such as permutation flow-shop scheduling problem (PFSP) and decrease its makespan, and (iii) avoid falling into local optimal by using SA-based mechanism [64]. Discrete crossover and mutation strategies of the SA-based mechanism are utilised as an improvement to avoid falling into local optimal and speed up the local search methods. The results of this research indicated that despite its ability to solve continuous problems, HBSA showed feasibility for improvement and usefulness for discrete problems such as PFSP. However, HBSA still required some improvements to determine optimal solutions. Based on this, two aspects could be considered for future research work. Firstly, improving the effectiveness of the local search methods and making them as effective as other local search methods, e.g., the local search of NHE-h based method [110], the local searches of pairwise-based and SA-based with a combination of adaptive meta-Lamarckian learning method [111], and the local search of RLS as a referenced search [112]. Secondly, HBSA was only applied to the discrete problem in this study [64]; however, it has the likelihood to be applied to complex and constraint problems owing to its excellent performance of a standard version of BSA [22].

In the second proposition, a hybridised BSA was proposed to solve both energy consumption and makespan for PFSP [65]. This HBSA was combined with SA to renew the population, and a local search approach was applied consequently. To improve the performance of this HBSA, the operators of BSA were analysed. Finally, this algorithm was applied to some popular benchmark functions to evaluate its performance. The evaluation results indicated HBSA's effectiveness compared to GA and the branch and bound algorithm.

In the third proposition, a modified and adaptive BSA hybridised with Hooke-Jeeves pattern search was suggested for global and numerical optimisations [67]. This HBSA comprises two parts. The exploitation phase was performed by the modified pattern search method, whereas the control parameters of BSA were generated by a powerful strategy. The introduced algorithm was tested on the IEEE CEC2014 benchmark functions, and it was compared to the standard BSA. The results showcased the efficiency and effectiveness of HBSA compared with BSA.

In the fourth proposition, according to a recent study [66], the standard BSA was combined with quadratic approximation (QA) to introduce a new hybridised BSA algorithm called HBSA for minimising complex optimisation problems. QA improved the accuracy and exploitation searchability of BSA. This HBSA was tested on 25 functions to validate its performance. The numerical results were evaluated against five variants of PSO, and analysis was conducted for the amplitude parameter (F) of BSA in terms of sensitivity. The results indicated that HBSA was statistically successful compared with other competitive algorithms and exhibited good performance.

**3.2.5. HBD.** In this study [68], a hybrid BSA with DE was proposed. The proposed algorithm is called HBD. In HBD, DE was utilised with the BSA algorithm to optimise each other to improve the convergence of BSA with the optimal solution. The performance of HBD was verified by using it on some benchmark test functions; it was seen that the efficiency and effectiveness of HBD were superior compared to BSA and DE. Future research can be based on the following two directions. Firstly, additional experiments can be conducted on a variety of probabilistic models in HBD. Secondly, HBD can be employed in large-scale and real-world optimisation problems because results indicated that HBD was stable in terms of scalability. In another piece of work [68], a hybridised algorithm comprising BSA and DE called BSA-DE was proposed and applied to sidelobe suppression problems such as concentric circular antenna arrays (CCAA). For the purpose of verification, BSA-DE was experimented on some complex benchmark test functions before applying it to CCAA. The results showed that BSA-DE was effective and showed satisfactory performance on most of the considered test functions. Additionally, BSA-DE showed improved performance and effectiveness on either structural or pattern parameters or both of them for the design problem of CCAA.

**3.2.6. Fast Hybrid BSA-DE-SA.** A hybrid algorithm that combines BSA, DE, and SA to improve the convergence and quality of BSA was introduced in [70]. This algorithm is called fast hybrid BSA-DE-SA. The algorithm when experimented on some benchmark test functions indicated that it had superior performance than BSA and the other two hybridised algorithms [68] in terms of quality improvement and convergence performance. Fast hybrid BSA-DE-SA was implemented in the graphics processing unit (GPU) and some high-dimensional benchmark problems. The following improvements can be implemented in the future. Firstly, this approach can be compared to other EAs such as PSO and other benchmark problems. Secondly, based on introducing a user-defined parameter, this algorithm can be improved with a self-adaptive technique to reduce dependency on the user and achieve better results. More importantly, it can be compared to other existing large-scale optimisation algorithms.

**3.2.7. IBSA.** According to [71], a hybrid algorithm that is an integration of hp-adaptive Gauss pseudo-spectral methods (hpGPM) and improved BSA was improved in two stages to overcome the drawbacks of the discrete points. The first stage comprised improving the BSA to enhance its global search (exploration) ability and the speed of convergence. The second stage comprised constructing an initialisation generator using IBSA during the first stage of the hybrid search. Therefore, two problems with complex dynamic constraints (CDC) were experimented by employing this hybrid algorithm. Further, Monte Carlo simulations were conducted. The results indicated that IBSA showed improved performance in terms of accuracy, convergence speed, and robustness.

**3.2.8. BSABCM.** In another research [72], a combination of BSA with Burger's chaotic map (BSABCM) was introduced to estimate the fuel cell electrochemical model's unknown parameters to achieve accurate solutions. Experiments were conducted based on two systems, viz., the Ballard Mark V FC and the SR-12 Modular PEM generator. BSABCM showed promising results than those of the competing metaheuristic algorithms. For further studies the other newly proposed algorithms such as artificial cooperative search algorithm [22], differential search [113], and others [114] can be investigated.

**3.2.9. BSANN.** Neural network is one of the most popular techniques used for human cognitive function, wherein mental signals are changed into control signals. On this basis, [73] proposed neural network BSA (BSANN) for classifying three mental tasks (hand movement generation of left-side and right-side and imagination of words). BSA is suitable for deciphering non-differentiable and non-linear problems; and it retains the old memory population that is used to generate a new solution and obtain the benefit of using the previous population's search results. Experimental results of the proposed algorithms on some brain-computer interfaces (BCI) indicated that BSANN showed superior results with respect to accuracy compared to other algorithms for the mental task classification.

**3.2.10. BSAF.** In [74], BSA with the fuzzy logic controller was proposed to control the three-phase induction motor (TIM). The proposed algorithm (BSAF) did not use the trial-and-error procedure to obtain the membership functions (MFs). In the speed motor controller design, the generated MFs were employed for both output and input based on the results of the fitness evaluation function developed by BSA. The results of BSAF were compared with its competitors such as PSO and GSA to validate the developed controller by using the simulation tests in the Simulink environment or MATLAB. BSAF provided improved performance than the other two aforementioned algorithms in terms of the transient response on distinct mechanical speeds and loads and the damping capability. Speeding up the response regarding the steady-state error, overshoot, and time setting can enhance the performance of TIM. This notion can be a motivation for further studies.

**3.2.11. BSA-WM.** A hybrid approach was presented in [75] to identify the types of non-linear Hammerstein models using BSA with wavelet-based mutation (BSA-WM). The hybridised algorithm was experimented on some benchmark test functions to examine its accuracy and precision. The results showed superior accuracy and precision of BSA-WWM compared to other heuristic algorithms.

**3.2.12. BGBSA.** To enhance the convergence performance of BSA, a guided BSA (BGBSA) was introduced in [76]. In the first stage of BGBSA's iteration, the historical information was used, and the best individual was employed on the iterations at a later stage. The efficiency and effectiveness of BGBSA were evaluated based on 28 benchmark functions. The experimental result showed improvements in both the efficiency and effectiveness of BGBSA.

**3.2.13. MFE-BSA.** Modified fuzzy entropy (MFE) with backtracking search optimisation was used to address the issue of colour image multilevel thresholding [77]. This technique is named MFE-BSA. Experimental results indicated that MFE-BSA showed effective result in terms of robustness, preciseness, and stability.

**3.2.14. w-BSAFCM.** In [78], an image clustering method based on a classic fuzzy c-means (FCM) algorithm combined with BSA was proposed. The new algorithm is called w-BSAFCM. In this algorithm, the objective function of BSA with FCM was minimised to obtain the image-clustering method. Additionally, for improving the exploration (local search) of the algorithm, a parameter (w) was introduced for BSA. This parameter (w) was used to determine the direction of the search matrix of BSA. The effectiveness of this new algorithm was evaluated on three benchmark functions. The evaluation results indicated that w-BSAFCM was effective to solve the problem of image clustering.

**3.2.15. BSA-SCA.** A research article [79] discusses combining BSA with the since-cosine algorithm (SCA) as the first application of meta-heuristic algorithms to design an optimal rube and shell evaporator. This new algorithm is called BSA-SCA. Accuracy of this hybrid algorithm was evaluated by using 10 benchmark functions. BSA-SCA when compared with six optimiser algorithms showed superior results than the other six techniques. Furthermore, future application of BSA-SCA to heat sink thermal design was suggested.

**3.2.16. BS-HH.** There were two hybridised algorithms named BS-HH in the literature, named BS-HH [80,81]. Firstly, in [80], the backtracking search-hyper heuristic algorithm was introduced to deal with DAPFSP in supply chain management. The proposed algorithm comprised some effective rules to create low-level strategies (LLSs), and backtracking search optimisation acted as high-level heuristics to employ the LLSs to work on space solution. Additionally, this solution was proposed to construct a feasible schedule. The effectiveness of BS-HH was evaluated for its effectiveness on two benchmark test functions. The results showed that BS-HH was more effective than its competitors. Further, it was suggested that BS-HH could be improved by using some faintness analysis techniques, and used to tackle PSFP for the distributed assembly permutation manufacturing system and supply chain management with distinctive assembly factories. Secondly, another hybridised BSA algorithm was suggested by [81] to solve the flexible job-shop scheduling problem with fuzzy processing time (FJSPF). The suggested scheme is called BS-HH. BSA was used to heuristically handle the low-level strategy in the solution domain. Finally, the performance of the proposed algorithm was evaluated on two well-known benchmark problems. The results indicated that BS-HH outperformed the results of its counterpart algorithms in the literature for minimising FJSPF.

**3.2.17. BSOA-SVM.** It is claimed that support vector machine (SVM) parameters can affect the classification results in terms of the accuracy rate. These parameters can be adjusted to improve the effectiveness and accuracy of the algorithm. Furthermore, [82] introduced a method by changing the SVM parameters based on backtracking search optimisation algorithm (BSOA in this study). The proposed method is called BSOA-SVM, and it was experimented on some benchmark problems. The results indicated that BSOA-SVM showed reduced computation time with high accuracy compared to CMAES-SVM, GA-SVM, and PSO-SVM. Additionally, BSOA-SVM was applied to real-world problems in combination with local mean decomposition-singular value decomposition (LMS-SVD) to diagnose roller bearing faults. The results showed that this combination of LMS-SVD with BSOA-SVM effectively reduced the cost, time, and errors.

**3.2.18. hDEBSA.** In [83], BSA was hybridised with DE to simultaneously update the value of the adaptive control parameter (F) of both the algorithms during the optimisation process. The aim of the proposed algorithm (hDEBSA) was to enhance the convergence speed and performance of the standard BSA. The proposed algorithm was evaluated and run on 28 benchmark functions of CEC2013 and 4 real-life problems against its competitor algorithms. The results showed that hDEBSA's performance was better than BSA and its competitive algorithms.

**3.2.19. E-BSADE.** Referring to [84], The standard BSA was ensembled with DE to create a new algorithm called E-BSADE. E-BSADE aimed to achieve optimum performance by tuning the time-consuming parameters of both BSA and DE. The performance of E-BSADE was validated for minimising several benchmark functions in comparison with the standard BSA, DE, and traditional DE mutation strategy. Further, the performance results were evaluated against the variants of PSO. The results suggested that the performance of E-BSADE was superior compared to other competitive algorithms.

**3.2.20. BSA-SQP.** In the latest study [85], the strength of the standard BSA and sequential quadratic programming (SQP) was utilised for active noise control systems (ANC). The newly proposed algorithm is called BSA-SQP. Exploration searchability of BSA in combination with the rapid local refinements and SQP were exercised for robust, powerful, reliable optimisation of non-linear ANC systems. The performance of BSA-SQP was evaluated on ANC. The results indicated that BSA-SQP was accurate, and it was able to minimise complex problems with fast convergence.

**3.2.21. BSA-based MPPT algorithm.** According to the recent study [86], the standard BSA was coupled with the maximum power point tracking (MPPT) algorithm to analyse the characteristics of the PV array under distinct configurations for non-uniform and uniform irradiance conditions. BSA-based MPPT was evaluated against perturb & observe (P&O)-, incremental conductance (INC)-, and ABC-based MPPT to analyse the I-V and P-V characteristics of different solar PV array configurations. BSA-based MPPT showed improved performance compared to other competitive algorithms (INC- and ABC-based MPPT) except the P&O-based MPPT.

**3.2.22. SRBSAMVS.** The researchers of [87] proposed a new algorithm, in which BSA is blended with stochastic ranking (SR) to minimise the MVS problems. This newly introduced algorithm is called SRBSAMVS. The motive behind proposing this method was the exploitation and exploration ability of BSA with the capability of SR to deal with constrained problems. SRBSAMVS was evaluated based on the TPC-H star schema benchmark datasets. The evaluation results of SRBSAMVS when compared with those of a constrained EA proposed by [115] indicated that SRBSAMVS was robust in terms of handling increased user queries and the processing cost of queries.

**3.2.23. BSA-based LSSVM.** The authors of [88] proposed a standard BSA based on the least squares support vector machine (LSSVM) called BSA-based LSSVM to predict short-term wind speed. In the proposed algorithm, BSA was used to optimise the parameters that influence the regression model of LSSVM, whereas LSSVM was utilised for predicting short-term wind speed. The results indicated that BSA-based LSSVM showed high reliability and accuracy in the prediction of short-term wind speed. Meanwhile, the performance indicators of prediction were enhanced. The introduced model was considered suitable for engineering applications.

**3.2.24. BSA-based ANN model.** As stated in [89], a model using BSA based on artificial neural network (ANN) for maximising the engine performance of unmanned aerial vehicles (UAV) and flight time was proposed. In the suggested model, BSA was utilised to optimise the structure of ANN to achieve the parameters required to maximise the engine performance and flight time. The simulated results showed that the BSA-based ANN model significantly assisted the designers in designing the UAV.

**3.2.25. APSS&AMFS–BSA-ESN.** As proposed in [90], the variants of BSA based on the echo state network (ESN) were utilised for identifying the most suitable output weights of ESN because it was considered as a complex optimisation problem. The suggested method, APSS&AMFS–BSA-ESN, is used to assist the prevention of overfitting in ESNs. Concurrently, optimised ESNs were introduced for effective time series forecasting. Experiments were conducted to evaluate the basic ESN with and without optimisation, APSS&AMFS–BSA-ESN, and other counterpart algorithms. The results indicated the following. Firstly, the optimised ESNs showed superior accuracy than the basic ESN. Secondly, APSS&AMFS–BSA-ESN showed relatively improved performance than the basic ESN, the three optimised ESNs, and the other competitive approaches.

**3.2.26. BSA-SVM.** The accuracy precision of SVM depends on the selection of its optimal parameters [91]. In the same study, a method to diagnose gear faults was proposed. The introduced algorithm, BSA-SVM, was based on BSA using the SVM. The accuracy of the introduced method was validated by comparing it with PSO and GA. Results indicated that BSA-SVM is more accurate than GA and PSO in diagnosing gear faults.

**3.2.27. EEMD/VMD-HBSA-DAWNN.** Considering the feature selection, signal decomposition, and parameter optimisation, a variant of the DAWNN method tuned by the hybrid BSA and merged with two-state decomposition EEMD/VMD was suggested for the modelling strategy to forecast wind speed [92]. The proposed method is called EEMD/VMD-HBSA-DAWNN. The effectiveness of this method was evaluated through its application for the WSF strategy, and the results indicated that the algorithm was effective for the multi-step WSF strategy.

**3.2.28. BSAxGWO.** Some researchers [93] stated that the combination of BSA with the grey wolf optimiser (GWO) algorithm enhances the search capabilities. The suggested combination model is called BSQxGWO. Experiments were conducted to validate the claims of the researchers. The simulated experimental results on the CEC2017 benchmark functions demonstrated BSAxGWO's efficiency in minimising global optimisation problems.

**3.2.29. BSANCS.** In [94], BSA was combined with the negatively correlated search (NCS) to enhance its searchability. The new algorithm called BSANCS was run on the CEC2017 benchmark functions to measure its validity. The results showed that BSANCS was effective and feasible in enhancing the solution efficiency and exploration ability. The authors of the study suggested applying BSANCS on combinatorial optimisation problems [116,117], multi-objective optimisation [118,119], and neural network learning tasks [120,121].

**3.2.30. IHBSA.** The researchers of [95] proposed a unique improved and hybrid BSA for fuzzy clustering and to minimise the cluster number and cluster centroids concurrently. The suggested method, IHBSA, is based on the T-S fuzzy model and is used in hydroelectric generating units. The generality and feasibility of IHBSA were evaluated on the essential benchmark functions, and the results were compared with those of the other counterpart algorithms in the literature. Furthermore, IHBSA was evaluated on the practical data of a hydraulic turbine, and the results indicated its accuracy and effectiveness. Further works will comprise designing a T-S fuzzy model suitable for industrial control systems. Later on, the LSSVM method was proposed on the basis of IHBSA, and it was used for model identification of a pumped turbine governing system [96]. The model was evaluated on two benchmark functions and an application for the pumped turbine governing system. The results indicated that the LSSVM-based IHBSA model not only achieved higher accuracy but also provided higher generalisation performance and improved robustness in comparison to other counterpart models.

**3.2.31. ELM-BSA.** Extreme learning machine (ELM) is a powerful method for flood forecasting. In some cases, ELM has a local minimum problem owing to the randomisation process of its input weights. Thereby, an enhanced ELM method was proposed based on the dual-population of BSA [97]. The suggested model called ELM-BSA was applied to a river in a case study to examine its efficiency. Compared to other counterpart models such as ELM and general regression neural network (GRNN), ELM-BSA provided improved results compared to its competitive models. The proposed method was suggested as an alternative for flood forecasting.

**3.2.32. MOBBSA.** Recently, a multi-objective feature selection technique that comprises the multi-objective binary BSA (MOBBSA) as a competent EA and an adaptive neuro-fuzzy inference system (ANFIS) approach for forecasting short-term electricity price were proposed [98]. The suggested hybrid method is called MOBBSA. Real-world datasets from electricity sales sector were used in a case study to evaluate the performance of MOBBSA. The simulated results showed that MOBBSA was more effective in forecasting the electricity price compared to ANN and ANFIS.

**3.2.33. IBSA.** In [99], BSA was combined with the chaotic local, adaptive mutation scale factor, search operator, elastic boundary processing strategy, and orthogonal initialisation technique to minimise the parameter identification problem for the pump turbine governing system (PTGS). The suggested algorithm, IBSA, was applied to an illustrative example to evaluate its efficiency and accuracy. The simulated results showed that IBSA provided superior performance compared to GSA, PSO, and the standard BSA in terms of accuracy and quality of the parameter identification.

**3.2.34. VMD-BSA-RELM.** In another study [100], researchers proposed a new optimisation algorithm by combining BSA, variational mode decomposition (VMD), and regularized extreme learning machine (RELM) to improve accuracy of wind forecast. The proposed model, VMD-BSA-RELM, firstly, the time series of the wind speed was observed, and it was further decomposed by VMD into a number of partial sequences. Secondly, BSA was used to identify the optimum parameters of RELM. Subsequently, the trained RELM was built to forecast the multi-step wind speed. To evaluate VMD-BSA-RELM, experiments were conducted on several benchmark functions. The results showed that the proposed approach showed improved performance than both multi- and single-step forecasting by 50%.

**3.2.35. BSO-FCM.** As suggested by [101], the standard BSA was combined with FCM to tackle the problem of high-order fuzzy-trend forecasting. The suggested model, BSO-FCM, aims to improve the capability of dividing the datasets into intervals by combining the global optimum capability of BSA to release the FCM from local minimum or local maximum. Meanwhile, an improved kidney-inspired approach was utilised to unify the high-order forecasting values. The approach showed remarkable forecasting accuracy than other benchmark approaches on the TAILEX.

**3.2.36. SEBAL-BSA.** As articulated in [102], a novel surface energy balance algorithm for land (SEBAL) based on BSA was proposed. This algorithm can use Landsat 8 images as datasets to automatically select the wet-hot pixels with ground control points. The proposed approach, SEBAL-BSA, was evaluated on Landsat 8 images in Turkey. The results showed that SEBAL-BSA was not only remarkably successful but also a user-friendly method suitable for use in water management related organisations.

**3.2.37. BSA-based clustering algorithm.** As reported in [103], a BSA-based clustering algorithm was proposed to develop a graphical user interface (GUI) for acute lymphoblastic leukaemia (ALL) image classification and segmentation. The evaluation results of this algorithm showed its improved efficiency and robustness compared to PSO, ABC, and DE.

**3.2.38. RBM-BSASA-BP.** In this article [104], a hybrid approach called RBM-BSASA-BP was proposed to forecast the time series. This hybrid method comprises the restricted Boltzmann machine (RBM) to provide the parameters for BPNN. Further, BSA with SA was used to tune the parameters to find the optimal biases and weights. In summary, RBM-BSASA-BP leveraged the advantages of both BSASA and RBM methods. It was concluded that the suggested method being robust and reliable, can be used for a wider-range of time series forecasting issues.

**3.2.39. Hybrid BSA-SQP.** A hybrid BSA with SQP was exploited for optimising the emissions and operative cost in transferring power to the generation units liable to operative constraints [105]. The proposed approach, hybrid BSQ-SQP, was applied to solve the dynamic EED problem on five- and ten-unit test systems, HDEED problem, six-unit and IEEE 57 bus, seven-unit test system, with and without renewable generation. The results demonstrated the efficiency of the algorithm in terms of reducing emissions and operative cost.

**3.2.40. HBSDE.** The performance of BSA requires further improvements. For this reason, a new method based on hybridising BSA with DE for minimising reactive power issue was developed [106]. In the proposed method, HBSDE, an exploitative strategy in DE was used to speed-up the convergence speed. HBSDE was evaluated on the standard IEEE 30 bus system. The simulation results indicated improved performance of HBSDE in decreasing power loss. Furthermore, the control variables achieved in HBSDE were within acceptable boundaries.

**3.2.41. BSA-NNRWs-N.** BSA was hybridised by using a neural network with a combination of random weights (NNRWs) to create a new hybridised algorithm (BSA-NNRWs-N) to minimise the hidden parameters of the single-layer feed-forward network [107]. Results of the experiments on the regression and classification datasets demonstrated that BSA-NNRWs-N showed promising performance.

**3.2.42. BSA-DE.** In [108], BSA was combined with DE to design two well-known analogue CMOS amplifier circuits optimally. The proposed algorithm, BSA-DE, was tested on the two analogue circuits to examine its effectiveness. The results indicated that BSA-DE showed promising effectiveness compared to DE, ABC, harmony search (HS), and PSO with respect to the design specifications of the parameters, practical design of the analogue circuits, and convergence speed.

**3.2.43. BA-ACO.** In another study [109], the standard BSA was combined with the ant colony optimisation (ACO) to optimise the analogue circuit performance. The role of the new hybridised algorithm, BA-ACO, was to reduce the computation time. The validity of BA-ACO was proved after testing it on optimising a radio-frequency circuit design.

### 3.3. Extensions of BSA

The first success of the BSA for numerical optimisation problems has strongly urged scholars to extend it to other areas. These areas are multi-objective optimisation, constrained optimisation, numerical optimisation, and binary optimisation. These extensions of BSA are briefed in Table 4.

**Table 4: The extended algorithms based on BSA**

| Algorithm name | Problem type | Source, author(s) | Aim of the extension | Result |
|---|---|---|---|---|
| BBSA | Binary | [122], M.G.M. Abdolrasol et al. | Finding the best scheduling controller for Microgrid Integration and virtual power plant | BBSA was able to reduce the cost of the power generation, minimise loss of power, deliver power in high-quality. It was also reliable to the loads, and had priority-based sustainable MGs into the grid |
| BBSA | Binary | [36], M.S. Ahmed et al. | Optimally solving schedule controller for managing home energy systems | It had better performance. |
| BSA | Numerical | [123], H. Zhao et al. | Minimal cost feature between misclassification costs and test costs by considering numerical data with measuring errors | BSA was efficient. |
| BSA-HA | Numerical | [124], H. Zhao et al. | Solving numeric data feature selection problem with measurement errors | It was efficient and effective. |
| BSA | Numerical | [125], P. Civicioglu | Solving circular antenna array design problem | Performance of S-DSA was better than BSA. |
| BSA | Numerical | [11], P. Civicioglu | Tackling numerical optimisation problem | Better performance and effectiveness of BSA were provided. |
| BS | Binary, numerical | [126], A.M. SHAHEEN, and R.A. EL-SEHIEMY | Optimally choosing types, routes, and number of circuits | BS had more capability with satisfactory technical and economic benefits. |
| BSAISA | Constrained | [35], H. Wang | Dealing with constrained problems in a variety of fields | It was powerful in terms of convergence speed. |
| BSA-SA$\varepsilon$ | Constrained | [127], C. Zhang et al | Minimising contained optimisation problems, and three constraint handling approaches | BSA-SAε had an excellent performance in comparison to SR, RPGA, a Simplex, MABC, and AMA. |
| MODBSA/D | Multi-objective | [128], F. Zou et al. | Detecting community in large and complex networks | It was effective, and its performance was promising. |
| MODBSA/D | Multi-objective | [128], F. Zou et al. | Dealing with energy efficiency and permutation scheduling with controllable and manageable transportation problem | It had better performance. |
| MOBSA | Multi-objective | [129], M. Modiri-Delshad, and N.A. Rahim | Coping with emission and ELD of power system generators | It had better performance in terms of robustness and efficiency for multi-objective problems. |
| Enhanced BSA | Multi-objective | [130], K. Bhattacharjee et al. | Dealing with emission and economic load dispatch of power system generators | BSAMO was capable of reaching an optimal solution with a short duration of time. |
| BSAMO | Multi-objective | [131], R. El Maani et al. | Solving multi-objective optimisation problem | BSAMO was considered as a possible alternative for minimising multi-objective optimisation problems. |
| MOBBSA | Binary, multi-objective | [98], A. Pourdaryaei et al. | Forecasting electricity price | Superiority of MOBBSA was provided compared to ANN and ANFIS was demonstrated. |
| BSAMO | Multi-objective | [132], A.T. Zeine et al | Optimising multi-objective problem | BSAMO was produced a better convergence compare to the results of NSGA-II. |
| MODBSA | Multi-objective | [133], C. Lu et al. | Optimising scheduling problem with controllable processing times (CPT) | MODBSA was helpful for such a schedulable program. |
| MOBSA | Constrained, multi-objective | [134], F. Daqaq et al. | Minimising high constrained, and multi-objective optimisation problem | MOBSA was capable of minimising OPF problems in large-scale power systems with thousands number of buses. |
| MOBSA | Multi-objective | [135], C. Lu et al. | Optimally solving power flow | MOBSA was outperformed the other competitive algorithms. |

**3.3.1. Binary Optimisation.** BSA is one of the algorithms that can be used for solving binary optimisation problems in different fields. For example, BBSA was used in two different fields. Firstly, BBSA was used to find the best scheduling controller for microgrid integration and virtual power plant [122]. BBSA shows promising results to solve the binary problem because it reduces the cost of power generation, minimises the loss of power, delivers high-quality and reliable power to the loads, and has priority-based sustainable micro-grids in the grid. In another piece of research, BBSA was used to solve the optimal scheduling and control problem to manage home energy systems [36]. The BBSA scheduling controller showed improved performance compared to the scheduling controller of BPSO to minimise energy use, electricity bill, and energy expenditure.

**3.3.2. Numerical Optimisation.** BSA has been used extensively to dealing with numerical optimisation problems. For instance, in this [123], a backtracking search optimisation was proposed for minimal cost feature between misclassification and test costs by considering numerical data with measuring errors. Results demonstrated that the proposed algorithm was efficient, for approximately one thousand datasets. This work can be improved in the future for a larger object dataset. Another study [124] was about solving the numeric data feature selection problem with measurement errors by using backtracking and heuristic algorithms. The algorithm proposed in this study is called BSA-HA. The evaluation showed that the pruning techniques of the backtracking algorithm were efficient, and the heuristic algorithm was effective. This research can be applied to real-world applications such as cost-sensitive learning. In another study [125], three different numerical optimisation problems were solved by using some EAs such as BSA and S-DSA. The aim was to solve the circular antenna array design problem. The results indicated that S-DSA outperformed the other EAs including BSA in solving the aforementioned problem. Similarly, [11] proposed an EA named BSA to tackle numerical optimisation problems. The performance of the new BSA was compared to some other heuristic algorithms such as CMAES, PSO, ABC, CLPSO, SADE, and JDE for solving numerical optimisation problems. Results indicated that the new BSA was the most successful of all the compared algorithms in terms of performance and effectiveness. Hence, success of BSA in terms of effectiveness and performance for numerical optimisation problems depends on the type of the numerical problem. Another recent research proposed a binary and integer coded BSA to minimise the transmission network expansion planning (TNEP) problem while considering the security limitation [126]. The aim of the newly introduced algorithm was to choose the types, routes, and number of circuits optimally to achieve the forecasted future load by considering the planning and operational constraints. The technique using this algorithm was applied to two realistic networks in Egypt. The results of these two networks indicated that the proposed technique had more capability with satisfactory technical and economic benefits in solving the two problems than other counterpart heuristic techniques, multi-verse optimiser (MVO), and the integer-based PSO (IBPSO) technique.

**3.3.3. Constrained Optimisation.** Recently, several solutions have been proposed to solve the constrained optimisation problems. BSA was used in a variety of fields to deal with the constrained problems. For example, a modified BSA based on SA was used to solve the constrained optimisation problems in the engineering field [35]. Meanwhile, the authors of [127] claimed that that they applied BSA to minimise the contained optimisation problems for the first time. They adopted three constraint handling approaches to BSA, viz., a constrained method, a proposed e-constrained method with self-adaptive control was of ε value (BSA-SAε), and feasibility and dominance (FAD) rules. The numerical results indicated that BSA-SAε provided excellent performance in comparison to SR, RPGA, Simplex, MABC, and AMA. Additionally, BSA-SAε outperformed its counterpart algorithms in some well-known engineering problems.

**3.3.4. Multi-objective Optimization.** Several solutions have been introduced by scholars for the multi-objective optimisation problems. For example, a novel multi-objective approach based on discrete BSA was proposed to detect community in large and complex networks [128]. The suggested approach is called MODBSA/D. In this approach, the individual updating rules were redesigned on the basis of network topology. This approach was used to reduce two objective functions with respect to the ratio cut (RC) of the community detection problem and negative ratio association (NRA). The proposed method was experimented on real-world networks to test its performance. The results indicated that MODBSA/D was effective, and its performance was promising in coping with the community detection in complex networks. Meanwhile, this algorithm was used to deal with the problem of energy efficiency and permutation scheduling with controllable and manageable transportation [128]. This modification was one of the effective energy-saving strategies. Results of the conducted experiments indicated that the algorithm performed satisfactorily compared to the other popular multi-objective EAs to solve the aforementioned type of problem. Similarly, multi-objective BSA (MOBSA) was used for EED by considering the valve-point effects and the transmission network loss [129]. In the proposed multi-objective approach, the non-dominated approach (NDA) and the weighted sum method (WSM) were used to solve the LED problem efficiently. Furthermore, according to [130], an enhanced BSA to handle the problem of emission and ELD of power system generators. In the same study, the enhanced BSA was used to design a solution by using the effect of two new mutation and crossover operations. Owing to the capability of ruthless exploitation and exploration, the enhanced BSA can deal with non-linear multi-objective problems. The performance of the enhanced BSA in terms of robustness and efficiency for multi-objective problems was evaluated and compared with its competitors. Results indicated that the enhanced BSA was capable of reaching an optimal solution with a short duration of time. In the future, the enhanced BSA's capacity can be used to tackle several complex and non-linear problems. Lately, a study [131] proposed a new BSA for solving multi-objective optimisation problems. The proposed algorithm, BSAMO, was successful in minimising two types of problems. It was first applied to benchmark functions and then to two real-world fluid-structure interaction (FSI) problems. The results indicated that BSAMO was more competitive and successful in coping with complex problems. Therefore, BSAMO is considered as a possible alternative to minimising multi-objective optimisation problems. Concurrently, another MOBBSA was used to forecast the price of electricity [98]. This algorithm was applied to real-world electricity sales datasets. The results indicated the superiority of MOBBSA compared to ANN and ANFIS.

Next, a new BSA was proposed to optimise multi-objective problems [132]. The proposed algorithm named BSAMO was run on some benchmark problems and two structural engineering design applications to evaluate its performance. The simulated results revealed that BSAMO produced improved convergence compared to the results of NSGA-II.

Further, another study proposed a multi-objective discrete BSA (MODBSA) for large scale and multi-objective problems [133]. MODBSA was used to optimise the scheduling problem with controllable processing times (CPT) as a multi-objective and large-scale problem, particularly, for minimising the total compression cost and tardiness of CPT concurrently. For assessing MODBSA, it was compared with NSGA-II, PAES, and SPEA2. The results demonstrated that MODBSA was suitable for such a schedulable problem.

In another research [134], an algorithm was proposed to minimise high constrained and multi-objective optimisation problems. In the proposed algorithm, MOBSA, BSA was utilised as the primary optimiser for optimally adjusting the control variables of the power system in the OPF problem subject to objectives of voltage stability enhancement, voltage improvement, and fuel cost minimisation. The suggested multi-objective optimiser was tested on the standard IEEE 30 bus system. The optimiser was capable of minimising the OPF problems in large-scale power systems with thousands number of buses

Finally, another MOBSA was proposed for the energy-efficient multi-pass turning operation problem [135]. The proposed algorithm was tested and compared with MOPSO, NSGA-II, MOHS, and cMOEA/D for various parameter optimisation problems. The results showed MOBSA outperformed the other competing algorithms. The authors of the same study suggested investigating the expansion of MOBSA into other machine processes such as grinding and milling in the future.

## 4. Implementations of BSA

Backtracking search optimisation Algorithm is one of the most used algorithms for dealing with the optimisation problems of artificial intelligence. In addition, BSA has many usages in a variety of industrial fields. The main implementation of BSA can be classified into two parts: the common problems solved by BSA; and real-world applications of BSA. These two parts are detailed in this section.

### 4.1. Common problems solved by BSA

The common problems solved by BSA were: basic problems of BSA; travelling salesperson problem (TSP); scheduling program; and BSA toolkit. This section gives details of the common problems solved by BSA.

**4.1.1. Basic problems solved by BSA.** Backtracking is an interesting approach to solve fundamental problems such as solving games and puzzles and creating computerised opponents in mind games such as chess. Table 5 summarises the basic problems that can be solved by BSA.

**Table 5: Basic problems solved by BSA**

| Problem name | Source, author(s) | Algorithm | Aim of the algorithm | Result |
|---|---|---|---|---|
| N- Queens Problem | [13], S. Güldal et al.; [136], M.Z. bin Mohd Zain et al. | BSA after removing the threatened cells | Solving an n-queen problem using backtracking algorithm after removing the threatened cells | Decreasing the number of trial and error steps was provided. |
| Knight's tours puzzle | [12], D. Ghosh et al. | Simple BSA | Proposing a simple algorithm to design night’s tours puzzle for 8X8 square chessboard | It was easy to implement, and it performed well compared to the existing ones. |
| Random word generator for Ambrosia Game | [14], I. Kuswardayan, and N. Suciati | BSA | Designing and implementing random word generator for Gameplay in Ambrosia Game | Extracting word solution and the gameplay attracted users to play ambrosia again successfully |
| Maze Generation | [16], B. Zhichao | New BSA | Solving the shortest path of the maze problem | The New BSA was secure, and it took less time and space complexity for the maze problem. |

**4.1.2 Traveling salesperson problem.** One of the well-known problems in computer science and mathematics is travelling salesperson problem (TSP) [137]. This problem has attracted several scholars in both practical computational and theoretical aspects. One of the highly performing methods to deal with TSP is BSA [17,18]. Using BSA had two-fold implication: firstly, it reduced the number of additions. Secondly, it showed improvements in factorial running time. In the future, some other aspects of TSP such as writing TSP software and solving 25-city TSP by using a backtracking algorithm can be investigated.

**4.1.3. Scheduling problem.** Scheduling deals with resource allocation over time to perform a set of tasks. The scheduling problem is seen in fields such as health care and manufacturing. Ideally, these problems are considered as constrained optimisation problems (COPs). Generally, to solve CSPs and COPs, BSA can be successfully employed [138]. On this basis, several types of research works have been conducted on the scheduling problem, especially the multi-objective scheduling problem. These two studies [19,20] were proposed on multi-objective scheduling problem. Their proposed algorithms based on backtracking search outperformed other heuristic algorithms for the backtrack beam search and PFSP, respectively in terms of energy-efficiency. Additionally, in [139], the use of BSA and GA were investigated for the personal scheduling problem to assign tasks to the employees of a laboratory. with the results of the performance evaluation of BSA based on the experiments on the personnel scheduling problem indicated that BSA showed improved performance than that of the previous use of BSA in the literature.

**4.1.4. BSA toolkit.** According to this study, [26], a toolkit was developed for BSA in the LabVIEW environment. This toolkit comprises a GUI for programming that is used to control applications and design their measurements. The developed toolkit was experimented on a variety of benchmark functions, and these functions were tested on the DE toolkit for a comparative study. The evaluation shows that the results of the BSA toolkit were more effective than those of the DE toolkit. Fig. 5 presents a high-level design of the BSA toolkit. The toolkit has two inputs: BSA parameters and stopping criteria, and it has three outputs: global minimiser, global minimum, and fitness v/s iteration graph.

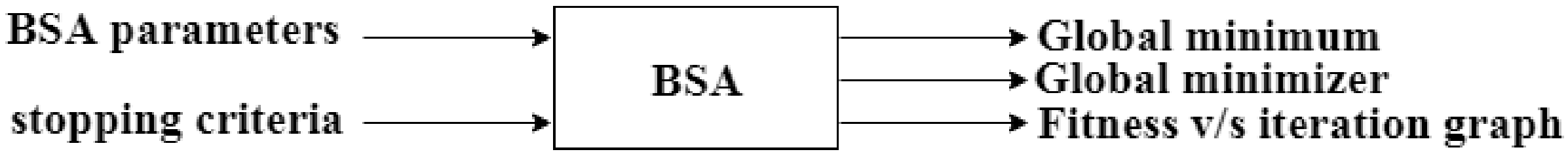

**Fig. 5: BSA Toolkit**

### 4.2. Real-world applications of BSA

BSA has many usages in a variety of industrial fields. Recently, BSA has been mainly used in five primary cohorts, which are control and robotics engineering, electrical and mechanical engineering, artificial intelligence, information and communication engineering, and material and civil engineering. Each of these applications of BSA is described below.

**4.2.1. Control and robotics engineering.** One of the leading and productive uses of BSA was in the field of control and robotics engineering. Table 6 summaries the primary uses of BSA in the mentioned fields.

**Table 6: Summary of BSA applications in control and robotics engineering**

| Source, author(s) | Algorithm | Aim of the algorithm | Result |
|---|---|---|---|
| [140], V. Kumar et al. | BSA | Tuning the parameters in developing a self-regulatory fractional-order fuzzy PID (SRFOFPID) | Superiority of SRFOFPID was demonstrated in the result compared to its counterpart approaches. |
| [47], M.A. Ahandani et al. | SBSA | Enhancing performance of the BSA | SBSA was robust and effective. |
| | | for identifying parameters of chaotic systems. | |
| [141], D. Guha et al. | BSA | Solving load frequency control (LFC) in power systems engineering | The proposed controller was adjusted by BSA was effective and robust; it had better performance compared with available algorithms in the literature. It can also be employed on nonlinear systems. |
| [142], H. Boudjefdjouf et al. | BSA | Diagnosing multi-wiring networks | BSA's efficiency was shown. |
| [49], J.A. Ali et al. | BSA-IFOC | Improving IFOC for operating TIM | BSA-IFOC was more robust than PSO-IFOC. |
| [143], A.O. de Sá et al. | BSA | Solving localisation problem | BSA had a better performance than the GA in terms of efficiency. |
| [144], A. Mehmood et al. | BSA | Estimating the parameters of nonlinear input systems based on the structure of Hammerstein control autoregressive (NHCARAR) | BSA was performing better on NHCARAR compared to its counterpart algorithms (GA and DE). |
| [91], Z. Wang et al. | BSA-SVM | Diagnosing the health status of gear faults. | BSA-SVM was more accurate than GA or PSO for gear fault diagnosis. |
| [42], J. Lu, and J. Ding | MBSA | Optimising the output weights of deep stochastic configuration networks (DSCN) for prediction intervals (PIS) constructions | MBSA was able to construct PIs with good quality. |
| [145], V. Goyal et al. | BSA | Turning fractional-order parallel control structure (FOPCS) | It had better performance compared to all other controllers. |
| [146], T.X. Dinh et al. | BSA-based inverse model | Identifying nonlinear dynamic system | The effectiveness of the model was promising compared to the PIDIM controller and conventional PID controller in hysteresis effects and dealing with nonlinearities of the system. |
| [147], V. Goyal et al. | BSA | Further tuning the parameters of control structures of fractional order parallel control structure (FOPCS) | The introduced FOPCS was capable of dealing with the issues of a pneumatic control valve in the flow control loop. |

**4.2.2. Electrical and mechanical engineering.** BSA was vastly used to deal with the problems in the field of electrical and mechanical engineering. Table 7 summarises the studies of BSA, which was used to solve electrical and mechanical engineering problems.

**Table 7: Summary of BSA applications in electrical and mechanical engineering**

| Source, author(s) | Algorithm | Aim of the algorithm | Result |
|---|---|---|---|
| [139], J. Yadav et al. | BSA-based approach with fuzzy expert rules | Optimally allocating multi-type distributed electric generators | It was capable to reduce power losses and voltage deviations substantially. |
| [148], A. Mehmood et al. | BSA | Estimating parameter for electrical muscle stimulation | Consistency of the valuable achievement of BSA was guaranteed. |
| [149], A. El-Fergany | BSA | Optimally allocate multi-objective of | Type-3 distributed electric generator unit (delivers P and injects Q) was |
| | | multi-type and distributed electric generators in distributed and radial networks | preferable in reducing losses of power in network lines and in boosting voltage stability indices and bus voltage profile. |
| [150], K. Guney, and A. Durmus | BSA | Solving linear antenna pattern with nulls prescription at interference directions | The use of BSA was provided an array of a linear antenna, deep null levels, and low side-lobe levels. |
| [151], A.E. Chaib et al. | BSA | Dealing with optimal power flow problem. | Performance of BSA was increased in comparison with other evolutionary algorithms. |
| [36], M.S. Ahmed | New binary BSA | Developing optimal controller in the real-time schedule for home energy management. | It was used for managing energy consumption for home. |
| [152], K. Bhattacharjee | BSA | Dealing with economic emission load dispatch problem | It was used for solving EELD problems with high performance. |
| [72], A. Askarzadeh, and L. dos Santos Coelho, | BSA with a combination of Burger's chaotic map. | Conducting parameters estimation for fuel cells in the electrochemical model. | It was efficient to deal with the problem of complexity of PEMFC parameter estimation. |
| [153], M. Shafiullah et al. | BSA | Tuning the parameters of power system stabilisers to propose a new method for multi-machine power system simultaneously | The proposed approach was compared with particle swarm optimisation, and it gave a confidence result. |
| [154], R.-E. Precup et al. | BSA based on tuning control parameters of proportional-integral-derivative | Pancaking direct current in torque motor systems. | It was able to enhance the performance of torque motor systems. |
| [69], S. Das et al. | BSA-DE | Solving sidelobe suppression problem of a uniformly exciting array of concentric circular antenna arrays | High performance and quality of BSA-DE on one or both structural parameters and pattern parameters was enhanced. |
| [155], C. Zhang et al. | Hybrid BSA | Forecasting wind speed | (i) It was able to decompose wind speed by optimising vibrational decomposition mode.<br>(ii) It was able to optimise the parameters of extreme learning machine (ELM) and the input matrix.<br>(iii) OVMD-HBSA-ELM was performed better with respect to predicting accuracy. |
| [156], U. Kılıç | BSA | Dealing with optimal power flow in power systems | OPF problem of power systems was solved. |
| [20], C. Lu et al. | HMOBSA | 1. Modelling an efficient-energy-scheduling efficiently with controlled transportations. | The HMOBSA was performed better than other algorithms for solving the studied problem. |
| | | 2. Proposing an effective energy-saving strategy. | |
| [157], A.M. Shaheen et al. | BSA | Reducing losses of power and voltage improvement based on variable constraints. | BSA was optimally able to tackle reactive power dispatch problem. |
| [152], K. Bhattacharjee | BSA | Dealing with ELD problem, such as power balance. | The proposed method was robust. |
| [158], D.S.K. Kanth et al. | Modified BSA with traditional significant bang-big crunch method | Sizing and placement of distributed electric generators in terms of voltage control | It was simulated in MATLAB and experimented on the IEEE 33-bus feeder system. |
| [159], P. Gupta et al. | Four optimisation techniques, including BSA and grey wolf optimiser (GWO) | Scaling factors optimisation for the controller of fuzzy proportional-integral | GWO was performed better than the other optimisation algorithms on the considered cases. |
| [85], W.U. Khan et al. | BSA-SQP | Solving non-linear ANC systems | BSA-SQP was accurate and able to minimise complex problems with fast convergence. |
| [86], M. Sriram, and K. Ravindra | BSA-based MPPT | Analysing I-V and P-V characteristics of different solar PV array configurations | BSA-based MPPT was performed better compared to competitive algorithms (INC and ABC based MPPT) except P&O. |
| [88], Z. Tian et al. | BSA-based LSSVM | Predicting short-term wind speed | Reliability and accuracy for predicting short-term wind speed of BSA-based LSSVM were high. Meanwhile, the performance indicators of prediction were enhanced. |
| [160], B. Baadji et al. | BSA | Designing power system stabilisers design for multi-machine systems (WAPSS) | The robustness of the controller was guaranteed, the system stability was enhanced effectively with significant time delay. |
| [161], N. Niu et al. | BSA, IBSA | Determining the global optima using hybrid PV/Wind system with battery storage for Dakhla smart city electrification | IBSA was more robustness and effectiveness compared to BSA. |
| [162], Y.C. Kuyu et al. | BSA | Minimising analogue filter group delay for Chebyshev filter by exploiting the structures of the all-pass filter with different orders. | All the competitive algorithms used in the experiment had the same performance for designing the second-order all-pass filter structure, whereas DE gave better in group delay than the rest of its counterpart methods. BSA found better group delay than the competitive algorithms BSA was the best option for optimising group delays. |
| [92], S. Sun et al. | EEMD/VMD-HSBA-DAWNN | Forecasting wind speed modelling strategy | It was effective for multi-step WSF strategy. |
| [163], A.R. Jordehi | BSA | Minimising problem of DG allocation | In most cases, GWO was outperformed BSA, PSO, and WOA. |
| [164], M. Shafiullah et al. | BSA | Solving optimal PMU placement problem to minimise the need number of PMU to ensure full distribution grids observability | Excellency of the introduced model and solution methodology in comparison other counterpart methodologies was demonstrated. |
| [131], R. El Maani et al. | BSAMO | Dealing with Fluid-structure interaction (FSI) problems | BSAMO was more competitive and successful in coping with complex problems. |
| [126], A.M. Shaheen, and R.A. El-Sehiemy | Binary and integer coded BSA | Minimising transmission network expansion planning (TNEP) taking the security limitation into account | The proposed technique had more capability with satisfactory technical and economic benefits in solving those two problems than other counterpart techniques. |
| [98], A. Pourdaryaei et al. | MOBBSA | Forecasting short period electricity sales | The superiority of MOBBSA was demonstrated compared to ANN and ANFIS. |
| [46], K. Yu et al. | MLBSA | Determining photovoltaic models parameters | MLBSA had better performance compared with other algorithms in the literature concerning reliability, computational efficiency, and accuracy. |
| [99], J. Zhou et al. | IBSA | Coping with the parameter identification problem for pump turbine governing system (PTGS) | IBSA performed better in comparison with GSA, PSO, and the standard BSA in terms of parameter identification accuracy and quality |
| [48], H. Yang et al. | Modified BSA | Enhancing the stability of photovoltaic (PV) | It was effectual for enhancing the stability of PV microgrids once it suffered from small disturbance. |
| [165], A. Khamis et al. | BSA | Introducing and developing an optimal load shedding scheme | It was more effectual than GA in finding the optimal amount of load. |
| [166], S.S. Khan et al. | BSA | Patterning the proton exchange membrane fuel cell (PEMFC), and retrieving model parameters. | BSA had better results than PSO in patterning PEMFC output voltage. |
| [167], N.N. Islam et al. | BSA | Designing power system stabilisers (PSSs) for an extensive power system. | A considerable number of parameters did not affect the performance of BSA for designing PSS compared to BFOA and PSO |
| [40], J. Kartite et al. | IBSA | Minimising renewable energy costs | Effectiveness of IBSA was demonstrated. |
| [133], C. Lu et al. | MODBSA | Minimising compression cost total and tardiness of CPT | MODBSA was effective for such a schedulable program. |
| [105], G. Mohy-ud-din | Hybrid BSA-SQP | Optimising emission and operative cost in transferring power to the generation units | Efficiency of hybrid BSA-SQP was promising in reducing emissions and operative cost. |
| [65], P. Chen et al. | HBSA | Solving energy consumption for PFSP and makespan | Effectiveness of HBSA was promising in comparison with GA and branch and bound algorithm. |
| [134], F. Daqaq et al. | MOBSA | Minimising high constrained and multi-objective optimisation problem | It was capable of minimising OPF problems in large-scale power systems with thousands number of buses. |
| [168], I. Elomary et al. | BSA | Validating the ability and efficiency of BSA by testing on brushless DC wheel motor (BLDC) design | The validation results were satisfactory for the Ability and efficiency of BSA on BLDC. |
| [169], B. Hiçdurmaz et al. | BSA | Predicting parameter values for a low-pass Sallen-Key topology Butterworth type active filter | It was successful in solving the design error in a short time. |
| [106], K. Lenin et al | HDSDE | Solving reactive power issue | It had better performance in reducing power loss. |
| [170], R. El Maani et al. | BSA | Minimising fluid-structure issue | BSA had better distribution and efficiency in comparison to NNIA. |
| [171], T.T. Nguyen et al. | BSA | Solving distribution network reconfiguration (DNR) issue | BSA was efficient and had promising results compared to CS and PSO. |
| [172], K. Dasgupta, and S.K. Ghorui | BSA | Solving ELD problem | BSA was a good tuner as it was found as the best algorithm compared PSO, PSOCFIWA, PSOCFA, and PSOIWA algorithms. |
| [173], A.M. Shaheen et al | BSA | Optimising reactive power dispatch (RPD) problem | BSA had fewer power losses in comparison to GA, PSO, LP, and ACO. |
| [135], C. Lu et al. | MOBSA | Optimising energy-efficient multi-pass turning operation problem | MOBSA was outperformed MOPSO, NSGA-II, MOHS, and cMOEA/D. |
| [174], M. Modiri-Delshad | BSA | Minimising economic dispatch problems with valve-point effects and variance fuel options | Robustness of BSA was confirmed. |
| [175], Z. Wei, and Q. Wei | BSA | Determining the optimum frequency band and the optimal combination of time segment and frequency band | Superiority of BSA’s performance was promising. |
| [108], S. Mallick et al. | BSA-DE | Designing two well-known analogue CMOS amplifier circuits optimally | Effectiveness of BSA-DE was promising in comparison to DE, ABC, harmony search (HS), and PSO. |
| [176], N. Tyagi et al. | BSA | Minimising ELD problem without or | BSA was substantially better than each of DE, GA, IPSO_TVAC, CRO, TVPSOGSA, and SA. |
| | | with the integration of solar power systems | |
| [177], W. Jianjun et al. | BSA | Finding the parameters of support vector regression (SVR) for forecasting annual power usage | Performance of BSA was better than BPNN, SVR, and regression forecasting methods in forecasting power loads annually. |
| [178], N.N. Islam et al. | BSA | Gaining the coordination of directional overcurrent relays optimally | (i) Performance of BSA was effective and viable for optimising relay coordination problem compared to its counterpart algorithms.<br>(ii) BSA enhanced the system's stability by up to 80%. |
| [179], A. El-Fergany | BSA | Finding optimal allocation of various distributed generators (DGs). | BSA significantly reduced voltage deviations and power losses. |
| [180], A. Garbaya et al. | BSA | Optimally designing of RF<br>circuits. | (i) BSA was outperformed other counterpart algorithms concerning computation time.<br>(ii) BSA was considered as an exciting algorithm for computer-aided design tool/approach. However, PSO was more robust than BSA. |
| [109], B. Benhala et al. | BA-ACO | Optimising analogue circuit performance | validity of BA-ACO was indicated. |

**4.2.3. Artificial intelligence.** Problems in artificial intelligence have also been dealt with by BSA. Table 8 presents summary studies of solving artificial intelligence problems by BSA.

**Table 8: Summary of BSA applications in artificial intelligence**

| **Source, author(s)** | **Algorithm** | **Aim of the algorithm** | **Result** |
|---|---|---|---|
| [73], S.K. Agarwal et al. | BSANN | Classifying three mental tasks (movement generation of left or right hand and imagination of word) | Accuracy of BSANN was better compared to other competitive algorithms for mental task classification. |
| [74], J.A. Ali et al. | BSA-based fuzzy logic | Controlling three-phase induction motor (TIM) variables | BSAF controller was outperformed PSO controllers and GSA and PSO controllers. |
| [128], F. Zou et al. | Multi-objective discrete BSA | Solving community detection problem in complex networks. | Performance of the proposed algorithm was promising and effective for solving the problem of community detection in complex networks. |

**4.2.4. Information and communication engineering.** Backtracking search optimisation algorithm has also been used for solving technology and network communication problems. Table 9 gives a summary of the studies conducted for solving the mentioned problems.

**Table 9: Summary of BSA applications in information and communications technology**

| Source, author(s) | Algorithm | Aim of the algorithm | Result |
|---|---|---|---|
| [181], N. Niu et al. | BSA of fault-tolerant topology reconfiguration | Tolerating faulty core in 2D REmesh based networks on chip | Successful reconfiguration rate was gained on distinctive sizes of topologies. |
| [182], K. Ayan, and U. Kılıç | BSA | Solving OPF of two-terminal HVDC systems. | Reducing CPU times and production costs were demonstrated compared to the other methods. |
| [150], K. Guney, and A. Durmus | BSA | Combining concentric circular antenna arrays (CCAAs) at a fixed beam width with the low sidelobe levels | BSA was able to design CCAA to provide fixed beam width and proper sidelobe levels. |
| [183], J. Yan, and J. Zhang | SAT-based approach and a BSA | Generating test studies for combinational testing | This method performed better than the other methods in small size problems. |
| [184], F. Zaman et al. | BSA | Analysing pattern and correcting faulty antenna array in wireless (mobile) communication | BSA was computationally produced better results in comparison with its counterpart algorithms (GA and GA-PS) |
| M.M. Badawy, et al., [185] | BSA | Introducing a dynamic QoS for providing a framework (QoPF) for service-oriented IoT | BSA was introduced worthier QoPF to meet QoS requirements in comparison with other algorithms in the literature review. |
| [90], Z. Wang et al | Basic ESN, the three optimised ESNs, APSS&AMFS–BSA-ESN | Determining the most suitable output weights of ESN | Accuracy of optimised ESNs was better than the basic ESN, and performance of APSS&AMFS–BSA-ESN was relatively better than the basic ESN, the three optimised ESNs, and the other competitive approaches. |
| [186], M. Eskandari, O. Sharifi | BSA | Selecting the effective feature sets of both modalities on gender classification | Robust work was produced for face-ocular multimodal biometric system for gender classification. |
| [81], J. Lin | BS-HH | Solving flexible job-shop scheduling problem with fuzzy processing time (FJSPF) | BS-HH was outperformed its counterpart algorithms for minimising FJSPF. |
| [97], L. Chen et al. | ELM-BSA | Forecasting flood occurrence | Performance of ELM-BSA was better than ELS and GRNN, and it was suggested as a useful alternative for flood forecasting. |
| [187], R.A. Osama et al. | BSA, graph-based theories, and probabilistic load flow approach | Accomplishing design of distributed system with several distributed generation units | The planned framework was sensitive to the system's security, reliability, and economics needs compared to other microgrid design. |
| [188], G.S. Walia et al. | BSA | Evaluating performance of Multimodal biometric system on the basis of a finger vein, iris, and fingerprint | High accuracy and low equal error rate of BSA were demonstrated. |
| [189], A.O. de Sá et al. | BSA | Improving accuracy of bio-inspired system identification (BiSI) attacks in noisy Networked Control Systems (NCS) by BSA along with noise processing technique | Accuracy of the proposed model was enhanced. |
| [100], J. Zhou et al. | VMD-BSA-RELM | Improving wind forecasting accuracy | The suggested approach performed better than its both multi- and single-step forecasting by 50%. |
| [101], W. Zhang et al. | BSO-FCM | Solving the problem of high-order fuzzy-trend forecasting | It had remarkable forecasting accuracy than other benchmark approaches on the TAILEX. |
| [190], N.N.A. Nazri et al. | BSA | Collaborating beamforming in wireless sensor networks (WSN). | The side-lobe level suppression for BSA was better in comparison with other counterpart approaches. |
| [191], A. Gosain, and K. Sachdeva | BSA | Nominating materialised views in a data warehouse. | (i) It was materialised view selection model (BSMVSA) was more excellent than PSO and GA.<br>(ii) It substantially reduced the cost in the storage constraint. |
| [104], H. Li et al. | RBM-BSASA-BP | Solving time series forecasting problem | The suggested method was robust and reliable and usable for large-scale problems. |
| [102], U.H. Atasever et al. | SEBAL-BSA | Selecting wet-hot pixel with ground control points automatically | It was significantly successful and a user-friendly method. |
| [103], G. Jothi et al. | BSA-based clustering | Developing GUI for ALL image clustering and classification | It had better efficiency and robustness compared with ABC, PSO, and DE. |
| [107], B. Wang et al. | BSA-NNRWs-N | Minimising hidden parameters of single-layer feed-forward | BSA-NNRWs-N had a promising performance. |

**4.2.5. Material and civil engineering.** According to the studies summarised in Table 10, material engineering problems have been solved by using a backtracking search optimisation algorithm.

**Table 10: Summary of BSA applications in material engineering**

| Source, author(s) | Algorithm | Aim of the algorithm | Result |
|---|---|---|---|
| [192], A. Montanaro | Quantum speedup algorithms based on backtracking technique | Solving constraint satisfaction problems (CSPs) | The algorithm was able to solve CSPs. |
| [193], Y.-K. Lin and T.-P. Nguyen | A multistate flight network (MSFN) with backtracking technique | Evaluating flight network reliability | A tool for supporting executive staffs in evaluating and observing their flight network was provided for sustainability. |
| [194], P. Mishra et al. | backtracking search algorithm (method 1 and method 2) | Estimating nonlinear system parameter efficiently | Estimating parameter using Method 2 were outperformed Method 1 by using lower SAE values. |
| [195] | Improved BSA | Casting charge plan problem of heat treatment | The proposed algorithm was effective and well performed. |
| [45], A. Chatzipavlis et al. | BSA-based neuro-fuzzy network | Modelling beach realignment | It had better performance compared to other competitive methods for modelling beach realignment. |
| [196], X. Song et al. | BSA | Analysing surface wave efficiently and effectively | BSA had outperformed GA in analysing surface wave. |
| [54], S. Vitayasak et al. | mBSAs | solve DFLP with a heterogeneous resource | (i) mBSAs had better solution than GA for large-scale problems, and it the |
| | | | costs used to generate the layout of the best mBSA were considerably lower than for the standard BSA.<br>(ii) mBSAs enlarged the diversity of solutions, and increased exploration capability accordingly.<br>(iii) The execution time needed for the mBSAs was less than GA by 70%. |

## 5. Analytical evaluation of BSA in the literature

In this section, the experiments and statistical analysis conducted in the literature on BSA by using the benchmark functions are presented. Additionally, the performance of BSA in previous studies is presented.

### 5.1. Statistical analysis

In this analysis, 75 benchmark functions were used in three tests to examine the success of BSA in comparison with the other heuristic algorithms [11]. Test 1 uses 50 benchmark test functions. The details of these benchmark test functions are provided in [197,198]. Furthermore, 25 benchmark functions are used in Test 2. The details of these benchmark functions are provided in [11]. Additionally, three benchmark test functions are used in Test 3. The details of these benchmark functions are provided in [199]. In the literature, the Wilcoxon signed-rank test was used to determine which two algorithms achieved better solution in terms of solving numerical problems statistically. In the same study [11], necessary statistics measures such as standard deviation, best, mean, and runtime were achieved by BSA, PSO, ABC, CMAES, JDE, SADE, and CLPSO in 75 mathematical benchmark problems for 30 solutions in 3 tests considering the statistical significant value (α) as 0.05 and the null hypothesis (H0). Tests 1, 2, and 3 were regarding the basic statistical measures (standard deviation, best, mean, and runtime) achieved by BSA, PSO, ABC, CMAES, JDE, SADE, and CLPSO for 30 solutions for 50 (F1-F50), 25 (F51-F75), and 3 (F76-F78) benchmark problems, respectively. These tests were used to compare BSA with PSO, ABC, CMAES, JDE, SADE, and CLPSO to statistically determine which of them provided the best solution. Fig. 6 illustrates a graphical comparison of BSA versus CMAES, PSO, ABS, CLPSO, JDE, and SADE to statistically determine which of them provided the best solution for the benchmark problems used in Tests 1, 2, and 3.

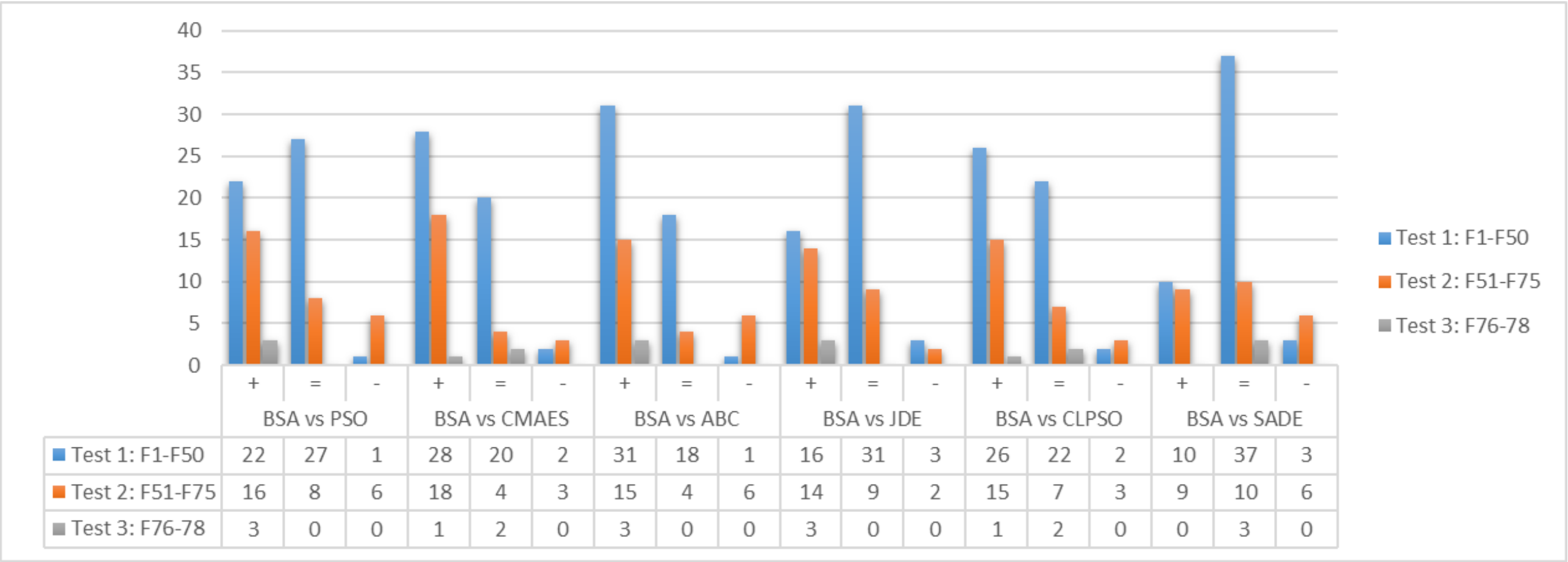

| | BSA vs PSO | | | BSA vs CMAES | | | BSA vs ABC | | | BSA vs JDE | | | BSA vs CLPSO | | | BSA vs SADE | | |
|---|---|---|---|---|---|---|---|---|---|---|---|---|---|---|---|---|---|---|
| | + | = | - | + | = | - | + | = | - | + | = | - | + | = | - | + | = | - |
| Test 1: F1-F50 | 22 | 27 | 1 | 28 | 20 | 2 | 31 | 18 | 1 | 16 | 31 | 3 | 26 | 22 | 2 | 10 | 37 | 3 |
| Test 2: F51-F75 | 16 | 8 | 6 | 18 | 4 | 3 | 15 | 4 | 6 | 14 | 9 | 2 | 15 | 7 | 3 | 9 | 10 | 6 |
| Test 3: F76-78 | 3 | 0 | 0 | 1 | 2 | 0 | 3 | 0 | 0 | 3 | 0 | 0 | 1 | 2 | 0 | 0 | 3 | 0 |

**Fig. 6: Statistical performance comparison of comparison algorithms on seventy-eight benchmark problems used in Tests 1,2, and 3**

In Fig. 6, '+' means that BSA has statistically better performance than the other algorithm at null hypothesis H0 and α = 0.05; '-' indicates that BSA has statistically lower performance compared to its competitor at null hypothesis H0 and α = 0.05; and '=' indicates that BSA has statistically the same performance as its competitor at null hypothesis H0 and α = 0.05. Furthermore, Table 6 compares BSA with other EAs for the statistical pairwise multi-problem at a significant statistical value (α = 0.05) and the null hypothesis (H0). The result of the experiment conducted by [11] showed that (i) CMAES, PSO, ABS, CLPSO, and JDE are statistically less successful than BSA in Tests 1 and 2, (ii) BSA is statistically as successful as SADE in Test 3, and (iii) BSA is more successful than PSO, ABC, CMAES, JDE, and CLPSO in Test 3.

### 5.2. Performance evaluation

In seen in the literature, BSA has been used to deal with different types of problems, and it has been evaluated against other popular EAs. Table 11 summarises the performance evaluation of BSA against its competitors for different types of problems. For most of the problems, BSA showed superior performance than its competitors of BSA performed better than BSA. For example, GA showed improved performance than BSA for the personal scheduling problem.

**Table 11: performance evaluation of the standard BSA against its competitors for different types of problems**

| Type of the problem | Source, author(s) | BSA's competitor(s) | Type of comparison | Winner |
|---|---|---|---|---|
| Numerical optimization problem | [11], P. Civicioglu | CMAES, PSO, ABS, CLPSO, JDE, and SADE | Effectiveness and performance | BSA |
| Personal scheduling | [200], A.C. Adamuthe, and R.S. Bichkar | GA | Performance for value and variable orderings and techniques of consistency enforcement | GA |
| Combination of centric circular antenna arrays | [150], K. Guney, and A. Durmus | Differential search (DS) and bacterial foraging algorithm (BFA) | Iterative performance | BSA |
| Three design problems circular antenna array | [125], P. Civicioglu | ABC, E2-DSA, CK, CLPSO, CMAES, ACS, JADE, DE, EPSDE, JDE, SADE, GSA, S-DSA, PSO | Side lobe levels minimization, maximum directivity acquisition, non-uniform null control, and array of planar circular antenna. | S-DSA |
| Economic dispatch problem | [61], A.F. Ali | GA, PSO, BA | Performance | Memetic BSA |
| Optimal power flow problem | [156], U. Kılıç | Those heuristic algorithms that have solved OPF previously, such as GA, SA, Hybrid SFLA-SA, and ABC | Effectiveness and efficiency | Effect and quick reach of BSA to global optimum |
| Optimal power flow problem | [151], A.E. Chaib et al. | DE, PSO, ABC, GA, BBO | Effectiveness and efficiency | BSA |
| Fed-batch fermentation optimisation | [136], M.Z. bin Mohd Zain et al. | DE, CMAES, AAA, and ABC | Performance and robustness of convergence | BSA |
| Global continuous optimisation | [201], S. Mandal et al | ALEP, CPSO-H6, HEA, HTGA, GAAPI | CPU Efficiency and solution of global optima | BSA |
| Fuzzy proportional-integral controller (EPCI) for scaling factor optimisation | [159], P. Gupta et al. | GWO, DE, BA | Performance | GWO |
| stabilisers design of multi-machine power systems | [153], M. Shafiullah et al. | PSO | Robustness | BSA |
| Allocating Multi-type distributed generators | [149], A. El-Fergany | Golden search, Grid search, Analytical, Analytical LSF, Analytical ELF, PSO, ABC | Optimal result | BSA |
| Linear antenna arrays | [150], K. Guney, and A. Durmus | PSO, GA, MTACO, QPM, BFA, TSA, MA, NSGA-2, MODE, MODE/D-DE, CLPSO, HAS, SOA, MVMO, | Pattern nulling | BSA |
| Estimating parameter for non-linear Muskingum model | [59], X. Yuan et al. | PSO, GA, DE | Efficiency | BSA |
| Cost-sensitive feature selection | [123], H. Zhao et al. | Heuristic algorithm | Efficiency and effectiveness | Efficiency: BSA Effectiveness: a heuristic algorithm |
| load frequency control (LFC) in power systems engineering | [141], V. Kumar et al. | BFOA, PSO, GA, DE, and GSA | Performance, robustness, and effectiveness | BSA |

## 6. Proposed operational framework of BSA

This section describes a proposed operational framework of the main expansions of BSA (hybridised BSA, modified BSA, and extensions of BSA) and the implementations of BSA that were concluded from the literature on BSA. The procedures of these three expansions from the basic BSA vary. The modification of BSA can be based on the combination of BSA with another meta-heuristic algorithm, control factor (F), historical information, routing problem, or some different cases such as binary BSA, hybridised MOBSA with a GA, and the use of BSA in the microprocessor. On the contrary, the procedure of hybridizing the BSA can be based on control factor (F), guided and memetic BSA, neural network, fuzzy logic, local and global searches, and hybridization of BSA with another meta-heuristic algorithm and mathematical functions such as combination of BSA with Burger's chaotic map with wavelet-based mutation and SVM. Meanwhile, BSA has been extended to be used for binary optimisation, numerical optimisation, constrained optimisation, and multi-objective optimisation. These extensions are based on the combination of BSA with another meta-heuristic algorithm, three constraint handlers, network topology, new mutation, and crossover operations. A brief summary of the different expansions of BSA and their expansion procedures is presented in an operational framework in Fig. 7.

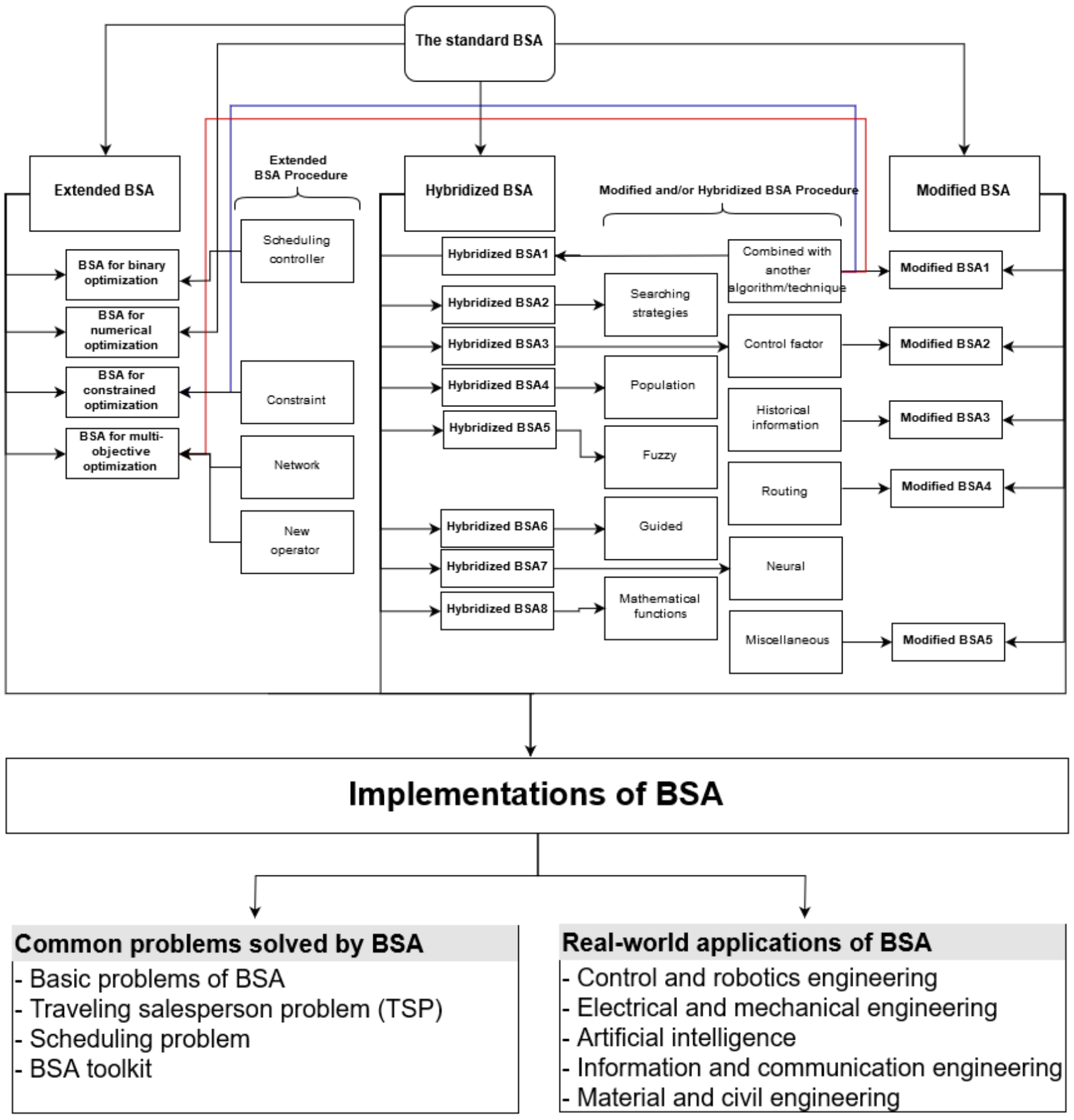

**Fig. 7: The proposed operational framework of BSA**

## 7. Experiments

From the previous studies that were conducted to compare the nature of the performance of BSA with its competitors, a mean result, standard deviation, the worst and best result were extracted [11,202,203]. Nevertheless, an equitable and fair comparison of a number of metaheuristic algorithms in terms of performance is a difficult task owing to several reasons. Particularly, selection and initialisation of the key parameters such as the problem dimension, the problem's search space for each algorithm, and the number of iterations required to minimise the problem are challenging. Another issue is the involvement of a randomness process in every algorithm. To weave out the effects of this randomisation and obtain an average result, the tests must be run on the algorithms several times. The performance of the system used to implement the algorithms and the programming style of the algorithms may also affect the algorithms' classification performance. To mitigate this effect, a consistent code style for all the algorithms on the same system can be utilized. A further challenging task is choosing the type of problems or benchmark test functions used to evaluate the algorithms. For instance, an algorithm may perform satisfactorily for a specific type of problem but not on other types of problems. Additionally, the size of a problem can affect an algorithm's performance. A set of standard problems or benchmark test functions with different levels of difficulties and different cohorts must be used to tackle the aforementioned issue, and thereby test the algorithms on various levels of optimisation problems. Therefore, an experimental setup to fairly compare BSA with its competitive meta-heuristic algorithms requires the consideration of aspects such as initialization of the control parameters, balance of the randomness process of the algorithms, computer performance used to implement the algorithms, programming style of the algorithms, and type of the tackled problems; this is owing to the lack of experimental study on BSA in the literature. Therefore, this section details the experimental setup, the list of benchmark functions used in the experiments with their control parameters, statistical analysis of the experiment, pairwise statistical testing tools, and the experimental results.

### 7.1. Experimental setup

The experiment is set up to fairly compare BSA with PSO, ABC, FF, and DE by considering the initialization of control parameters such as the problem dimensions, problem search space, and number of iterations required to minimise the problem, performance of the system used to implement the algorithms, the programming style of the algorithms, achieving balance on the effect of randomization, and the use of different types of optimisation problems in terms of hardness and cohorts. This experiment is about the unbiased comparison of BSA with PSO, ABC, FF, and DE on 16 benchmark problems with different levels of hardness [204–206] in three tests as follows:

1. Several iterations are needed to minimise a specific function with Nvar variables with the default search space for the population size of 30. Nvars take values of 10, 30, and 60. For each benchmark function, each algorithm is run for 30 times with 2000 iterations for Nvar values of 10, 30, and 60.
2. Several iterations are needed to minimise the functions with two variables for three different sized solution spaces for the population size of 30. For each benchmark function, each algorithm is run for 30 times with 2000 iterations for three different ranges (R1, R2, and R3), where

(a) R1: [-5, 5]

(b) R2: [-250, 250]

(c) R3: [-500, 500]

3. Determining the ratio of successful minimisation of the functions for Tests 1 and 2 is needed to compare the successful rate of BSA with its competitive algorithms.

Fig. 8 depicts the framework for the experimental setup that includes the processes of Tests 1, 2, and 3.

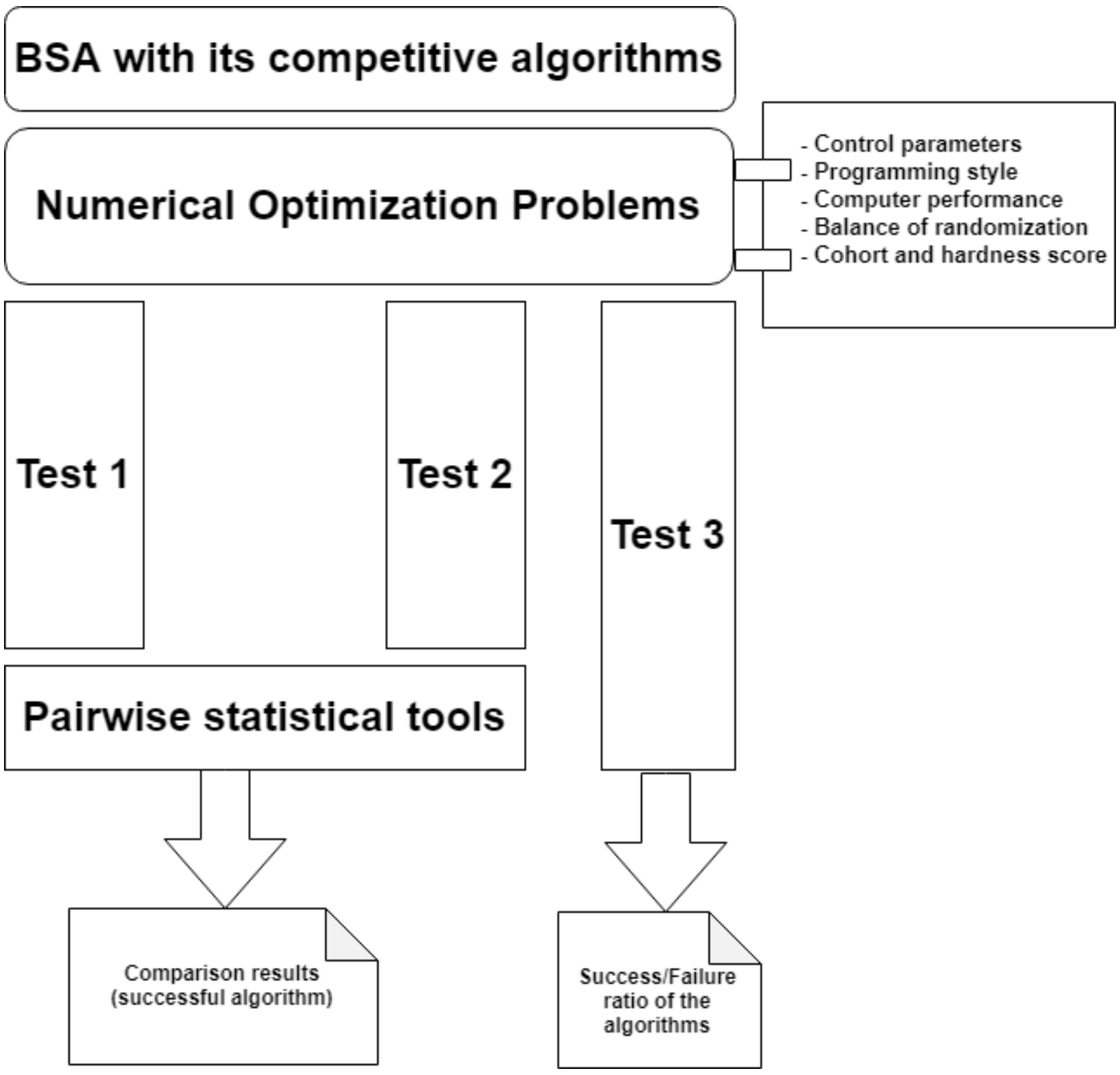

**Fig. 8: The proposed framework for the experimental setup**

### 7.2. Benchmark functions and control parameters

To examine the relative success of BSA compared to PSO, DE, ABC, and FF in solving a variety of optimisation problems based on their level of hardness and cohort [204–206], 16 benchmark problems were used in all the three tests. Regardless of the algorithm chosen to solve the test functions, some of the chosen problems are harder to minimise than the others. Table 12 lists the chosen optimisation algorithms with their search space, global minima, dimension, and percentage level of hardness. The percentage level of hardness ranges between 4.92 to 82.75%. Among the problems listed in the table, the Whitley function is the hardest, whereas the Sphere is the easiest.

**Table 12: Benchmark testing functions.**

| ID | Function | Search space | Global min. | Dim. | Overall success (%) |
|---|---|---|---|---|---|
| F1 | Ackley | [-32, 32] | 0 | n | 48.25 |
| F2 | Alpine01 | [0, 10] | 0 | 2 | 65.17 |
| F3 | Bird | [-6.283185, | -106.76453 | 2 | 59.00 |
| F4 | Leon | [0, 10] | 0 | 2 | 41.17 |
| F5 | CrossInTray | [-10, 10] | -2.062611 | 2 | 74.08 |
| F6 | Easom | [-100, 100] | -1 | 2 | 26.08 |
| F7 | Whitley | [-10.24, | 0 | 2 | 4.92 |
| F8 | EggCrate | [-5, 5] | 0 | 2 | 64.92 |
| F9 | Griewank | [-600, 600] | 0 | n | 6.08 |
| F10 | HolderTable | [-10, 10] | -19.2085 | 2 | 80.08 |
| F11 | Rastrigin | [-5.12, 5.12] | 0 | n | 39.50 |
| F12 | Rosenbrock | [-5, 10] | 0 | n | 44.17 |
| F13 | Salomon | [-100, 100] | 0 | 2 | 10.33 |
| F14 | Sphere | [-1, 1] | 0 | 2 | 82.75 |
| F15 | StyblinskiTang | [-5, 5] | -39.1661 for | n | 70.50 |
| F16 | Schwefel26 | [-500, 500] | 0 | 2 | 62.67 |

Additionally, for each benchmark function, each algorithm is run for 30 times with maximum 2000 iterations with population size of 30 for all the three tests. Meanwhile, the search spaces and dimensions of the benchmark problems vary from Test 1, 2, and 3. For Test 1, the dimension of each benchmark function is set to default, and its search space is categorised as R1, R2, and R3. For Test 2, the search space of each benchmark function is set to two variables, and its dimension is categorised as Nvar1, Nvar2, and Nvar3. Finally, determining the ratio of successful minimisation of the functions for Tests 1 and 2 is needed to compare the successful rate of BSA with its competitive algorithms. This initialisation of parameters is described in Table 13.

**Table 13: Control parameters of Test 1 and 2**

<table>
<tr><th rowspan="2">Tests</th><th colspan="3">Features of optimisation problems</th><th colspan="3">Initial parameters</th></tr>
<tr><th>Dimension</th><th>Search space</th><th>Hardness score</th><th>Maximum run</th><th>iterations</th><th>Population size</th></tr>
<tr><td rowspan="3">Test 1</td><td rowspan="3">Default dimension</td><td>R1: [-5, 5]</td><td rowspan="6">4.92% - 82.75%</td><td rowspan="6">30</td><td rowspan="6">2000</td><td rowspan="6">30</td></tr>
<tr><td>R2: [-250, 250]</td></tr>
<tr><td>R3: [-500, 500]</td></tr>
<tr><td rowspan="3">Test 2</td><td>Nvar1: 10</td><td rowspan="3">2</td></tr>
<tr><td>Nvar2: 30</td></tr>
<tr><td>Nvar3: 60</td></tr>
</table>

### 7.3. Statistical analysis

Occasionally, EAs may provide the worst and best solutions from time to time for a specific problem, e.g., if an algorithm runs for two times on a specific problem, it may obtain the best solution initially and the worst solution later, and vice versa. Based on this, statistical tools were used in the literature [11,22] to compare the success or failure of the problem-solving of BSA with the other EAs. In our experiment, the following seven statistical measures were used to solve the numerical optimisation problems: mean, standard deviation, best, worst, average computation time, number of successful minimisations, and number of failed minimisations.

### 7.4. Pairwise statistical tools

Pairwise statistical testing tools such as the Wilcoxon signed-rank test can be utilised to compare two algorithms and determine which algorithm can solve a specific optimisation problem with higher statistical success [11,207]. In the experiment, BSA was compared with other algorithms by using the Wilcoxon signed-rank test, where the significant statistical value (α) was considered as 0.05, and the null hypothesis (H0) for a specific benchmark problem was defined in Eq. (17) as follows:

$$Median\ (Algorithm\ A) = Median\ (Algorithm\ B) \quad (17)$$

To determine the algorithm that achieved a better solution in terms of statistics, R+, R-, and p-value were provided by the Wilcoxon signed-rank test to determine the ranking size. In the same experiment, GraphPad Prism was used to determine the R+ and R- values that ranged from 0 to 465. P-value was similar to the mathematical precision of current software and application development tools; it was 4-6. Generally, the precise value used for statistical Tests 1 and 2 was 6 because this precision level could be required in practical applications.

### 7.5. Result analysis

This section discusses the results of the three tests described in the previous section.

In Test 1, all the algorithms successfully minimised the four optimisation problems: F3, F4, F5, and F8. In this test, the algorithms were run on the optimisation problems on their default search spaces and three different dimensions: Nvar1, 2, and 3. Nevertheless, the dimensions of the numerical optimisation problems did not affect the success rate of the algorithms to minimise the problems because the algorithms could minimise four problems in three variable dimensions. Moreover, the level of hardness of the minimised benchmark functions was relatively low. The minimum and maximum levels of the overall success of these problems were 41.17 and 59%, respectively. Consequently, none of the algorithms minimised the problems with high hardness score in different variable dimensions. Meanwhile, F10 with overall success score of 80.08% was not solved by any of the algorithms in any variable dimensions. This implies that a clear conclusion cannot be drawn regarding the success of BSA and other algorithms in minimising the optimisation problems based on their hardness score. Therefore, there is no implicit correlation between the success of these algorithms and the difficulty level of minimising optimisation problems. A problem-based statistical comparison method was used to determine which of the algorithms used in the experiment could statistically solve the benchmark functions. This method uses the computation time required by the algorithms to reach the global minimum as a result of 30 runs. In the experiment, BSA was compared with other algorithms by using the Wilcoxon signed-rank test by considering the significant statistical value (α) to be 0.05. Further, the null hypothesis (H0) for a specific benchmark problem defined in Eq. (17) was considered. Tables 14, 15, and 16 list the algorithms that statistically gained better solutions compared to the other algorithms in Test 1 according to the Wilcoxon signed-rank test.

**Table 14: Statistical comparison to find the algorithm that provides the best solution for the solved benchmark function used in Test 1 (Nvar1) using two-sided Wilcoxon Signed-Rank Test (α=0.05)**

| **Problems** | **BSA vs DE** | | | | **BSA vs PSO** | | | | **BSA vs ABC** | | | | **BSA vs FF** | | | |
|---|---|---|---|---|---|---|---|---|---|---|---|---|---|---|---|---|
| | p-value | R+ | R- | Win | p-value | R+ | R- | Win | p-value | R+ | R- | Win | p-value | R+ | R- | Win |
| F3 | 0.0001 | 0 | 465 | + | 0.0004 | 49 | 329 | + | 0.0004 | 69 | 396 | + | 0.0001 | 463 | 2 | + |
| F4 | 0.0001 | 465 | 0 | + | 0.0003 | 399 | 66 | + | 0.6431 | 167 | 133 | - | <0.0001 | 465 | 0 | + |
| F5 | <0.0001 | 465 | 0 | + | 0.1094 | 311 | 154 | + | <0.0001 | 450 | 15 | + | <0.0001 | 463 | 2 | + |
| F8 | <0.0001 | 465 | 0 | + | <0.0001 | 465 | 0 | + | <0.0001 | 21 | 279 | + | 0.0006 | 347 | 59 | + |
| **+/=/-** | **4/0/0** | | | | **4/0/0** | | | | **3/0/1** | | | | **4/0/0** | | | |

**Table 15: Statistical comparison to find the algorithm that provides the best solution for the solved benchmark function used in Test 1 (Nvar2) using two-sided Wilcoxon Signed-Rank Test (α=0.05)**

| **Problems** | **BSA vs DE** | | | | **BSA vs PSO** | | | | **BSA vs ABC** | | | | **BSA vs FF** | | | |
|---|---|---|---|---|---|---|---|---|---|---|---|---|---|---|---|---|
| | p-value | R+ | R- | Win | p-value | R+ | R- | Win | p-value | R+ | R- | Win | p-value | R+ | R- | Win |
| F3 | <0.0001 | 450 | 15 | + | 0.0001 | 51 | 384 | + | <0.0001 | 0 | 465 | + | <0.0001 | 441 | 24 | + |
| F4 | <0.0001 | 465 | 0 | + | <0.0001 | 455 | 10 | + | <0.0001 | 465 | 0 | + | <0.0001 | 465 | 0 | + |
| F5 | <0.0001 | 463 | 2 | + | 0.3285 | 281 | 184 | + | 0.6263 | 208 | 257 | - | 0.0879 | 316 | 149 | + |
| F8 | <0.0001 | 465 | 0 | + | <0.0001 | 465 | 0 | + | <0.0001 | 3 | 432 | + | 0.0772 | 263 | 115 | + |
| **+/=/1** | **4/0/0** | | | | **4/0/0** | | | | **3/0/1** | | | | **4/0/0** | | | |

**Table 16: Statistical comparison to find the algorithm that provides the best solution for the solved benchmark function used in Test 1 (Nvar3) using two-sided Wilcoxon Signed-Rank Test (α=0.05)**

| **Problems** | **BSA vs DE** | | | | **BSA vs PSO** | | | | **BSA vs ABC** | | | | **BSA vs FF** | | | |
|---|---|---|---|---|---|---|---|---|---|---|---|---|---|---|---|---|
| | p-value | R+ | R- | Win | p-value | R+ | R- | Win | p-value | R+ | R- | Win | p-value | R+ | R- | Win |
| F3 | <0.0001 | 447 | 18 | + | <0.0001 | 1 | 350 | + | <0.0001 | 1 | 464 | + | <0.0001 | 426 | 39 | + |
| F4 | <0.0001 | 465 | 0 | + | <0.0001 | 421 | 44 | + | 0.9906 | 190 | 188 | - | <0.0001 | 465 | 0 | + |
| F5 | 0.0879 | 316 | 149 | + | <0.0001 | 0 | 465 | + | <0.0001 | 0 | 465 | + | <0.0001 | 2 | 463 | + |
| F8 | <0.0001 | 465 | 0 | + | <0.0001 | 465 | 0 | + | <0.0001 | 0 | 406 | + | <0.0001 | 12 | 339 | + |

| +/=/- | 4/0/0 | 4/0/0 | 3/0/1 | 4/0/0 |
|---|---|---|---|---|

In contrast, all the algorithms in Test 2 solved some of the optimisation functions from the highest hardness score of difficulty to the lowest. In this test, the algorithms were run on the optimisation problems in two dimensions and three different search spaces: R1, 2, and 3. Unlike Test 1, the variety of the search spaces affected the success rates of the algorithms to minimise the problems because the algorithms could have minimised a different number of problems in each search space. For example, out of 16, the number of optimisation problems minimised by all the algorithms in R1, 2, and 3 was 11, 9, and 8, respectively. Further, the level of difficulty of the solved benchmark functions varied from high to low. For example, the minimum and maximum levels of the overall success of these problems in search space R3 were 6.08 and 82.75%, respectively. Furthermore, none of the algorithms could minimise the problems with a high score of hardness in search space R3. F16 with overall success score of 62.67% could not be solved by any of the algorithms in search space R3. Similarly, F7 with overall success score of 4.92% could not be minimised by any of the algorithms in the same search space. This implies that a clear conclusion cannot be drawn on the success of BSA and other algorithms in minimising optimisation problems based on their hardness scores. Therefore, there is no implicit correlation between the success of these algorithms and the difficulty level of minimising optimisation problems. Similar to Test 1, Test 2 was about comparing BSA with other algorithms by using the Wilcoxon signed-rank test by considering the significant statistical value (α) as 0.05. Here, the considered null hypothesis (H0) for a specific benchmark problem is defined in Equation (7). Tables 17, 18, and 19 list the algorithms that statistically gained better solutions compared to the other algorithms in Test 2 according to the Wilcoxon signed-rank test.

**Table 17: Statistical comparison to find the algorithm that provides the best solution for the solved benchmark function used in Test 2 (R1) using two-sided Wilcoxon Signed-Rank Test (α=0.05)**

| **Problem** | **BSA vs DE** | | | | **BSA vs PSO** | | | | **BSA vs ABC** | | | | **BSA vs FF** | | | |
|---|---|---|---|---|---|---|---|---|---|---|---|---|---|---|---|---|
| | p-value | R+ | R- | Win | p-value | R+ | R- | Win | p-value | R+ | R- | Win | p-value | R+ | R- | Win |
| F2 | <0.0001 | 465 | 0 | + | <0.0001 | 406 | 0 | + | <0.0001 | 465 | 0 | + | <0.0001 | 465 | 0 | + |
| F4 | 0.0027 | 290 | 61 | + | <0.0001 | 465 | 0 | + | 0.1780 | 198 | 102 | - | 0.0004 | 44 | 307 | + |
| F5 | <0.0001 | 411 | 54 | + | <0.0001 | 433 | 32 | + | <0.0001 | 460 | 5 | + | <0.0001 | 450 | 5 | + |
| F6 | <0.0001 | 465 | 0 | + | 0.0137 | 114 | 351 | + | <0.0001 | 465 | 0 | + | 0.7611 | 217 | 248 | - |
| F8 | <0.0001 | 14 | 364 | + | <0.0001 | 465 | 0 | + | <0.0001 | 325 | 0 | + | <0.0001 | 371 | 7 | + |
| F9 | <0.0001 | 351 | 0 | + | 0.0001 | 105 | 0 | + | <0.0001 | 321 | 30 | + | <0.0001 | 435 | 0 | + |
| F11 | <0.0001 | 465 | 0 | + | <0.0001 | 462 | 3 | + | <0.0001 | 374 | 4 | + | <0.0001 | 322 | 3 | + |
| F12 | 0.7024 | 114 | 139 | - | <0.0001 | 465 | 0 | + | 0.0005 | 326 | 52 | + | 0.0179 | 68 | 232 | + |
| F13 | <0.0001 | 435 | 0 | + | <0.0001 | 325 | 0 | + | <0.0001 | 465 | 0 | + | <0.0001 | 377 | 1 | + |
| F14 | <0.0001 | 0 | 210 | + | <0.0001 | 465 | 0 | + | <0.0001 | 15 | 285 | + | <0.0001 | 0 | 325 | + |
| F15 | <0.0001 | 465 | 0 | + | <0.0001 | 465 | 0 | + | <0.0001 | 465 | 0 | + | <0.0001 | 465 | 0 | + |
| **+/=/-** | **10/0/1** | | | | **11/0/0** | | | | **10/0/1** | | | | **10/0/1** | | | |

**Table 18: Statistical comparison to find the algorithm that provides the best solution for the solved benchmark function used in Test 2 (R2) using two-sided Wilcoxon Signed-Rank Test (α=0.05)**

| **Problem** | **BSA vs DE** | | | | **BSA vs PSO** | | | | **BSA vs ABC** | | | | **BSA vs FF** | | | |
|---|---|---|---|---|---|---|---|---|---|---|---|---|---|---|---|---|
| | **p-val** | **R+** | **R-** | **Win** | **p-val** | **R+** | **R-** | **Win** | **p-val** | **R+** | **R-** | **Win** | **p-val** | **R+** | **R-** | **Win** |
| **F2** | <0.0001 | 407 | 28 | + | <0.0001 | 377 | 1 | + | <0.0001 | 457 | 8 | + | <0.0001 | 378 | 0 | + |
| **F4** | <0.0001 | 403 | 3 | + | <0.0001 | 231 | 0 | + | <0.0001 | 433 | 2 | + | <0.0001 | 325 | 0 | + |
| **F8** | 0.9789 | 164 | 161 | - | <0.0001 | 465 | 0 | + | <0.0001 | 325 | 0 | + | <0.0001 | 465 | 0 | + |
| **F9** | <0.0001 | 465 | 0 | + | <0.0001 | 267 | 0 | + | <0.0001 | 325 | 0 | + | <0.0001 | 435 | 0 | + |
| **F11** | <0.0001 | 646 | 1 | + | <0.0001 | 432 | 3 | + | <0.0001 | 351 | 0 | + | <0.0001 | 378 | 0 | + |
| **F12** | <0.0001 | 378 | 0 | + | <0.0001 | 465 | 0 | + | <0.0001 | 153 | 0 | + | <0.0001 | 374 | 4 | + |
| **F13** | <0.0001 | 464 | 1 | + | <0.0001 | 378 | 0 | + | <0.0001 | 378 | 0 | + | <0.0001 | 465 | 0 | + |
| **F14** | <0.0001 | 13 | 365 | + | <0.0001 | 465 | 0 | + | <0.0001 | 299 | 1 | + | <0.0001 | 405 | 1 | + |
| **F15** | <0.0001 | 465 | 0 | + | <0.0001 | 465 | 0 | + | <0.0001 | 465 | 0 | + | <0.0001 | 465 | 0 | + |
| +/=/- | **8/0/1** | | | | **9/0/0** | | | | **9/0/0** | | | | **9/0/0** | | | |

**Table 19: Statistical comparison to find the algorithm that provides the best solution for the solved benchmark function used in Test 2 (R3) using two-sided Wilcoxon Signed-Rank Test (α=0.05)**

| **Problem** | **BSA vs DE** | | | | **BSA vs PSO** | | | | **BSA vs ABC** | | | | **BSA vs FF** | | | |
|---|---|---|---|---|---|---|---|---|---|---|---|---|---|---|---|---|
| | p-value | R+ | R- | Win | p-value | R+ | R- | Win | p-value | R+ | R- | Win | p-value | R+ | R- | Win |
| F2 | <0.0001 | 433 | 32 | + | <0.0001 | 406 | 0 | + | <0.0001 | 450 | 15 | + | <0.0001 | 465 | 0 | + |
| F4 | <0.0001 | 406 | 0 | + | <0.0001 | 351 | 0 | + | <0.0001 | 378 | 0 | + | <0.0001 | 351 | 0 | + |
| F8 | 0.0027 | 290 | 61 | + | <0.0001 | 465 | 0 | + | <0.0001 | 325 | 0 | + | <0.0001 | 276 | 0 | + |
| F9 | <0.0001 | 6 | 345 | + | <0.0001 | 300 | 0 | + | <0.0001 | 325 | 0 | + | <0.0001 | 276 | 0 | + |
| F11 | <0.0001 | 422 | 43 | + | 0.2286 | 292 | 173 | + | <0.0001 | 604 | 0 | + | <0.0001 | 435 | 0 | + |
| F12 | <0.0001 | 350 | 1 | + | <0.0001 | 465 | 0 | + | 0.0156 | 28 | 0 | + | <0.0001 | 405 | 1 | + |
| F13 | <0.0001 | 377 | 1 | + | <0.0001 | 406 | 0 | + | <0.0001 | 351 | 0 | + | <0.0001 | 465 | 0 | + |
| F14 | 0.0118 | 94 | 312 | + | <0.0001 | 465 | 0 | + | <0.0001 | 431 | 4 | + | <0.0001 | 406 | 0 | + |
| +/=/- | **8/0/0** | | | | **8/0/0** | | | | **8/0/0** | | | | **8/0/0** | | | |

In Tables 14-19, '-' represents the cases of rejecting the null hypothesis and displaying BSA as statistically inferior performance in the statistical comparison of the problems; '+' represents that the cases of rejecting the null hypothesis and displaying BSA as statistically superior performance; '=' represents the cases with no statistical difference between the two comparison algorithms in the evaluation of the success of minimising the problems. The last rows of Tables 14-19 present the summation in the (+/=/-) format for these statistically significant cases: '+', '=', and '-' in the pair-wise problem-based statistical comparisons of the algorithms. Based on the examination of the (+/=/-) values in Tests 1 and 2, it can be concluded that BSA showed superior performance statistically than the other comparison algorithms in minimising numerical optimisation problems.

Overall, the results obtained from Tests 1 and 2 reveal that BSA is relatively more successful in solving numerical optimisation problems with different levels of hardness, variable size, and search spaces. However, none of the algorithms could minimise all the 16 benchmark functions successfully.

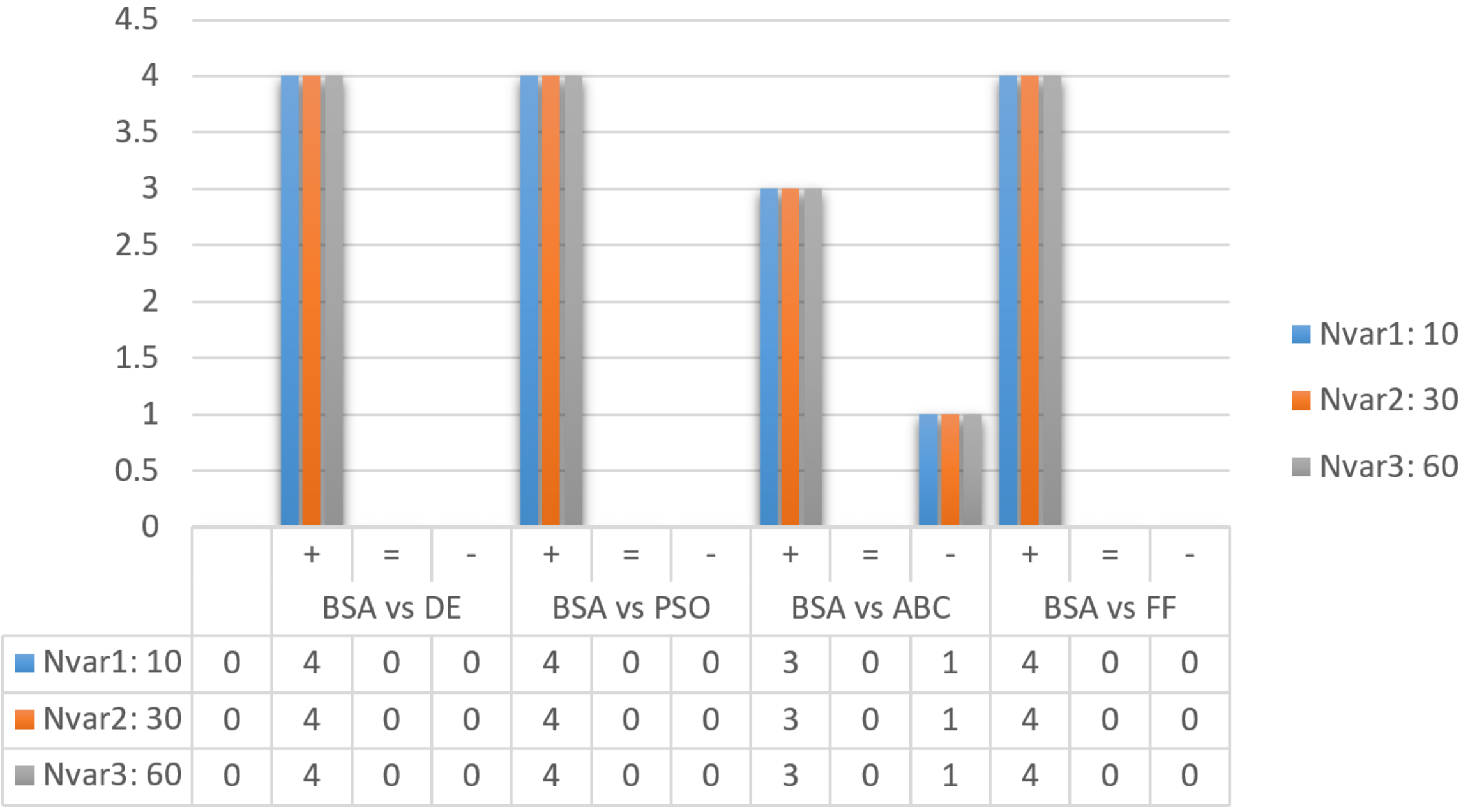

**Fig. 9: A graphical representation of Test 1**

The results listed in Tables 14, 15, and 16 are graphically represented in Fig. 9. Likewise, the results in Tables 17, 18, and 19 are graphically represented in Fig. 10.

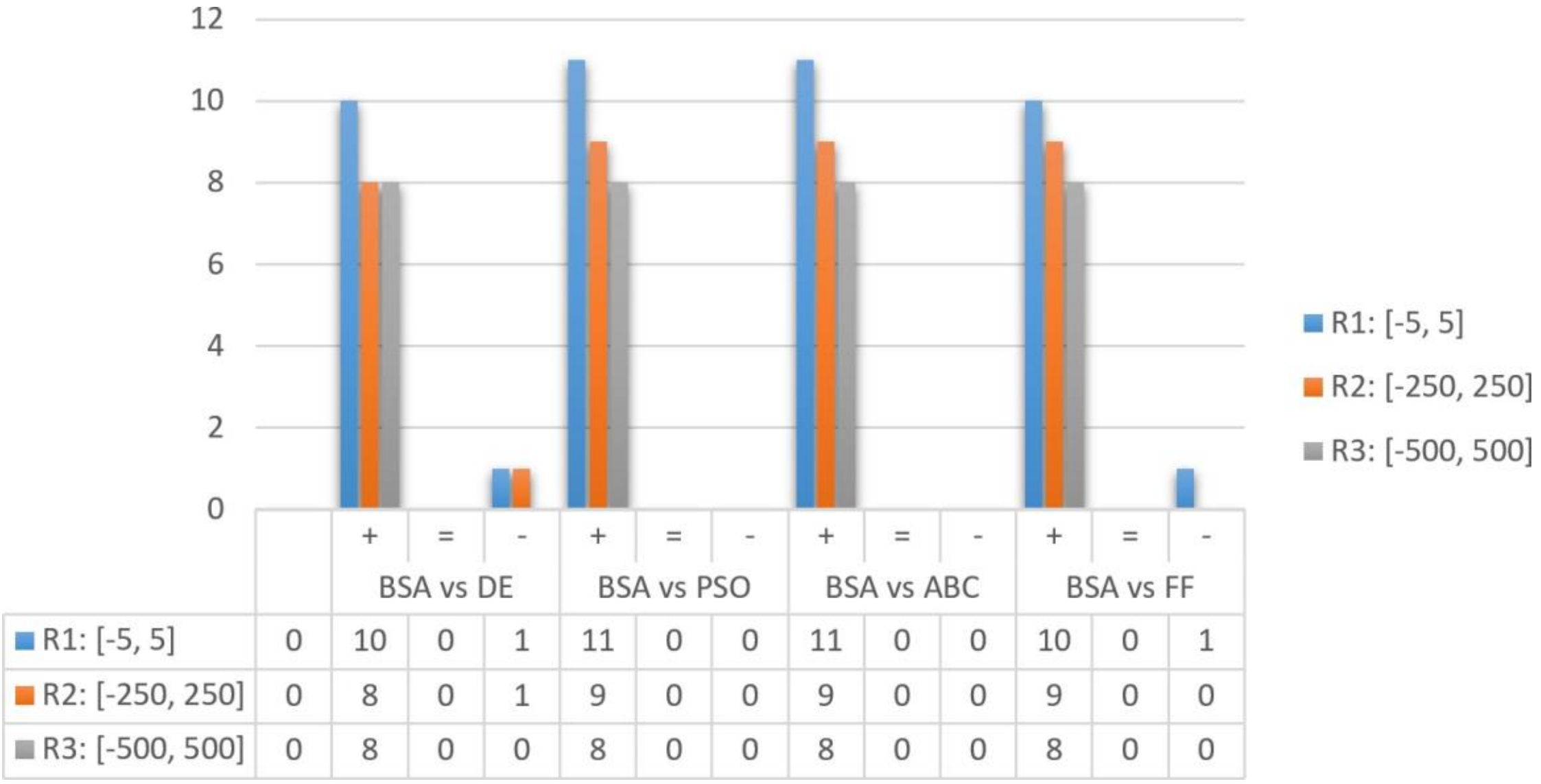

**Fig. 10: A graphical representation of Test 2**

Referring to Test 1 and 2, none of the algorithms was successful in solving all benchmark problems. In regard with Test 3, the ratio of successful minimisation of all sixteen benchmark functions varies in Nvar 1, 2, and 3 with default search space, and in two variable dimensions with three different search spaces (R1, R2, and R3). As a more illustration, Fig. 11 depicts the success and failure ratio for minimising the sixteen benchmark functions in Test 1.

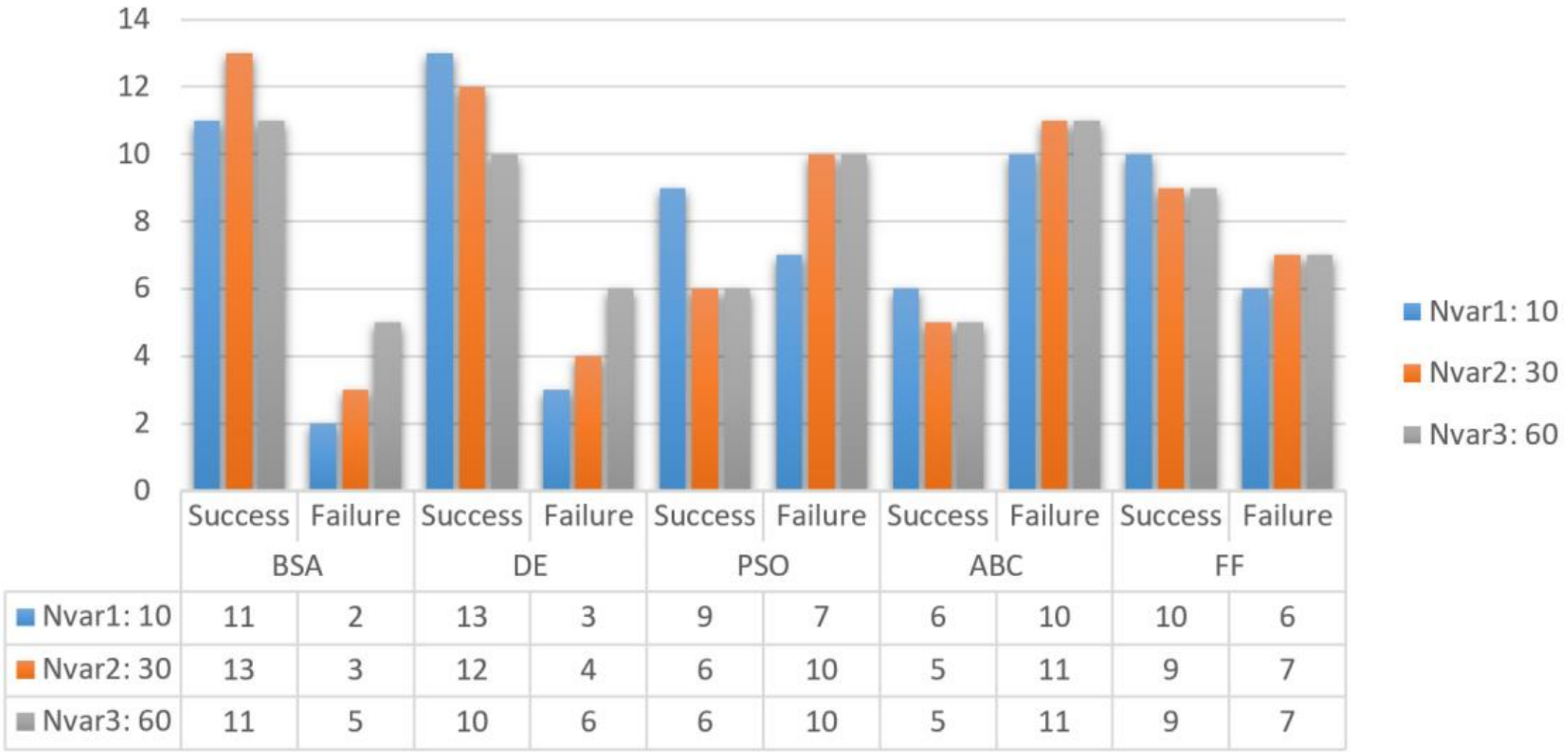

**Fig. 11: The success and failure ratio for minimising the sixteen benchmark functions in Test 1**

Regarding Tests 1 and 2, it was seen that none of the algorithms was successful in solving all the benchmark problems. Regarding Test 3, the ratio of successful minimisation of all sixteen benchmark functions varied in Nvar1, 2, and 3 with the default search space, and in two variable dimensions with three different search spaces (R1, R2, and R3). Fig. 11 and 12 depict the success and failure ratios in minimising the 16 benchmark functions in Tests 1 and 2, respectively.

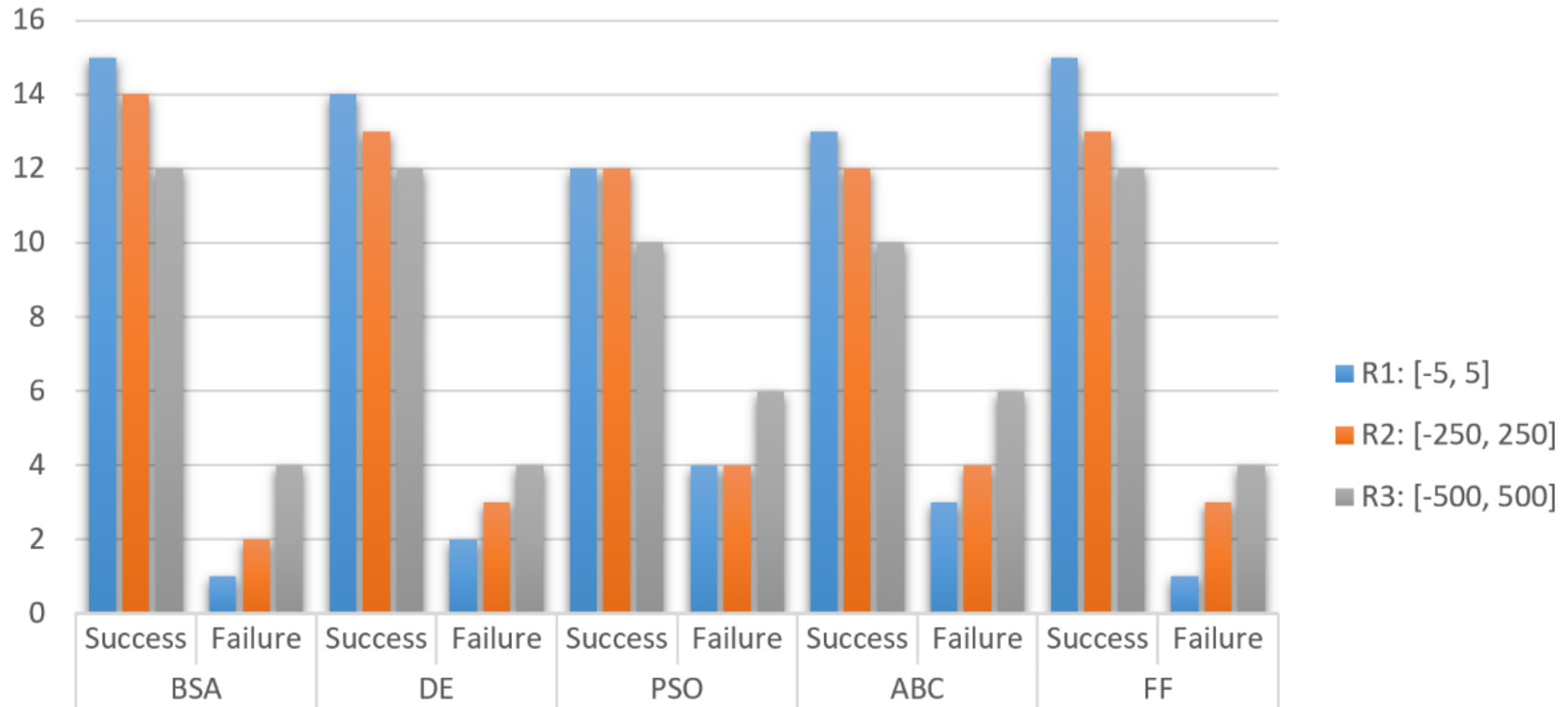

| | BSA Success | BSA Failure | DE Success | DE Failure | PSO Success | PSO Failure | ABC Success | ABC Failure | FF Success | FF Failure |
|---|---|---|---|---|---|---|---|---|---|---|
| R1: [-5, 5] | 15 | 1 | 14 | 2 | 12 | 4 | 13 | 3 | 15 | 1 |
| R2: [-250, 250] | 14 | 2 | 13 | 3 | 12 | 4 | 12 | 4 | 13 | 3 |
| R3: [-500, 500] | 12 | 4 | 12 | 4 | 10 | 6 | 10 | 6 | 12 | 4 |

**Fig. 12: The success and failure ratio in minimising the sixteen benchmark functions in Test 2**

Test 3 revealed that BSA was marginally the most successful algorithm in minimizing the maximum number of the given 16 benchmark functions, whereas ABC was the least successful in minimising the same. It also revealed that BSA could minimise the maximum number of optimisation problems; however, the minimum ratio of successful minimisation of F1-F16 functions in Nvar1, 2, 3 variables with the default search space was seen for ABC followed by PSO, whereas the maximum ratio of successful minimisation of these optimisation problems with variable size of two with three different search spaces (R1, 2, and 3) was seen for both ABC and PSO, followed by DE.

## 8. Contributions and Limitations of BSA

Literature discusses the positives and negatives of BSA along with those of the other meta-heuristic algorithms. Certain factors that could be responsible for higher success of BSA compared to other EAs are as follows:

1) The trial populations are produced very efficiently in each generation by using BSA’s crossover and mutation operators.
2) BSA balances the relationship between exploration (global search) and exploitation (local search) appropriately owing to control parameter, *F*. This parameter controls the direction of the local and global searches in an efficient and balanced manner. This balance can produce a wide range of numerical values necessary for the local and global searches. An extensive range of numerical values is produced for the global search, whereas a small range is produced for the local search. This helps BSA to solve a variety of problems.
3) The search direction of BSA is calculated by using the historical population *(oldP)*. In other words, the historical populations that are used in more capable individuals are relative to those that are used in older generations. This operation helps BSA to generate the trial individuals efficiently.
4) BSA’s crossover is a complex process compared to DE’s crossover. In each generation, the crossover operator in BSA creates new trial individuals to enhance BSA’s ability to deal with the problems.

5) Population diversity can be effectively achieved in BSA by a boundary control mechanism to ensure efficient searches.
6) Unlike other meta-heuristic algorithms, BSA uses historical and current populations; therefore, it is called a dual-population algorithm.
7) The simplicity of BSA may facilitate its usage by researchers and professionals.

To the best of our knowledge, no clear and precise drawbacks of BSA were explicitly addressed in the literature. Nonetheless, scholars have attempted to modify and hybridise the standard version of BSA as a compromise to further improve its efficiency and effectiveness to maximise its application domain. These attempts imply that researchers implicitly pointed out some drawbacks of BSA. Similar to the other meta-heuristic algorithms, BSA has some drawbacks that are listed below:

1) BSA requires memory space for storing historical populations. It is a dual- population algorithm because it utilises both historical and current populations. In each iteration, the old populations are stored on the stack for use in the next iterations. We entirely support the notion of employing a lexical database as the historical population of BSA to store its historical population. This lexical database is beneficial to reduce the memory size.
2) Setting the initial population in BSA entirely depends on the randomisation. We could naively mitigate the un-typical random nature of BSA by taking an average run of the algorithm. Alternatively, we could effectively deal with the un-typical random efforts of BSA by using the initial population of another nature-inspired algorithm. BSA and cuckoo search (CS) are both population-based algorithms [208]; therefore, the initial population of BSA can be set using CS. This helps the new proposed algorithm to reach the convergence faster than the standard BSA.
3) In the literature, it is stated that BSA could be improved further [29,35,60,68,71,75,76,82]. The improvements can be related to efficiency, effectiveness, computing time, speed of convergence, and improvement of local and global searches. The reasons behind these issues are complex. One possible reason can be the control parameters of BSA. Another reason could be related to the reality that BSA may appropriately need to find the best relationship between the local (exploitation) and global searches (exploration) that will suit each type of application to converge the best solution quickly.
4) This study [60] claims three shortcomings of BSA: difficulty in adjusting control parameter F, lack of learning from optimal individuals, and easily falling into local optima. The same research states that BSA is effective in exploration (global optima), whereas it is relatively weak in exploitation (local optima). On this basis, the convergence speed may be affected. Based on our knowledge, we suggest modifying BSA to improve its convergence speed and other drawbacks by redesigning the amplitude control factor (F) based on Levy flights and dynamic penalty to balance the local and global optima. This balance decreases the number of iterations and speeds up the convergence of BSA. In a recent study [209], F is designed to balance the exploration and exploitation of the algorithm. We believe that learning knowledge from the optimal individuals in BSA is not an intricate task. Recently, a new algorithm based on BSA such that each individual learns knowledge by utilising the global and local best information and the better historical information was proposed [209].

5) Compared to other meta-heuristic algorithms such as GA and PSO, BSA has not been used many applications and a variety of fields. BSA has mostly been used in engineering and information technology but not in the fields of text mining, natural language processing, semantic web, health, and medical sciences. The limited popularity of BSA compared to GA and PSO is not related to any weaknesses of BSA. Instead, it can probably be attributed to the newness of BSA. Time may be a factor in revealing the uses and expansions of BSA.

## 9. Conclusions and future directions

### 9.1. Conclusions

This study tried to understand BSA and its advances and applications. In this study, initially, the standard version of BSA and its convergence analysis were explained. Further, the expansions of BSA in the literature were presented. Later, the initial implementation of BSA as an optimiser to optimise common problems and using it in real-world applications was detailed. Further, the statistical experiments conducted on BSA in the literature were analysed, and the performance of BSA was evaluated against its counterpart algorithms for different types of real-world problems. Further, a general and an operational framework of the main expansions and implementations of BSA were proposed. Further, experiments were set up to evaluate BSA with other competitive algorithms statistically. Finally, the main shortcomings and limitations of BSA were summarised. It was found from these frameworks that BSA can be improved by amending its parameters, combining it with some other techniques or meta-heuristic algorithms to improve its performance, and adapting it to minimise different optimisation problems. The experimental results in the literature showed the statistical success of BSA over the other competitive algorithms to minimise different types of numerical optimisation problems without being excessively sensitive to the problem dimensions. Although, it is found in the experiments conducted in this study that BSA has statistically superior performance compared to the other algorithms, it does not lead us to conclude that BSA can be used in all cohorts of optimisation problems in terms of hardness score, search space, and problem dimensions. BSA was not able to minimise all the 16 benchmark functions used in the experiment in this study. This failure can probably be attributed to four factors: different hardness scores of the problem and the variety of dimensions, search spaces, and cohorts of the problems. A clear conclusion cannot be drawn to determine the success and failure ratios of BSA in solving different types of problems with different hardness scores, problem dimensions, and search spaces. Therefore, the experiment in this study is an attempt to reveal that BSA is sensitive towards the hardness score of the optimisation problem, problem dimension and type, and search space of the problem to a certain extent.

### 9.2. Future directions

This study puts forth several advances in SI in identifying more research gaps and principles for conducting further studies and future research direction on BSA. Most importantly, this study has four-fold contribution in the direction of future research and development.

Firstly, the frameworks proposed in this study can guide researchers who are working on the improvement of BSA. Notably, the proposed operational framework presents a clear overview of the main expansions and implementations of BSA and the relationships among them. The primary role of these frameworks is to act as a useful guide on the improved areas of BSA, and the research gaps can represent topics for future studies, in which scholars could effectively follow the procedures and skeletons of the proposed frameworks as footprints for further expansions and uses of BSA. The secondary role of the proposed frameworks is to outline the main extensions of BSA.

Secondly, it was seen that BSA was mostly used in the engineering fields [21], particularly in the areas of electrical and mechanical engineering and information and communication engineering. Furthermore, the standard BSA has substantially expanded in the context of hybridisation, modification, and extension to increase its application domain and improve its suitability for different types of optimisation problems. Significantly, BSA has been substantially hybridised and modified for different purposes, and it has been most extended for optimising multi-objective problems. These indications demonstrated the maturity of BSA in its use in the engineering fields and hybridisation with other techniques. Probably, the future of BSA depends on its use in medicine and pure sciences. Designing a better alternative EA based on the operators of BSA may replace further extensions of this approach. In brief, extensions of BSA can either be in a different direction than the current one or could be a motive to introduce a new and effective algorithm.

Thirdly, the experiments conducted in this study have a role of revealing the sensitivity of BSA towards the hardness score of the optimisation problem, problem dimension and type, and search space of the problem to a certain extent. Accordingly, BSA may be applicable to minimise combinatorial and large-scale optimisation problems appropriately. Although BSA has focused on small and medium-scale problems, it could be beneficial if further research focuses on minimising large-scale and combinatorial optimisation problems.

Lastly, we suggest conducting more studies in the area of tackling the limitations of BSA. A lexical database such as WordNet can be employed as the historical population of BSA in the text mining field to reduce the memory size for storing the historical population. On the contrary, combining BSA with other machine learning techniques can produce a better algorithm. It is foreseen that further advances in this area will lead to more generic and robust algorithms that would be mainly useful for dynamic optimisation problems. Specifically, the evolutionary operators of BSA can be coupled with machine learning techniques to outperform the standard BSA and the current methods of data mining [210–212]. However, the simplicity of BSA has reached a contradictory conclusion. Simplicity may facilitate the use of BSA by researchers and professionals; however, this simplicity brings the following paradoxical thought—whether or not BSA is technically effective and efficient as the other EAs. Based on this context, scholars may reluctantly use BSA for large-scale and combinatorial optimisation problems. Also, the performance of BSA can be further evaluated against other popular algorithms, such as Whale Optimisation Algorithm [213], Donkey and Smuggler Optimisation Algorithm [214], Fitness Dependent Optimiser [215], and Modified Grey Wolf Optimiser [216]. Nonetheless, the simplicity of BSA will not impede its uses and expansions. As described in this study, the studies conducted on BSA conclude that BSA has mostly been used in engineering optimisation problems. Despite the current use of BSA in a limited number of applications, a further research direction may address the issue of application of BSA to a large number of optimisation problems. Therefore, further developments in application-based BSA algorithms can certainly be seen in the future.

Acknowledgements

The authors would like to thank the referees for their remarkable suggestions. This paper's technical content has significantly improved based on their suggestions.

## Compliance with Ethical Standards

**Conflict of interest:** The authors declare that they have no conflict of interest.

**Funding:** Funding information is not applicable / No funding was received.

## References

[1] B.L. Agarwal, Basic statistics, New Age International, 2006.

[2] K.E. Voges, N. Pope, Business applications and computational intelligence, Igi Global, 2006.

[3] J. Kacprzyk, W. Pedrycz, Springer handbook of computational intelligence, Springer, 2015.

[4] G. Beni, J. Wang, Swarm intelligence in cellular robotic systems, in: Robot. Biol. Syst. Towar. a New Bionics?, Springer, 1993: pp. 703–712.

[5] V. Kothari, J. Anuradha, S. Shah, P. Mittal, A survey on particle swarm optimization in feature selection, in: Glob. Trends Inf. Syst. Softw. Appl., Springer, 2012: pp. 192–201.

[6] A.K. Kar, Bio inspired computing–A review of algorithms and scope of applications, Expert Syst. Appl. 59 (2016) 20–32.

[7] E. Falkenauer, Genetic algorithms and grouping problems, Wiley New York, 1998.

[8] R. Storn, K. Price, Differential evolution–a simple and efficient heuristic for global optimization over continuous spaces, J. Glob. Optim. 11 (1997) 341–359.

[9] R.J. Mullen, D. Monekosso, S. Barman, P. Remagnino, A review of ant algorithms, Expert Syst. Appl. 36 (2009) 9608–9617.

[10] R. Eberhart, J. Kennedy, A new optimizer using particle swarm theory, in: Micro Mach. Hum. Sci. 1995. MHS'95., Proc. Sixth Int. Symp., IEEE, 1995: pp. 39–43.

[11] P. Civicioglu, Backtracking search optimization algorithm for numerical optimization problems, Appl. Math. Comput. 219 (2013) 8121–8144.

[12] D. Ghosh, U. Bhaduri, A simple recursive backtracking algorithm for knight's tours puzzle on standard 8× 8 chessboard, in: Adv. Comput. Commun. Informatics (ICACCI), 2017 Int. Conf., IEEE, 2017: pp. 1195–1200.

[13] S. Güldal, V. Baugh, S. Allehaibi, N-Queens solving algorithm by sets and backtracking, in: SoutheastCon, 2016, IEEE, 2016: pp. 1–8.

[14] I. Kuswardayan, N. Suciati, Design and Implementation of Random Word Generator using Backtracking Algorithm for Gameplay in Ambrosia Game, Int. J. Comput. Appl. 158 (2017).

[15] S. Mukherjee, S. Datta, P.B. Chanda, P. Pathak, Comparative Study of Different Algorithms To

Solve N-Queens Problem, Int. J. Found. Comput. Sci. Technol. 5 (2015) 15–27.

[16] B. Zhichao, Solution for backtracking based on maze problem and algorithm realization [J], Electron. Test. 14 (2013) 83.

[17] D. Carlson, Sophomores meet the traveling salesperson problem, J. Comput. Sci. Coll. 33 (2018) 126–133.

[18] B. Korte, J. Vygen, Kombinatorische Optimierung: Theorie und Algorithmen, Springer-Verlag, 2012.

[19] N. Honda, Backtrack beam search for multiobjective scheduling problem, in: Multi-Objective Program. Goal Program., Springer, 2003: pp. 147–152.

[20] C. Lu, L. Gao, X. Li, Q. Pan, Q. Wang, Energy-efficient permutation flow shop scheduling problem using a hybrid multi-objective backtracking search algorithm, J. Clean. Prod. 144 (2017) 228–238.

[21] T. Dede, M. Kripka, V. Togan, V. Yepes, Usage of Optimization Techniques in Civil Engineering During the Last Two Decades, Curr. Trends Civ. Struct. Eng. 2 (2019) 1–17.

[22] P. Civicioglu, Artificial cooperative search algorithm for numerical optimization problems, Inf. Sci. (Ny). 229 (2013) 58–76.

[23] K. DUSKO, Backtracking Tutorial using C Program Code Example for Programmers, (2014). https://www.thegeekstuff.com/2014/12/backtracking-example/.

[24] Huybers, Backtracking is an implementation of Artificial Intelligence, (n.d.). http://www.huybers.net/backtrack/backe.html (accessed October 6, 2018).

[25] P. Civicioglu, E. Besdok, A+ Evolutionary search algorithm and QR decomposition based rotation invariant crossover operator, Expert Syst. Appl. 103 (2018) 49–62.

[26] Y. Sheoran, V. Kumar, K.P.S. Rana, P. Mishra, J. Kumar, S.S. Nair, Development of backtracking search optimization algorithm toolkit in LabViewTM, Procedia Comput. Sci. 57 (2015) 241–248.

[27] X.-S. Yang, X.-S. He, Mathematical foundations of nature-inspired algorithms, Springer, 2019.

[28] R.R. Sharapov, A. V Lapshin, Convergence of genetic algorithms, Pattern Recognit. Image Anal. 16 (2006) 392–397.

[29] L. Zhao, Z. Jia, L. Chen, Y. Guo, Improved Backtracking Search Algorithm Based on Population Control Factor and Optimal Learning Strategy, Math. Probl. Eng. 2017 (2017).

[30] S. Nama, A.K. Saha, S. Ghosh, Improved backtracking search algorithm for pseudo dynamic active earth pressure on retaining wall supporting c-Φ backfill, Appl. Soft Comput. 52 (2017) 885–897.

[31] P.N. Agrawal, R.N. Mohapatra, U. Singh, H.M. Srivastava, Mathematical Analysis and its Applications, Springer, 2015.

[32] L. Qi, Convergence analysis of some algorithms for solving nonsmooth equations, Math. Oper. Res. 18 (1993) 227–244.

[33] D. Lahaye, J. Tang, K. Vuik, Modern solvers for Helmholtz problems, Springer, 2017.

[34] S.P. Brooks, A. Gelman, General methods for monitoring convergence of iterative simulations, J. Comput. Graph. Stat. 7 (1998) 434–455.

[35] H. Wang, Z. Hu, Y. Sun, Q. Su, X. Xia, Modified Backtracking Search Optimization Algorithm Inspired by Simulated Annealing for Constrained Engineering Optimization Problems, Comput. Intell. Neurosci. 2018 (2018).

[36] M.S. Ahmed, A. Mohamed, T. Khatib, H. Shareef, R.Z. Homod, J.A. Ali, Real time optimal schedule controller for home energy management system using new binary backtracking search algorithm, Energy Build. 138 (2017) 215–227.

[37] M. Akhtar, M.A. Hannan, R.A. Begum, H. Basri, E. Scavino, Backtracking search algorithm in CVRP models for efficient solid waste collection and route optimization, Waste Manag. 61 (2017) 117–128.

[38] H. Wang, Z. Hu, Y. Sun, Q. Su, X. Xia, A novel modified BSA inspired by species evolution rule and simulated annealing principle for constrained engineering optimization problems, Neural Comput. Appl. (n.d.) 1–28.

[39] D. Chen, F. Zou, R. Lu, P. Wang, Learning backtracking search optimisation algorithm and its application, Inf. Sci. (Ny). 376 (2017) 71–94.

[40] J. Kartite, M. Cherkaoui, Improved backtracking search algorithm for renewable energy system, Energy Procedia. 141 (2017) 126–130.

[41] H.-C. Tsai, Improving backtracking search algorithm with variable search strategies for continuous optimization, Appl. Soft Comput. 80 (2019) 567–578.

[42] J. Lu, J. Ding, Construction of prediction intervals for carbon residual of crude oil based on deep stochastic configuration networks, Inf. Sci. (Ny). 486 (2019) 119–132.

[43] D. Chen, R. Lu, F. Zou, S. Li, P. Wang, A learning and niching based backtracking search optimisation algorithm and its applications in global optimisation and ANN training, Neurocomputing. 266 (2017) 579–594.

[44] Y.Ç. Kuyu, F. Vatansever, The Chaos-Based Approaches for Actual Metaheuristic Algorithms, Uludağ Univ. J. Fac. Eng. 23 (n.d.) 103–116.

[45] A. Chatzipavlis, G.E. Tsekouras, V. Trygonis, A.F. Velegrakis, J. Tsimikas, A. Rigos, T. Hasiotis, C. Salmas, Modeling beach realignment using a neuro-fuzzy network optimized by a novel backtracking search algorithm, Neural Comput. Appl. 31 (2019) 1747–1763.

[46] K. Yu, J.J. Liang, B.Y. Qu, Z. Cheng, H. Wang, Multiple learning backtracking search algorithm for estimating parameters of photovoltaic models, Appl. Energy. 226 (2018) 408–422.

[47] M.A. Ahandani, A.R. Ghiasi, H. Kharrati, Parameter identification of chaotic systems using a shuffled backtracking search optimization algorithm, Soft Comput. 22 (2018) 8317–8339.

[48] H. Yang, J. Yu, Y. Qiu, Q. Li, W. Chen, A coordinated optimization method considering time-delay effect of islanded photovoltaic microgrid based on modified backtracking search algorithm, J. Renew. Sustain. Energy. 10 (2018) 23503.

[49] J.A. Ali, M.A. Hannan, A. Mohamed, Backtracking search algorithm approach to improve indirect field-oriented control for induction motor drive, in: 2015 IEEE 3rd Int. Conf. Smart Instrumentation, Meas. Appl., IEEE, 2015: pp. 1–6.

[50] Q. Xu, L. Guo, N. Wang, L. Xu, Opposition-based backtracking search algorithm for numerical optimization problems, in: Int. Conf. Intell. Sci. Big Data Eng., Springer, 2015: pp. 223–234.

[51] J. Lin, Oppositional backtracking search optimization algorithm for parameter identification of hyperchaotic systems, Nonlinear Dyn. 80 (2015) 209–219.

[52] M. Brévilliers, O. Abdelkafi, J. Lepagnot, L. Idoumghar, Idol-guided backtracking search optimization algorithm, in: 12th Int. Conf. Artif. Evol. (EA 2015), Lyon, Fr., 2015.

[53] L.A. Passos, D. Rodrigues, J.P. Papa, Quaternion-Based Backtracking Search Optimization Algorithm, in: 2019 IEEE Congr. Evol. Comput., IEEE, 2019: pp. 3014–3021.

[54] S. Vitayasak, P. Pongcharoen, C. Hicks, A tool for solving stochastic dynamic facility layout problems with stochastic demand using either a Genetic Algorithm or modified Backtracking Search Algorithm, Int. J. Prod. Econ. 190 (2017) 146–157.

[55] H. Ferradi, R. Géraud, D. Maimuţ, D. Naccache, H. Zhou, Backtracking-assisted multiplication, Cryptogr. Commun. 10 (2018) 17–26.

[56] M. Li, D. Lin, J. Kou, A hybrid niching PSO enhanced with recombination-replacement crowding strategy for multimodal function optimization, Appl. Soft Comput. 12 (2012) 975–987.

[57] E.L. Yu, P.N. Suganthan, Ensemble of niching algorithms, Inf. Sci. (Ny). 180 (2010) 2815–2833.

[58] B. Costin, B. Amelia, SOLVING COMBINATORIAL OPTIMISATION PROBLEMS USING SIMULATED ANNEALING, (n.d.).

[59] X. Yuan, X. Wu, H. Tian, Y. Yuan, R.M. Adnan, Parameter identification of nonlinear Muskingum model with backtracking search algorithm, Water Resour. Manag. 30 (2016) 2767–2783.

[60] F. Zou, D. Chen, R. Lu, Hybrid Hierarchical Backtracking Search Optimization Algorithm and Its Application, Arab. J. Sci. Eng. 43 (2018) 993–1014.

[61] A.F. Ali, A memetic backtracking search optimization algorithm for economic dispatch problem, Egypt. Comput. Sci. J. 39 (2015) 56–71.

[62] M.A. Hannan, M.S.H. Lipu, A. Hussain, M.H. Saad, A. Ayob, Neural network approach for estimating state of charge of lithium-ion battery using backtracking search algorithm, IEEE Access. 6 (2018) 10069–10079.

[63] M.S.H. Lipu, A. Hussain, M.H.M. Saad, M.A. Hannan, Optimal neural network approach for estimating state of energy of lithium-ion battery using heuristic optimization techniques, in: Electr. Eng. Informatics (ICEEI), 2017 6th Int. Conf., IEEE, 2017: pp. 1–6.

[64] Q. Lin, L. Gao, X. Li, C. Zhang, A hybrid backtracking search algorithm for permutation flow-shop scheduling problem, Comput. Ind. Eng. 85 (2015) 437–446.

[65] P. Chen, L. Wen, R. Li, X. Li, A hybrid backtracking search algorithm for permutation flow-shop scheduling problem minimizing makespan and energy consumption, in: 2017 IEEE Int. Conf. Ind. Eng. Eng. Manag., IEEE, 2017: pp. 1611–1615.

[66] S. Nama, A. Saha, A novel hybrid backtracking search optimization algorithm for continuous function optimization, Decis. Sci. Lett. 8 (2019) 163–174.

[67] S. Wang, X. Da, M. Li, T. Han, Adaptive backtracking search optimization algorithm with pattern search for numerical optimization, J. Syst. Eng. Electron. 27 (2016) 395–406.

[68] L. Wang, Y. Zhong, Y. Yin, W. Zhao, B. Wang, Y. Xu, A hybrid backtracking search optimization algorithm with differential evolution, Math. Probl. Eng. 2015 (2015).

[69] S. Das, D. Mandal, R. Kar, S. Prasad Ghoshal, A new hybridized backtracking search optimization algorithm with differential evolution for sidelobe suppression of uniformly excited concentric circular antenna arrays, Int. J. RF Microw. Comput. Eng. 25 (2015) 262–268.

[70] M. Brévilliers, O. Abdelkafi, J. Lepagnot, L. Idoumghar, Fast Hybrid BSA-DE-SA Algorithm on GPU, in: Int. Conf. Swarm Intell. Based Optim., Springer, 2016: pp. 75–86.

[71] Z. Su, H. Wang, P. Yao, A hybrid backtracking search optimization algorithm for nonlinear optimal control problems with complex dynamic constraints, Neurocomputing. 186 (2016) 182–194.

[72] A. Askarzadeh, L. dos Santos Coelho, A backtracking search algorithm combined with Burger's chaotic map for parameter estimation of PEMFC electrochemical model, Int. J. Hydrogen Energy. 39 (2014) 11165–11174.

[73] S.K. Agarwal, S. Shah, R. Kumar, Classification of mental tasks from EEG data using backtracking search optimization based neural classifier, Neurocomputing. 166 (2015) 397–403.

[74] J.A. Ali, M.A. Hannan, A. Mohamed, M.G.M. Abdolrasol, Fuzzy logic speed controller optimization approach for induction motor drive using backtracking search algorithm, Measurement. 78 (2016) 49–62.

[75] P.S. Pal, R. Kar, D. Mandal, S.P. Ghoshal, A hybrid backtracking search algorithm with wavelet

mutation-based nonlinear system identification of Hammerstein models, Signal, Image Video Process. 11 (2017) 929–936.

[76] W. Zhao, L. Wang, B. Wang, Y. Yin, Best guided backtracking search algorithm for numerical optimization problems, in: Int. Conf. Knowl. Sci. Eng. Manag., Springer, 2016: pp. 414–425.

[77] S. Pare, A.K. Bhandari, A. Kumar, V. Bajaj, Backtracking search algorithm for color image multilevel thresholding, Signal, Image Video Process. 12 (2018) 385–392.

[78] G. Toz, İ. Yücedağ, P. Erdoğmuş, A fuzzy image clustering method based on an improved backtracking search optimization algorithm with an inertia weight parameter, J. King Saud Univ. Inf. Sci. (2018).

[79] O.E. Turgut, Thermal and Economical Optimization of a Shell and Tube Evaporator Using Hybrid Backtracking Search—Sine–Cosine Algorithm, Arab. J. Sci. Eng. 42 (2017) 2105–2123.

[80] J. Lin, Z.-J. Wang, X. Li, A backtracking search hyper-heuristic for the distributed assembly flow-shop scheduling problem, Swarm Evol. Comput. 36 (2017) 124–135.

[81] J. Lin, Backtracking search based hyper-heuristic for the flexible job-shop scheduling problem with fuzzy processing time, Eng. Appl. Artif. Intell. 77 (2019) 186–196.

[82] H. Ao, T.N. Thoi, V.H. Huu, L. Anh-Le, T. Nguyen, M.Q. Chau, Backtracking search optimization algorithm and its application to roller bearing fault diagnosis, Int. J. Acoust. Vib. 21 (2016).

[83] S. Nama, A.K. Saha, A new hybrid differential evolution algorithm with self-adaptation for function optimization, Appl. Intell. 48 (2018) 1657–1671.

[84] S. Nama, A. Saha, S. Ghosh, A new ensemble algorithm of differential evolution and backtracking search optimization algorithm with adaptive control parameter for function optimization, Int. J. Ind. Eng. Comput. 7 (2016) 323–338.

[85] W.U. Khan, Z. Ye, N.I. Chaudhary, M.A.Z. Raja, Backtracking search integrated with sequential quadratic programming for nonlinear active noise control systems, Appl. Soft Comput. 73 (2018) 666–683.

[86] M. Sriram, K. Ravindra, Backtracking Search Optimization Algorithm Based MPPT Technique for Solar PV System, in: Adv. Decis. Sci. Image Process. Secur. Comput. Vis., Springer, 2020: pp. 498–506.

[87] A. Gosain, K. Sachdeva, Selection of materialized views using stochastic ranking based Backtracking Search Optimization Algorithm, Int. J. Syst. Assur. Eng. Manag. (n.d.) 1–10.

[88] Z. Tian, Y. Ren, G. Wang, An application of backtracking search optimization–based least squares support vector machine for prediction of short-term wind speed, Wind Eng. (2019) 0309524X19849843.

[89] M. Konar, GAO Algoritma tabanlı YSA modeliyle İHA motorunun performansının ve uçuş süresinin maksimizasyonu, Avrupa Bilim ve Teknol. Derg. (n.d.) 360–367.

[90] Z. Wang, Y.-R. Zeng, S. Wang, L. Wang, Optimizing echo state network with backtracking search optimization algorithm for time series forecasting, Eng. Appl. Artif. Intell. 81 (2019) 117–132.

[91] V. Thai, J. Cheng, V. Nguyen, P. Daothi, Optimizing SVM's parameters based on backtracking search optimization algorithm for gear fault diagnosis, J. Vibroengineering. 21 (2019) 66–81.

[92] S. Sun, L. Wei, J. Xu, Z. Jin, A New Wind Speed Forecasting Modeling Strategy Using Two-Stage Decomposition, Feature Selection and DAWNN, Energies. 12 (2019) 334.

[93] D. Jia, Y. Tong, Y. Yu, Z. Cai, S. Gao, A Novel Backtracking Search with Grey Wolf Algorithm for Optimization, in: 2018 10th Int. Conf. Intell. Human-Machine Syst. Cybern., IEEE, 2018: pp. 73–76.

[94] Z. Xu, Z. Lei, L. Yang, X. Li, S. Gao, Negative Correlation Learning Enhanced Search Behavior in

Backtracking Search Optimization, in: 2018 10th Int. Conf. Intell. Human-Machine Syst. Cybern., IEEE, 2018: pp. 310–314.

[95] S. Yan, J. Zhou, Y. Zheng, C. Li, An improved hybrid backtracking search algorithm based T–S fuzzy model and its implementation to hydroelectric generating units, Neurocomputing. 275 (2018) 2066–2079.

[96] C. Zhang, C. Li, T. Peng, X. Xia, X. Xue, W. Fu, J. Zhou, Modeling and synchronous optimization of pump turbine governing system using sparse robust least squares support vector machine and hybrid backtracking search algorithm, Energies. 11 (2018) 3108.

[97] L. Chen, N. Sun, C. Zhou, J. Zhou, Y. Zhou, J. Zhang, Q. Zhou, Flood forecasting based on an improved extreme learning machine model combined with the backtracking search optimization algorithm, Water. 10 (2018) 1362.

[98] A. Pourdaryaei, H. Mokhlis, H.A. Illias, S.H.A. Kaboli, S. Ahmad, Short-Term Electricity Price Forecasting via Hybrid Backtracking Search Algorithm and ANFIS Approach, IEEE Access. 7 (2019) 77674–77691.

[99] J. Zhou, C. Zhang, T. Peng, Y. Xu, Parameter identification of pump turbine governing system using an improved backtracking search algorithm, Energies. 11 (2018) 1668.

[100] J. Zhou, N. Sun, B. Jia, T. Peng, A novel decomposition-optimization model for short-term wind speed forecasting, Energies. 11 (2018) 1752.

[101] W. Zhang, S. Zhang, S. Zhang, Two-factor high-order fuzzy-trend FTS model based on BSO-FCM and improved KA for TAIEX stock forecasting, Nonlinear Dyn. 94 (2018) 1429–1446.

[102] U.H. Atasever, C. Ozkan, A New SEBAL Approach Modified with Backtracking Search Algorithm for Actual Evapotranspiration Mapping and On-Site Application, J. Indian Soc. Remote Sens. 46 (2018) 1213–1222.

[103] G. Jothi, H.H. Inbarani, A.T. Azar, K.R. Devi, Rough set theory with Jaya optimization for acute lymphoblastic leukemia classification, Neural Comput. Appl. (2018) 1–20.

[104] H. Li, L. Pan, M. Chen, X. Chen, Y. Zhang, RBM-Based Back Propagation Neural Network with BSASA Optimization for Time Series Forecasting, in: 2017 9th Int. Conf. Intell. Human-Machine Syst. Cybern., IEEE, 2017: pp. 218–221.

[105] G. Mohy-ud-din, Hybrid dynamic economic emission dispatch of thermal, wind, and photovoltaic power using the hybrid backtracking search algorithm with sequential quadratic programming, J. Renew. Sustain. Energy. 9 (2017) 15502.

[106] K. Lenin, B. Ravindhranathreddy, M. Suryakalavathi, Hybridisation of backtracking search optimisation algorithm with differential evolution algorithm for solving reactive power problem, Int. J. Adv. Intell. Paradig. 8 (2016) 355–364.

[107] B. Wang, L. Wang, Y. Yin, Y. Xu, W. Zhao, Y. Tang, An improved neural network with random weights using backtracking search algorithm, Neural Process. Lett. 44 (2016) 37–52.

[108] S. Mallick, R. Kar, D. Mandal, S.P. Ghoshal, CMOS analogue amplifier circuits optimisation using hybrid backtracking search algorithm with differential evolution, J. Exp. Theor. Artif. Intell. 28 (2016) 719–749.

[109] B. Benhala, M. Kotti, A. Ahaitouf, M. Fakhfakh, Backtracking ACO for RF-circuit design optimization, in: Perform. Optim. Tech. Analog. Mix. Radio-Frequency Circuit Des., IGI Global, 2015: pp. 158–179.

[110] T. Aldowaisan, A. Allahverdi, New heuristics for no-wait flowshops to minimize makespan, Comput. Oper. Res. 30 (2003) 1219–1231.

[111] B. Liu, L. Wang, Y.-H. Jin, An effective PSO-based memetic algorithm for flow shop scheduling, IEEE Trans. Syst. Man, Cybern. Part B. 37 (2007) 18–27.

[112] Q.-K. Pan, M.F. Tasgetiren, Y.-C. Liang, A discrete differential evolution algorithm for the permutation flowshop scheduling problem, Comput. Ind. Eng. 55 (2008) 795–816.

[113] S. Das, P.N. Suganthan, Differential evolution: a survey of the state-of-the-art, IEEE Trans. Evol. Comput. 15 (2011) 4–31.

[114] P. Civicioglu, E. Besdok, A conceptual comparison of the Cuckoo-search, particle swarm optimization, differential evolution and artificial bee colony algorithms, Artif. Intell. Rev. 39 (2013) 315–346.

[115] J.X. Yu, X. Yao, C.-H. Choi, G. Gou, Materialized view selection as constrained evolutionary optimization, IEEE Trans. Syst. Man, Cybern. Part C (Applications Rev. 33 (2003) 458–467.

[116] J. Cheng, X. Wu, M. Zhou, S. Gao, Z. Huang, C. Liu, A novel method for detecting new overlapping community in complex evolving networks, IEEE Trans. Syst. Man, Cybern. Syst. (2018).

[117] S. Gao, R.-L. Wang, H. Tamura, Z. Tang, A multi-layered immune system for graph planarization problem, IEICE Trans. Inf. Syst. 92 (2009) 2498–2507.

[118] Y. Zhou, J. Wang, J. Chen, S. Gao, L. Teng, Ensemble of many-objective evolutionary algorithms for many-objective problems, Soft Comput. 21 (2017) 2407–2419.

[119] S. Song, S. Gao, X. Chen, D. Jia, X. Qian, Y. Todo, AIMOES: Archive information assisted multi-objective evolutionary strategy for ab initio protein structure prediction, Knowledge-Based Syst. 146 (2018) 58–72.

[120] J. Ji, S. Gao, J. Cheng, Z. Tang, Y. Todo, An approximate logic neuron model with a dendritic structure, Neurocomputing. 173 (2016) 1775–1783.

[121] T. Zhou, S. Gao, J. Wang, C. Chu, Y. Todo, Z. Tang, Financial time series prediction using a dendritic neuron model, Knowledge-Based Syst. 105 (2016) 214–224.

[122] M.G.M. Abdolrasol, M.A. Hannan, A. Mohamed, U.A.U. Amiruldin, I.B.Z. Abidin, M.N. Uddin, An Optimal Scheduling Controller for Virtual Power Plant and Microgrid Integration Using the Binary Backtracking Search Algorithm, IEEE Trans. Ind. Appl. 54 (2018) 2834–2844.

[123] H. Zhao, F. Min, W. Zhu, A backtracking approach to minimal cost feature selection of numerical data, J. Inf. &COMPUTATIONAL Sci. 10 (2013) 4105–4115.

[124] H. Zhao, F. Min, W. Zhu, Cost-sensitive feature selection of numeric data with measurement errors, J. Appl. Math. 2013 (2013).

[125] P. Civicioglu, Circular antenna array design by using evolutionary search algorithms, Prog. Electromagn. Res. 54 (2013) 265–284.

[126] A.M. SHAHEEN, R.A. EL-SEHIEMY, Binary and Integer Coded Backtracking Search Optimization Algorithm for Transmission Network Expansion Planning, (n.d.).

[127] C. Zhang, Q. Lin, L. Gao, X. Li, Backtracking Search Algorithm with three constraint handling methods for constrained optimization problems, Expert Syst. Appl. 42 (2015) 7831–7845.

[128] F. Zou, D. Chen, S. Li, R. Lu, M. Lin, Community detection in complex networks: Multi-objective discrete backtracking search optimization algorithm with decomposition, Appl. Soft Comput. 53 (2017) 285–295.

[129] M. Modiri-Delshad, N.A. Rahim, Multi-objective backtracking search algorithm for economic emission dispatch problem, Appl. Soft Comput. 40 (2016) 479–494.

[130] K. Bhattacharjee, A. Bhattacharya, S.H. nee Dey, Backtracking search optimization based economic environmental power dispatch problems, Int. J. Electr. Power Energy Syst. 73 (2015) 830–842.

[131] R. El Maani, B. Radi, A. El Hami, Multiobjective backtracking search algorithm: application to FSI, Struct. Multidiscip. Optim. 59 (2019) 131–151.

[132] A.T. Zeine, A. El Hami, R. Ellaia, E. Pagnacco, Backtracking search algorithm for multi-objective design optimisation, Int. J. Math. Model. Numer. Optim. 8 (2017) 93–107.

[133] C. Lu, L. Gao, X. Li, Q. Wang, W. Liao, Q. Zhao, An efficient multiobjective backtracking search algorithm for single machine scheduling with controllable processing times, Math. Probl. Eng. 2017 (2017).

[134] F. Daqaq, R. Ellaia, M. Ouassaid, Multiobjective backtracking search algorithm for solving optimal power flow, in: 2017 Int. Conf. Electr. Inf. Technol., IEEE, 2017: pp. 1–6.

[135] C. Lu, L. Gao, X. Li, P. Chen, Energy-efficient multi-pass turning operation using multi-objective backtracking search algorithm, J. Clean. Prod. 137 (2016) 1516–1531.

[136] M.Z. bin Mohd Zain, J. Kanesan, G. Kendall, J.H. Chuah, Optimization of fed-batch fermentation processes using the Backtracking Search Algorithm, Expert Syst. Appl. 91 (2018) 286–297.

[137] K.H. Rosen, K. Krithivasan, Discrete mathematics and its applications: with combinatorics and graph theory, Tata McGraw-Hill Education, 2012.

[138] H. El Sakkout, M. Wallace, Probe backtrack search for minimal perturbation in dynamic scheduling, Constraints. 5 (2000) 359–388.

[139] J. Yadav, J. Chandel, N. Gupta, Personnel Scheduling: Comparative Study of Backtracking Approaches and Genetic Algorithms, (2015).

[140] V. Kumar, K.P.S. Rana, D. Kler, Efficient control of a 3-link planar rigid manipulator using self-regulated fractional-order fuzzy PID controller, Appl. Soft Comput. (2019) 105531.

[141] D. Guha, P.K. Roy, S. Banerjee, Application of backtracking search algorithm in load frequency control of multi-area interconnected power system, Ain Shams Eng. J. 9 (2018) 257–276.

[142] H. Boudjefdjouf, H.R.E.H. Bouchekara, R. Mehasni, M.K. Smail, A. Orlandi, F. de Paulis, Wire fault diagnosis using time-domain reflectometry and backtracking search optimization algorithm, in: 2015 31st Int. Rev. Prog. Appl. Comput. Electromagn., IEEE, 2015: pp. 1–2.

[143] A.O. de Sá, N. Nedjah, L. de M. Mourelle, Genetic and backtracking search optimisation algorithms applied to localisation problems, Int. J. Innov. Comput. Appl. 6 (2015) 223–228.

[144] A. Mehmood, N.I. Chaudhary, A. Zameer, M.A.Z. Raja, Backtracking search optimization heuristics for nonlinear Hammerstein controlled auto regressive auto regressive systems, ISA Trans. (2019).

[145] V. Goyal, P. Mishra, A. Shukla, V.K. Deolia, A. Varshney, A fractional order parallel control structure tuned with meta-heuristic optimization algorithms for enhanced robustness, J. Electr. Eng. 70 (2019) 16–24.

[146] T.X. Dinh, N.P. Luan, K.K. Ahn, A novel inverse modeling control for piezo positioning stage, J. Mech. Sci. Technol. 32 (2018) 5875–5888.

[147] V. Goyal, P. Mishra, V.K. Deolia, A robust fractional order parallel control structure for flow control using a pneumatic control valve with nonlinear and uncertain dynamics, Arab. J. Sci. Eng. 44 (2019) 2597–2611.

[148] A. Mehmood, A. Zameer, N.I. Chaudhary, M.A.Z. Raja, Backtracking search heuristics for identification of electrical muscle stimulation models using Hammerstein structure, Appl. Soft Comput. (2019) 105705.

[149] A. El-Fergany, Multi-objective allocation of multi-type distributed generators along distribution networks using backtracking search algorithm and fuzzy expert rules, Electr. Power Components Syst. 44 (2016) 252–267.

[150] K. Guney, A. Durmus, Pattern nulling of linear antenna arrays using backtracking search optimization algorithm, Int. J. Antennas Propag. 2015 (2015).

[151] A.E. Chaib, H. Bouchekara, R. Mehasni, M.A. Abido, Optimal power flow with emission and non-smooth cost functions using backtracking search optimization algorithm, Int. J. Electr. Power Energy Syst. 81 (2016) 64–77.

[152] K. Bhattacharjee, Economic Dispatch Problems Using Backtracking Search Optimization, Int. J. Energy Optim. Eng. 7 (2018) 39–60.

[153] M. Shafiullah, M.A. Abido, L.S. Coelho, Design of robust PSS in multimachine power systems using backtracking search algorithm, in: Intell. Syst. Appl. to Power Syst. (ISAP), 2015 18th Int. Conf., IEEE, 2015: pp. 1–6.

[154] R.-E. Precup, A.-D. Balint, M.-B. Radac, E.M. Petriu, Backtracking Search Optimization Algorithm-based approach to PID controller tuning for torque motor systems, in: Syst. Conf. (SysCon), 2015 9th Annu. IEEE Int., IEEE, 2015: pp. 127–132.

[155] C. Zhang, J. Zhou, C. Li, W. Fu, T. Peng, A compound structure of ELM based on feature selection and parameter optimization using hybrid backtracking search algorithm for wind speed forecasting, Energy Convers. Manag. 143 (2017) 360–376.

[156] U. Kılıç, Backtracking search algorithm-based optimal power flow with valve point effect and prohibited zones, Electr. Eng. 97 (2015) 101–110.

[157] A.M. Shaheen, R.A. El-Sehiemy, S.M. Farrag, Optimal reactive power dispatch using backtracking search algorithm, Aust. J. Electr. Electron. Eng. 13 (2016) 200–210.

[158] D.S.K. Kanth, N.S.R. Reddy, R.S.G. Reddy, Optimal placement & sizing of DG's using backtracking search algorithm in IEEE 33-bus distribution system, in: Comput. Methodol. Commun. (ICCMC), 2017 Int. Conf., IEEE, 2017: pp. 163–169.

[159] P. Gupta, V. Kumar, K.P.S. Rana, P. Mishra, Comparative study of some optimization techniques applied to Jacketed CSTR control, in: Reliab. Infocom Technol. Optim. (ICRITO)(Trends Futur. Dir. 2015 4th Int. Conf., IEEE, 2015: pp. 1–6.

[160] B. Baadji, H. Bentarzi, A. Mati, Robust Wide Area Power System Stabilizers Design in Multimachine System based on Backtracking Search Optimization, in: 2018 Int. Conf. Appl. Smart Syst., IEEE, 2018: pp. 1–5.

[161] J. Kartite, M. Cherkaoui, Towards 100% Renewable Production: Dakhla Smart City Electrification, in: Proc. Third Int. Conf. Smart City Appl., Springer, 2018: pp. 1146–1156.

[162] Y.C. Kuyu, F. Vatansever, Analog Filter Group Delay Optimization using Metaheuristic Algorithms: A Comparative Study, in: 2018 Int. Conf. Artif. Intell. Data Process., IEEE, 2018: pp. 1–5.

[163] A.R. Jordehi, DG allocation and reconfiguration in distribution systems by metaheuristic optimisation algorithms: a comparative analysis, in: 2018 IEEE PES Innov. Smart Grid Technol. Conf. Eur., IEEE, 2018: pp. 1–6.

[164] M. Shafiullah, M. Abido, M. Hossain, A. Mantawy, An improved OPP problem formulation for distribution grid observability, Energies. 11 (2018) 3069.

[165] A. Khamis, H. Shareef, A. Mohamed, Z.Y. Dong, A load shedding scheme for DG integrated islanded power system utilizing backtracking search algorithm, Ain Shams Eng. J. 9 (2018) 161–172.

[166] S.S. Khan, M.A. Rafiq, H. Shareef, M.K. Sultan, Parameter optimization of PEMFC model using backtracking search algorithm, in: 2018 5th Int. Conf. Renew. Energy Gener. Appl., IEEE, 2018: pp. 323–326.

[167] N.N. Islam, M.A. Hannan, H. Shareef, A. Mohamed, An application of backtracking search algorithm in designing power system stabilizers for large multi-machine system, Neurocomputing. 237 (2017) 175–184.

[168] I. Elomary, A. Abbou, L. Idoumghar, Backtracking Search Algorithm Optimization for the Brushless Direct Current (BLDC) Motor Parameter Design, in: 2017 Int. Renew. Sustain. Energy Conf., IEEE,

2017: pp. 1–5.

[169] B. Hiçdurmaz, B. Durmuş, H. Temurtaş, S. Özyön, The Prediction of Butterworth Type Active Filter Parameters in Low-Pass Sallen-Key Topology by Backtracking Search Algorithm, (n.d.).

[170] R. El Maani, A.T. Zeine, B. Radi, A. El Hami, R. Ellaia, Backtracking search optimization algorithm for fluid-structure interaction problems, in: 2016 4th IEEE Int. Colloq. Inf. Sci. Technol., IEEE, 2016: pp. 690–695.

[171] T.T. Nguyen, H.N. Pham, A.V. Truong, T.A. Phung, T.T. Nguyen, A Backtracking Search Algorithm for Distribution Network Reconfiguration Problem, in: AETA 2015 Recent Adv. Electr. Eng. Relat. Sci., Springer, 2016: pp. 223–234.

[172] K. Dasgupta, S.K. Ghorui, An analysis of economic load dispatch with prohibited zone constraints using BSA algorithm, in: 2016 Int. Conf. Comput. Electr. Commun. Eng., IEEE, 2016: pp. 1–5.

[173] A.M. Shaheen, R.A. El-Sehiemy, S.M. Farrag, Integrated strategies of backtracking search optimizer for solving reactive power dispatch problem, IEEE Syst. J. 12 (2016) 424–433.

[174] M. Modiri-Delshad, S.H.A. Kaboli, E. Taslimi-Renani, N.A. Rahim, Backtracking search algorithm for solving economic dispatch problems with valve-point effects and multiple fuel options, Energy. 116 (2016) 637–649.

[175] Z. Wei, Q. Wei, The backtracking search optimization algorithm for frequency band and time segment selection in motor imagery-based brain–computer interfaces, J. Integr. Neurosci. 15 (2016) 347–364.

[176] N. Tyagi, H.M. Dubey, M. Pandit, Economic load dispatch of wind-solar-thermal system using backtracking search algorithm, Int. J. Eng. Sci. Technol. 8 (2016) 16–27.

[177] W. Jianjun, L. Li, L. Ding, Application of SVR with backtracking search algorithm for long-term load forecasting, J. Intell. Fuzzy Syst. 31 (2016) 2341–2347.

[178] N.N. Islam, M.A. Hannan, A. Mohamed, H. Shareef, Improved power system stability using backtracking search algorithm for coordination design of PSS and TCSC damping controller, PLoS One. 11 (2016) e0146277.

[179] A. El-Fergany, Optimal allocation of multi-type distributed generators using backtracking search optimization algorithm, Int. J. Electr. Power Energy Syst. 64 (2015) 1197–1205.

[180] A. Garbaya, M. Kotti, M. Fakhfakh, P. Siarry, The backtracking search for the optimal design of low-noise amplifiers, in: Comput. Intell. Analog Mix. Radio-Frequency Circuit Des., Springer, 2015: pp. 391–412.

[181] N. Niu, F.-F. Fu, H. Li, F.-C. Lai, J.-X. Wang, A Novel Topology Reconfiguration Backtracking Algorithm for 2D REmesh Networks-on-Chip, in: Int. Symp. Parallel Archit. Algorithm Program., Springer, 2017: pp. 51–58.

[182] K. Ayan, U. Kılıç, Optimal power flow of two-terminal HVDC systems using backtracking search algorithm, Int. J. Electr. Power Energy Syst. 78 (2016) 326–335.

[183] J. Yan, J. Zhang, Backtracking algorithms and search heuristics to generate test suites for combinatorial testing, in: Null, IEEE, 2006: pp. 385–394.

[184] F. Zaman, S.U. Khan, M.A.Z. Raja, S.A. Niazi, Backtracking Search Optimization Paradigm for Pattern Correction of Faulty Antenna Array in Wireless Mobile Communications, Wirel. Commun. Mob. Comput. 2019 (2019).

[185] M.M. Badawy, Z.H. Ali, H.A. Ali, QoS provisioning framework for service-oriented internet of things (IoT), Cluster Comput. (2019) 1–17.

[186] M. Eskandari, O. Sharifi, Effect of face and ocular multimodal biometric systems on gender classification, IET Biometrics. 8 (2019) 243–248.

[187] R.A. Osama, A.F. Zobaa, A.Y. Abdelaziz, A Planning Framework for Optimal Partitioning of Distribution Networks Into Microgrids, IEEE Syst. J. (2019).

[188] G.S. Walia, T. Singh, K. Singh, N. Verma, Robust multimodal biometric system based on optimal score level fusion model, Expert Syst. Appl. 116 (2019) 364–376.

[189] A.O. de Sá, A. Casimiro, R.C.S. Machado, L.F.R. da Costa Carmo, Bio-inspired System Identification Attacks in Noisy Networked Control Systems, in: Int. Conf. Bio-Inspired Inf. Commun., Springer, 2019: pp. 28–38.

[190] N.N.A. Nazri, N.N.N.A. Malik, L. Idoumghar, N.M.A. Latiff, S. Ali, Backtracking Search Optimization for Collaborative Beamforming in Wireless Sensor Networks, Telkomnika. 16 (2018) 1801–1808.

[191] A. Gosain, K. Sachdeva, Materialized view selection using backtracking search optimization algorithm, in: Intell. Eng. Informatics, Springer, 2018: pp. 241–251.

[192] A. Montanaro, Quantum walk speedup of backtracking algorithms, ArXiv Prepr. ArXiv1509.02374. (2015).

[193] Y.-K. Lin, T.-P. Nguyen, Reliability evaluation of a multistate flight network under time and stopover constraints, Comput. Ind. Eng. 115 (2018) 620–630.

[194] P. Mishra, V. Kumar, K.P.S. Rana, An efficient method for parameter estimation of a nonlinear system using Backtracking Search Algorithm, Eng. Sci. Technol. an Int. J. (2018).

[195] J. Zhou, H. Ye, X. Ji, W. Deng, An improved backtracking search algorithm for casting heat treatment charge plan problem, J. Intell. Manuf. (2017) 1–16.

[196] X. Song, X. Zhang, S. Zhao, L. Li, Backtracking search algorithm for effective and efficient surface wave analysis, J. Appl. Geophys. 114 (2015) 19–31.

[197] D. Karaboga, B. Akay, A comparative study of artificial bee colony algorithm, Appl. Math. Comput. 214 (2009) 108–132.

[198] D. Karaboga, B. Basturk, A powerful and efficient algorithm for numerical function optimization: artificial bee colony (ABC) algorithm, J. Glob. Optim. 39 (2007) 459–471.

[199] S. Das, P.N. Suganthan, Problem definitions and evaluation criteria for CEC 2011 competition on testing evolutionary algorithms on real world optimization problems, Jadavpur Univ. Nanyang Technol. Univ. Kolkata. (2010).

[200] A.C. Adamuthe, R.S. Bichkar, Personnel scheduling: Comparative study of backtracking approaches and genetic algorithms, Int. J. Comput. Appl. 38 (2012).

[201] S. Mandal, R.K. Sinha, K. Mittal, Comparative Analysis of Backtrack Search Optimization Algorithm (BSA) with other Evolutionary Algorithms for Global Continuous Optimization., Int J Comput Sci Inf Technol. 6 (2015) 3237–3241.

[202] P. Agarwal, S. Mehta, Empirical analysis of five nature-inspired algorithms on real parameter optimization problems, Artif. Intell. Rev. 50 (2018) 383–439.

[203] G. Lindfield, J. Penny, Introduction to Nature-Inspired Optimization, Academic Press, 2017.

[204] M. Jamil, X.-S. Yang, A literature survey of benchmark functions for global optimization problems, ArXiv Prepr. ArXiv1308.4008. (2013).

[205] M.M. Ali, C. Khompatraporn, Z.B. Zabinsky, A numerical evaluation of several stochastic algorithms on selected continuous global optimization test problems, J. Glob. Optim. 31 (2005) 635–672.

[206] Global Optimization Benchmarks and AMPGO, (n.d.). http://infinity77.net/global_optimization (accessed November 24, 2018).

[207] J. Derrac, S. García, D. Molina, F. Herrera, A practical tutorial on the use of nonparametric statistical tests as a methodology for comparing evolutionary and swarm intelligence algorithms, Swarm Evol. Comput. 1 (2011) 3–18.

[208] A.S. Joshi, O. Kulkarni, G.M. Kakandikar, V.M. Nandedkar, Cuckoo search optimization-a review, Mater. Today Proc. 4 (2017) 7262–7269.

[209] D. Chen, F. Zou, R. Lu, S. Li, Backtracking search optimization algorithm based on knowledge learning, Inf. Sci. (Ny). 473 (2019) 202–226.

[210] D. Chakrabarti, R. Kumar, A. Tomkins, Evolutionary clustering, in: Proc. 12th ACM SIGKDD Int. Conf. Knowl. Discov. Data Min., ACM, 2006: pp. 554–560.

[211] K.S. Xu, M. Kliger, A.O. Hero Iii, Adaptive evolutionary clustering, Data Min. Knowl. Discov. 28 (2014) 304–336.

[212] A.A. Freitas, Data mining and knowledge discovery with evolutionary algorithms, Springer Science & Business Media, 2013.

[213] H.M. Mohammed, S.U. Umar, T.A. Rashid, A systematic and meta-analysis survey of whale optimization algorithm, Comput. Intell. Neurosci. (2019) 2019.

[214] A.S. Shamsaldin, T.A. Rashid, R.A.A.-R. Agha, N.K. Al-Salihi, M. Mohammadi, Donkey and smuggler optimization algorithm: a collaborative working approach to path finding, J. Comput. Des. Eng. (2019).

[215] J.M. Abdullah, T. Ahmed, Fitness dependent optimizer: inspired by the bee swarming reproductive process, IEEE Access 7 (2019) 43473–43486.

[216] T.A. Rashid, D.K. Abbas, Y.K. Turel, A multi hidden recurrent neural network with a modified grey wolf optimizer, PLoS One 14 (2019) e0213237